\documentclass{article}

\usepackage[dvipsnames,table]{xcolor} 
\usepackage[preprint]{corl_2026} 

\usepackage{xspace}
\usepackage{graphicx}
\usepackage{wrapfig}
\usepackage{float}
\usepackage{adjustbox}
\usepackage{etoc}

\usepackage{enumitem}
\usepackage{algorithm}
\usepackage{algpseudocode}
\usepackage{booktabs}
\usepackage{amsmath}
\usepackage{iftex}
\ifPDFTeX\else
  \usepackage{newtxtext,newtxmath}
\fi

\usepackage[mathcal]{eucal}

\newcommand{\coolname}{UniWALM\xspace}

\newcommand{\walmlong}{World Action Language Model\xspace}
\newcommand{\walm}{WALM\xspace}

\newcommand{\parahead}[1]{\paragraph*{#1.}}

\usepackage[capitalize]{cleveref}

\crefname{section}{Sec.}{Secs.}
\Crefname{section}{Section}{Sections}
\crefname{table}{Tab.}{Tabs.}
\Crefname{table}{Table}{Tables}
\crefname{figure}{Fig.}{Figs.}
\Crefname{figure}{Figure}{Figures}
\crefname{equation}{Eq.}{Eqs.}
\Crefname{equation}{Equation}{Equations}
\Crefname{algorithm}{Algorithm}{Algorithms}
\crefname{algorithm}{Algo.}{Algos.}

\usepackage{bm}

\usepackage{listings}

\definecolor{json_key}{rgb}{0,0.44,0.34}       
\definecolor{json_string}{rgb}{0.75,0.0,0.0}   
\definecolor{json_number}{rgb}{0.0,0.0,1.0}   
\definecolor{json_background}{rgb}{0.97,0.97,0.97}

\lstdefinelanguage{json}{
    basicstyle=\normalfont\ttfamily\scriptsize,
    numberstyle=\scriptsize\color{gray},
    stepnumber=1,
    numbersep=8pt,
    showstringspaces=false,
    breaklines=true,
    backgroundcolor=\color{json_background},
    frame=single,
    rulecolor=\color{lightgray},
    string=[s]{"}{"},
    stringstyle=\color{json_key},
    morestring=[s]{""}{""},
    literate=
     *{0}{{{\color{json_number}0}}}{1}
      {1}{{{\color{json_number}1}}}{1}
      {2}{{{\color{json_number}2}}}{1}
      {3}{{{\color{json_number}3}}}{1}
      {4}{{{\color{json_number}4}}}{1}
      {5}{{{\color{json_number}5}}}{1}
      {6}{{{\color{json_number}6}}}{1}
      {7}{{{\color{json_number}7}}}{1}
      {8}{{{\color{json_number}8}}}{1}
      {9}{{{\color{json_number}9}}}{1}
      {:}{{{\color{black}{:}}}}{1}
      {,}{{{\color{black}{,}}}}{1}
      {\{}{{{\color{black}{\{}}}}{1}
      {\}}{{{\color{black}{\}}}}}{1}
      {[}{{{\color{black}{[}}}}{1}
      {]}{{{\color{black}{]}}}}{1},
}

\definecolor{goldcell}{RGB}{255, 245, 200}
\definecolor{silvercell}{RGB}{235, 235, 235}

\definecolor{cb-black}      {RGB}{  0,   0,   0}
\definecolor{cb-blue-green} {RGB}{  0,  073,  073}
\definecolor{cb-green-sea}  {RGB}{  0, 146, 146}
\definecolor{cb-rose}       {RGB}{255, 109, 182}
\definecolor{cb-salmon-pink}{RGB}{255, 182, 119}
\definecolor{cb-purple}     {RGB}{ 194,   106, 119}
\definecolor{cb-blue}       {RGB}{ 0, 109, 219}
\definecolor{cb-lilac}      {RGB}{182, 109, 255}
\definecolor{cb-blue-sky}   {RGB}{109, 182, 255}
\definecolor{cb-blue-light} {RGB}{182, 219, 255}
\definecolor{cb-burgundy}   {RGB}{146,   0,   0}
\definecolor{cb-brown}      {RGB}{146,  73,   0}
\definecolor{cb-clay}       {RGB}{219, 209,   0}
\definecolor{cb-green-lime} {RGB}{ 36, 255,  36}
\definecolor{cb-yellow}     {RGB}{255, 255, 109}
\definecolor{cb-red}{RGB}{240, 94, 0} 
\definecolor{cb-edit}     {RGB}{ 60, 170, 200}
\definecolor{cb-maroon}{RGB}{192, 0, 0} 
\definecolor{cb-dark-blue}{RGB}{0, 79, 106} 

\definecolor{cb-cream-yellow}{RGB}{255, 239, 220} 

\definecolor{material-edit}{RGB}{255, 0, 255}
\definecolor{image-edit}{RGB}{0, 255, 255}
\definecolor{prompt-edit}{RGB}{255,0,0}

\definecolor{grey}{RGB}{100, 100, 100}

\title{\coolname: Unified World Action Language Model for Embodied Robot Control}

\vspace{-0.1in}
\author{
\mdseries%
Rahul Sajnani$^{1,2}$, Diego Romeres$^{2}$, Mingyuan Zhang$^{2,3}$, Kai Gao$^{2}$
Carl Qi$^{2,4}$, Baichuan Huang$^{2}$ \\ Srinath Sridhar$^{1}$, Dylan Randle$^{2}$,
Lillian J.~Ratliff$^{2,5}$, Jose Barreiros$^{2}$, Manikantan Nambi$^{2}$, Jeroen Vanbaar$^{2}$\\[0.4em]
$^{1}$Brown University \quad $^{2}$Amazon \quad $^{3}$Northeastern University\quad
$^{4}$UT Austin \quad $^{5}$University of Washington\\[0.3em]
\normalfont\small Project page: \href{https://rahulsajnani.github.io/projects/uniwalm/}{\texttt{rahulsajnani.github.io/projects/uniwalm/}}%
}

\begin{document}
\newgeometry{left=0.9in,right=0.9in,top=1in,bottom=1in} 
\maketitle
\etocsettocdepth.toc{none}

\begin{figure}[H]
    \vspace{-0.4in}
    \centering
    \includegraphics[width=\textwidth]{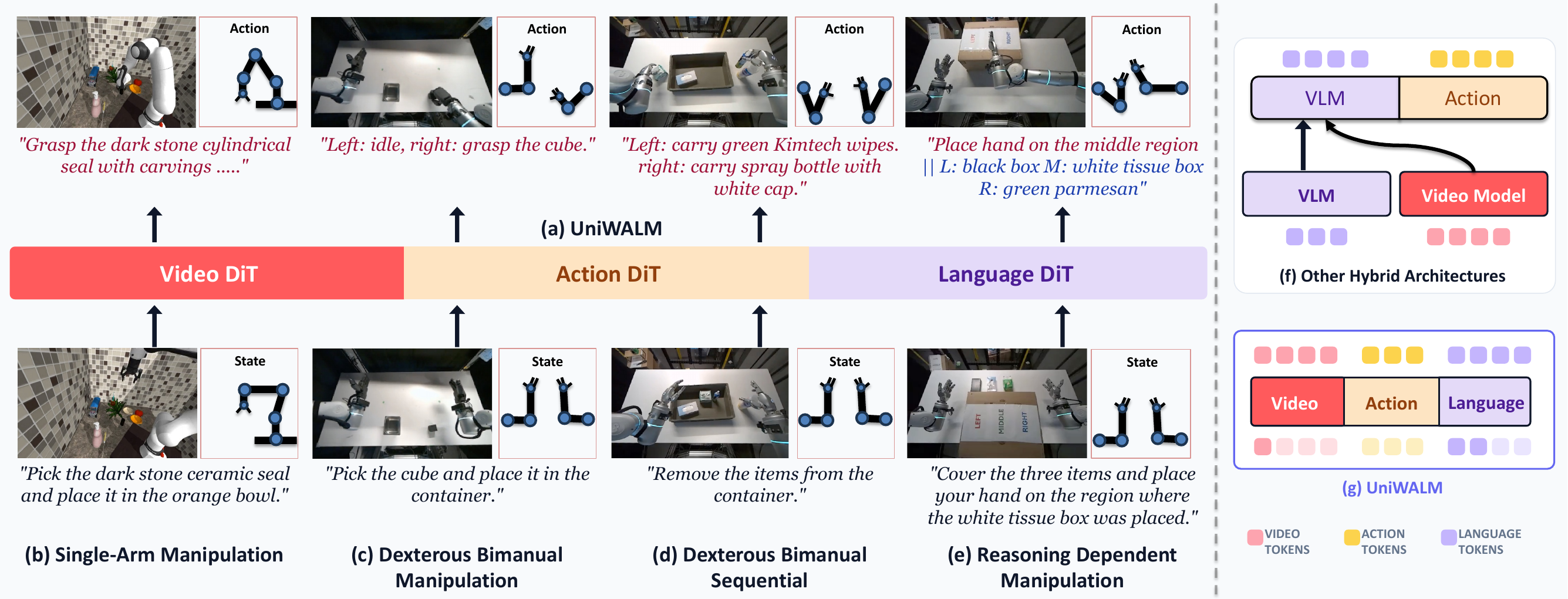}
    \vspace{-0.2in}
    \caption{\textbf{(a) \coolname} is a unified \walmlong that extends a single video-model backbone to \emph{act} and \emph{reason} by co-generating video, action, and language through one shared diffusion process --- no separate language model fused onto a world-action model. This supports \textbf{(b) single and (c) bimanual dexterous manipulation} with \textbf{(d) multi-stage manipulation} and \textbf{(e) memory-dependent reasoning}. Prior work \textbf{(f) composes a VLM with a VLA and a video model} in pipeline or parallel; \coolname{} \textbf{(g) tunes a single unified representation} end-to-end. Each manipulation example displays the input task instruction at the bottom and the generated video, action, and {\color{cb-maroon}subtask} at the top. In \textbf{(e)} the predicted {\color{cb-maroon}subtask} and {\color{cb-dark-blue}textual memory} (shown as `{\color{cb-maroon}\textit{subtask}}~\textbar\textbar~{\color{cb-dark-blue}\textit{memory}}') are re-consumed as the next-step prompt.}
    \label{fig:teaser}
\end{figure}

\begin{abstract}
Motivated by the human ability to integrate experience across language, perception, and action, we introduce \coolname{}, a unified world-action-language model that learns video, action, and language in a single shared representation, where each modality is both a conditioning signal and a supervision target.
We ask: \emph{can jointly generating language enrich this shared representation, improving manipulation and supporting memory-dependent behavior?}
Unlike prior approaches that \emph{attach} a reasoning or vision-language model to a video-action backbone and run them as distinct stages, \coolname{} unifies all three modalities inside a single mixture-of-transformers backbone whose per-modality diffusion heads co-denoise synchronously, decoding subtask and textual memory in the same pass as action with no separate planning stage; the generated language is re-consumed on the next step, closing a language-as-memory loop without an external planner or memory bank.
To keep the action stream reliable under this richer objective, we further introduce Flow Matching with Correction (FMC), a training objective that teaches recovery from perturbed action trajectories at no inference-time cost.
\coolname{} consistently improves over same-backbone baselines (e.g., FastWAM~\cite{FastWAM}) on compositional pick-and-place (MolmoSpaces) and across the memory-dependent subset of RMBench~\cite{rmbench}, where it reaches \textbf{34.5\%} without any pretrained language or reasoning model and exceeds Mem-0's keyframe-equipped setup (28.5\%) using only a single anchor frame.
On a real-world Find-the-Item task, generating textual memory lifts success from \textbf{11.1\%} to \textbf{61.1\%}, while state-of-the-art VLAs stay at or below chance ($\leq 33.3\%$).
\end{abstract}

\section{Introduction}
\label{sec:introduction}

Humans integrate perception, language, and action in an \emph{any-to-any} manner: regions like the posterior parietal and premotor cortex continuously fuse afferent inputs with efferent motor copies~\cite{Andersen2003Sensorimotor}, and cognitive representations are inherently shaped by what an agent can \emph{do} in the world~\cite{Goodale1992Separate, Cisek2007}. For embodied agents, language, video, and action are three complementary views of a single interaction; a unified representation should let each one inform the others: language guiding physical prediction, visual dynamics grounding action, and action anchoring predictions in embodied behavior. Multimodal models have operationalized this framing for vision and text~\cite{massively-multimodal, wu2024nextgptanytoanymultimodalllm, zhan2025anygptunifiedmultimodalllm} but largely omit action and motor control.

The same substrate that integrates these modalities also \emph{remembers}: the hippocampus, which the brain uses to recall past experience, is also recruited to imagine the future and locate the self in time~\cite{eagleman2015brain, zadbood2017transmit}. We routinely enlist \emph{language itself} as working memory (counting out loud, repeating a phone number, silently rehearsing a passage while reading to retain it) to track state through long-horizon, memory-dependent tasks. A truly unified representation should therefore not only let modalities inform one another but also let \emph{generated} language serve as the working memory that threads behavior across step boundaries. Concretely, because language is co-denoised with action in one representation, the generated string serves as a learned, in-loop plan (next subtask) and textual memory (what has been done so far), while an anchor-frame visual memory preserves scene context across occlusion; re-consumed on the next step, these close a memory loop with no external planner, VLM, or retrieval buffer.

Recent robotic foundation models connect video, language, and action, but typically along only a subset of these directions. Vision-language-action models~\cite{pi0.5,lin2026cotraininglbm} consume language as fixed conditioning but do not generate it. Chain-of-thought VLAs~\cite{innermonologue, rt22023arxiv, ecot, rth, cotvla} generate intermediate language \emph{within} a control step, but autoregressively, so action must wait for token decoding, and the generated string is not re-consumed across steps. World-action models (DreamZero~\cite{dreamzero}, FastWAM~\cite{FastWAM}) co-model video and action but drop language entirely. Memory-augmented systems~\cite{Torne2026MEMME, pi07, rmbench} attach retrieval buffers or an external planner to a pretrained backbone; $\pi_{0.7}$~\cite{pi07} supplies language from a high-level semantic policy while a separate world model generates subgoal images that condition a VLA (\cref{fig:teaser}(f)). On memory-dependent evaluations like RMBench~\cite{rmbench}'s $M(n)$ subset, where decisions require $n$ task-relevant past observations, policies that condition only on the current frame and prompt collapse.

A natural response is to \emph{compose} a pretrained vision-language model with a video-action backbone and supply reasoning via auxiliary conditioning or an additional loss (\cref{fig:teaser}(f)). This family (hybrid planner + VLA systems~\cite{pi07}, concurrent compositional WALMs) is more accurately described as \emph{WAM + VLM features}: it typically uses two vision backbones optimized for different objectives and only loosely coupled, treats language as an external hidden state rather than a generation target, and bolts memory on as a separate retrieval module. \coolname{} (\cref{fig:teaser}(g)) uses a \emph{single video backbone jointly optimized for all three modalities}, with video, action, and language as first-class diffusion heads co-denoised in one pass; the generated language is re-consumed next step, closing a language-as-memory loop with no external planner, VLM, or memory bank. Where prior world-action models supervise the backbone with \emph{frame-level} predictive structure (future video), \coolname{} adds a \emph{task-level} predictive stream (future subtask and task state as language), a second, longer-horizon supervision signal that frame prediction alone cannot cheaply express.

In this work we study two questions: \emph{(i) does jointly generating language enrich the shared representation of a video-action model and improve manipulation?} (\cref{fig:teaser}(g)) and \emph{(ii) can a policy use its own generated language, re-consumed across steps, as working memory on memory-dependent tasks?} (\cref{fig:arch_reasonvam})

\parahead{Contributions} We introduce \coolname{}, a multimodal world-model architecture that takes a step toward any-to-any modeling across language, video, and actions, where each modality serves as both a conditioning signal and a supervision target. Specifically:

\begin{itemize}[leftmargin=*]
    \item \textbf{World-Action-Language Model (\walm):} We introduce \walm, a unified \walmlong implemented as a mixture of transformers that takes video, language, and actions as input and generates all three as output. The language stream emits a \textbf{subtask and textual memory} in the same denoising pass as action, and re-consumes them as prompt on the next control step, closing a language-as-memory loop. \coolname's language and action experts are trained from scratch on a video-only backbone: \emph{no pretrained language or reasoning model is used}.
    \item \textbf{Flow Matching with Correction (FMC).} We introduce FMC to improve the reliability of generated actions. During training, FMC perturbs action trajectories and supervises a corrective velocity that returns the perturbed denoising path toward the clean trajectory. This trains recovery from action-space perturbations without adding model capacity or sampling steps at inference.
    \item \textbf{Evaluation in simulation and real-world manipulation.} \coolname{} achieves the highest success on all three MolmoSpaces~\cite{MolmoSpaces} compositional pick-and-place variants, improving over the same-backbone FastWAM baseline by \textbf{+14.8\%} on average (\cref{tab:molmobot}). On RMBench's memory-dependent M(n) subset, \coolname{} reaches \textbf{34.5\%} without pretrained language or reasoning model, improving over same-backbone FastWAM (13.3\%) by \textbf{+21.2 points} and above Mem-0's keyframe-equipped setup (28.5\%) using a single anchor frame instead of per-task privileged keyframes (\cref{tab:rmbench}). On a real-world bimanual Find-the-Item task, generating textual memory lifts success from \textbf{11.1\%} (FastWAM, same video backbone, no language) to \textbf{61.1\%}, while state-of-the-art VLAs stay at or below chance ($\leq 33.3\%$) (\cref{tab:real_evaluation}).
\end{itemize}

Together, these results motivate further investigation of any-to-any world modeling as a foundation for generalist embodied agents.

\section{Related Works}
\label{sec:related_works}

\parahead{Generalist Visuomotor Policies} Generalist robot policies~\cite{pi0.5, pi0, Molmobot, Nvidia2025GR00TNA, qwenvla, rt22023arxiv, octo_2023, lbmtri2025} improve over single-task models~\cite{diffusion_policy, zhao2023act} by initializing from vision-language backbones~\cite{qwen3, molmo2, eagle2.5, shi2025eagle, dino, vjepa2} and training for action. These policies map observations to actions but do not model world dynamics or reason over long-horizon subtasks.

\parahead{World-Action Models} A recent line of works~\cite{FastWAM, dreamzero, cosmospolicy, Motus, unipi, leWorldModel, Lin2026JEPAWAMLV} adds a video-prediction objective, jointly modeling world dynamics and action. FastWAM~\cite{FastWAM}, closest to our work, couples a pretrained Wan2.2~\cite{wan2025} video DiT with a parallel action DiT through unified attention. These models treat language purely as a conditioning input, never as a generation target and provide no mechanism for subtask reasoning or memory.

\parahead{Reasoning, Memory, and Hybrid Policies} Long-horizon and memory-dependent tasks have motivated explicit reasoning and memory~\cite{rmbench, memoryvla}, and hybrid systems~\cite{pi07, mem, tau0, lingbotva, flexpi} that combine a high-level VLM planner, a low-level policy, and a video model to imagine action effects. Because these components are trained separately on heterogeneous data, they lack a shared representation and execute planning and control asynchronously.

\parahead{Unified and Omnimodal Models} Any-to-any modeling originates in multimodal LLMs~\cite{zhan2025anygptunifiedmultimodalllm, wu2024nextgptanytoanymultimodalllm} which omit generating actions, and vision works like 4M~\cite{massively-multimodal}, which unify many modalities via masked token modeling but target perception rather than control and do not reason. Concurrent work~\cite{flexpi} additionally generates actions, DINO features, semantics, and video, but cannot reason or preserve memory during manipulation. Concurrent omnimodal models such as Cosmos~3~\cite{cosmos3} extend any-to-any modeling to action within a single framework, but decode text autoregressively in one tower and continuous modalities by diffusion in another, running reasoning and generation as separate stages in a 64B model. \coolname instead generates language, video, and action \emph{entirely by diffusion in one shared representation}, denoising language and action \emph{synchronously} so that reasoning and control co-refine in a single unified model.

An extended discussion contrasting \coolname{} with CoT / inner-monologue policies, multi-scale memory VLAs, and recent unified models is deferred to \cref{app:related_work}.

\section{\coolname}
\coolname{} jointly models the future video, robot actions, and language within a shared multimodal representation, enabling reasoning and control to be learned together.
Next, we describe its world-action-language formulation, multimodal co-training, and corrective flow-matching objective.

\subsection{Reasoning with Video Action Models}

Video Action Models typically map the current observation and a coarse prompt to near-term video and actions, with no explicit mechanism to generate language that can reason about subtasks or maintain task memory.
We introduce \coolname{}, a \walmlong (\walm) that jointly generates future video, actions, and language within a unified model.
Its language generation expert is a diffusion model that provides language reasoning in the form of subtasks and memory.

\begin{figure*}[t]
    \vspace{-0.1in}
  \centering
  \includegraphics[width=\textwidth]{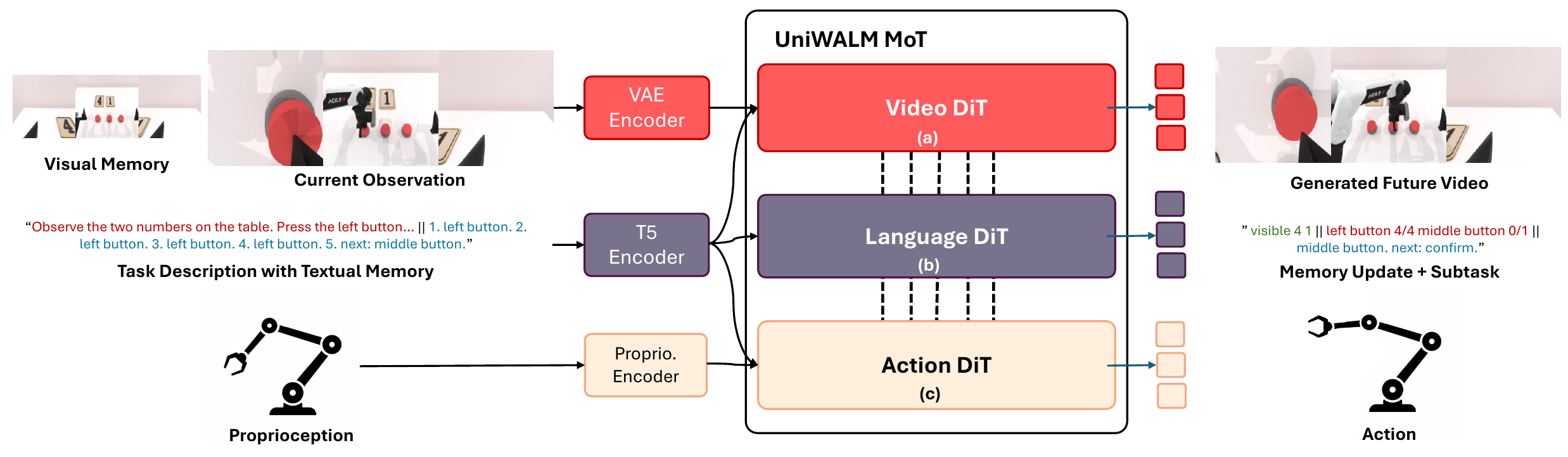}
  \vspace{-0.3in}
  \caption{
    \textbf{Overview of \emph{\coolname}}.
    \coolname{} is a unified mixture-of-transformers (MoT) world-action-language model that co-denoises video, language, and action in a single shared representation.
    Given the current observation, task instruction, proprioception, an optional \textbf{visual memory} $\mathcal{F}$ (one or more frames; in our experiments a single anchor at episode start), and the previously generated \textbf{textual memory}, \coolname{} processes inputs through a \textbf{(a) video}, \textbf{(b) language}, and \textbf{(c) action DiT} that share features via unified self-attention.
    The language stream emits a {\color{cb-maroon}subtask} together with a {\color{cb-dark-blue}textual memory} string, which is re-consumed as the prompt on the next control step, closing a language-as-memory loop across time. Video generation is disabled at inference, so the language and action DiTs run without producing future frames.
  }
  \vspace{-0.2in}
  \label{fig:arch_reasonvam}
\end{figure*}

\parahead{\walmlong (\walm)} World Action Language Models are a new class of models that take as input the current observation ($\mathcal{O}$, $k$ RGB views\footnote{We use $k=2$ in MolmoSpaces sim and $k=3$ in real-world and RMBench sim.}), robot state ($\mathcal{S}$), visual memory $\mathcal{F}$ (a set of one or more frames), and task prompt ($T$) to predict future video ($\mathcal{V}$), action ($\mathcal{A}$), language ($L$), and a learned task-completion signal ($\beta$) using a Mixture of Transformers (MoT) architecture with a single shared representation.
\coolname (\cref{fig:arch_reasonvam}) is a step towards a single-representation, synchronous \walm. Formally, \coolname{} jointly generates
\[
    \mathcal{V}_{1:H},\ (\mathcal{A},\ \beta)_{1:N},\ L\ :=\ \text{\coolname}\big(\mathcal{O},\ \mathcal{S},\ \mathcal{F},\ T\big),
\]
where $\mathcal{V}_{1:H}$ is $H$ future frames, $(\mathcal{A},\beta)_{1:N}$ is a chunk of $N$ future actions each augmented with a 1-D \emph{task-completion signal} $\beta$ that fires at subtask completion, and $L$ is a compact string (subtask $\|$ textual memory) re-consumed as prompt on the next control step. The visual memory $\mathcal{F}$ is an optional input, enabled for memory-dependent tasks (see~\cref{subsec:memory}) and omitted for standard manipulation. $\beta$ gates when the textual memory is advanced, so memory updates only at subtask completions.

\parahead{Generating Video, Action, and Language}
Given a pretrained video model, the observation frames are encoded first and passed through the video DiT.
We replicate the video DiT's structure, matching its depth (30 layers) while reducing hidden dimension, to define the action and language DiTs, whose weights are initialized from scratch.
Each layer of the video, action, and language DiT processes features via unified self-attention enabling feature sharing between modalities.

\parahead{Training the Language Expert for Reasoning}
\label{method:reasoning}
Per-frame RGB prediction captures short-horizon dynamics but cannot cheaply carry task-level information (what subtask is active, what has been done), and we need this signal produced in the same pass as action rather than through a separate autoregressive planner.
We therefore adopt a \emph{discrete language diffusion}~\cite{diffusiongemma} expert that denoises in tandem with video and action, giving us speed and architectural symmetry with the video model.
At each training step we perturb ground-truth text ($L_{GT}$) depending on the diffusion time step, $t$, and predict the correct token embedding ($\hat{L}$) using a cross entropy objective ($\mathcal{L}_{lang}$):
\begin{align*}
    & L_t := \text{RandomReplace}(L_{GT}, t) \\
    & \hat{L} := \text{DiffusionStep}(L_t, t)\\
    & \mathcal{L}_{lang} := \text{CrossEntropy}(\hat{L}, L_{GT}).
\end{align*}
This teaches the model to freeze already-correct tokens and update the incorrect ones based on the condition.
We use the language expert to predict the next subtask the robot should perform, and additionally emit textual memory for memory-dependent tasks.

\subsection{Multi-Modal Co-Training}
\label{method:cotraining}
\parahead{Co-training Objective}
\coolname trains Video DiT, Action DiT, and Language DiT in a co-training objective that combines the following:
\begin{align*}
    & \mathcal{L}_{video} := ||{}^{v}v^{\theta}_t - {}^{v}v^{GT}_t ||_1 \\
    & \mathcal{L}_{action} := ||{}^{a}v^{\theta}_t - {}^{a}v^{GT}_t ||_1 \\
    & \mathcal{L} := \mathcal{L}_{video} + \mathcal{L}_{action} + \alpha \mathcal{L}_{lang}
\end{align*}
Here, $\mathcal{L}_{video}, \mathcal{L}_{action}$ are flow-matching losses;
$\mathcal{L}_{lang}$ is the discrete diffusion loss (\cref{method:reasoning}).
The ${}^{v/a}v^{\theta}_t$ and ${}^{v/a}v^{GT}_t$ are predicted and ground truth velocities for video/action modality respectively.

We use an $\ell_1$ penalty for the flow-matching terms, which we find more stable than $\ell_2$ for learning actions. 
Together, these three terms supervise the shared representation at \emph{two levels of abstraction} in parallel: $\mathcal{L}_{video}$ is \textbf{frame-level} predictive structure (per-step dense RGB prediction, which prior world-action models like FastWAM~\cite{FastWAM} and DreamZero~\cite{dreamzero} already use), while $\mathcal{L}_{lang}$ is \textbf{task-level} predictive structure (future subtask and task state as a short string), a longer-horizon signal that frame prediction alone cannot cheaply express; the action loss anchors both to physical behavior.
Training with both predictive levels in one backbone, rather than only one, is what lets a single end-to-end policy internalize the structure usually externalized to a planner or VLM.
During training, either all or a subset of the mixture of transformers can be trained in tandem depending on the availability of data to reap benefits from all modalities.

\subsection{Memory}
\label{subsec:memory}
\coolname{} integrates three memory mechanisms directly into the shared representation: a \emph{textual memory} that is re-consumed as the next-step prompt (closing a language-as-memory loop without an external planner), a \emph{visual memory} that admits one or more keyframes, and a \emph{task-completion channel} that gates when the textual memory is updated. Full implementation details (RoPE-shifted keyframe tokenization, task-completion supervision, and RMBench-specific training configuration) are in \cref{app:memory_details}.

\parahead{Textual Memory Re-consumption}\label{para:textual_memory} Policies that reason once per step cannot carry information across step boundaries, which is exactly what long-horizon and memory-dependent tasks require. At inference, the language head emits a compact string (subtask $\|$ textual memory) in the same denoising pass as action. For memory-dependent tasks, this string is re-consumed as prompt on the next control step, closing a language-as-memory loop without an external planner or memory bank.

\parahead{Visual Memory}\label{para:visual_memory} Textual memory alone struggles with scene state that is hard to describe in words (precise object locations, occluded layouts), so we add a lightweight visual-memory stream $\mathcal{F}$ the policy can always attend to. For tasks that hide or move objects mid-episode, $\mathcal{F}$ preserves the layout otherwise lost as observations advance.

\parahead{Task-Completion Signal ($\beta$)}\label{para:task_completion} If the textual memory is rewritten at every control step, intermediate reasoning overwrites past progress and the summary ceases to be a reliable record of what has been done, especially since the language stream has no explicit supervision for \emph{when} a subtask is actually complete. We therefore co-train a 1-D task-completion channel $\beta$, appended to the action vector, that fires at subtask completion and gates when the textual memory is updated, so the memory advances only at subtask boundaries rather than every step. Because $\beta$ is concatenated into the action vector, it is supervised jointly with the action by $\mathcal{L}_{action}$ (\cref{method:cotraining}) and requires no additional loss term.

Taken together, these three mechanisms correspond to three distinct learned abstractions over the task that live in the shared representation: the generated subtask is a \emph{plan} (what to do next), the generated textual memory is a \emph{running summary} (what has been done), and $\beta$ is a \emph{task-completion signal} (when to commit a subtask as done). All three are first-class outputs of the same backbone that generates video and action, not external modules bolted onto a pretrained VLM or retrieval buffer.

\subsection{Flow Matching with Correction (FMC)}
\label{sec:background_flow_matching}
Flow matching~\cite{diffusion_policy, flow_matching, Esser2024ScalingRF} learns a parameterized velocity field $v^\theta_t$ that transports samples from a Gaussian prior $\mathcal{N}(0,I)$ to the data distribution $p_{\text{data}}(x)$ via a continuous normalizing flow.
During training, a data sample $x\sim p_{data}(x)$ is linearly noised to $x_t= (1-t)x + t \epsilon$ at timestep $t$ where $t\in [0, 1]$.
The parameterized function learns to estimate the velocity $v_t := \frac{d}{dt}x_t = (\epsilon - x)$ using a mean-square-error objective $||v^{\theta}_t - v_t||_2$ where $v^{\theta}_t:=p_{\theta}(x_t, t)$ is the predicted velocity.
Then data is sampled  $x_1\sim \mathcal{N}(0, I)$ and denoised to the clean sample $x:= x_1 - \int_0^1 v^{\theta}_tdt$.
\label{method:fc}

Flow matching is a ubiquitous objective in training large scale video and action policies because it supports high-quality generation with relatively few sampling steps.
However, policies trained with flow matching often exhibit minor deviations during the generation process resulting in imprecise rollouts.
To curb this, we implement a flow correction strategy that simulates a perturbed noisy trajectory to teach the model to self-correct to a clean denoising trajectory, which we call Flow Matching with Correction (FMC).
\begin{wrapfigure}{r}{0.5\textwidth}
    \vspace{-0.1in}
  \centering
  \includegraphics[width=0.48\textwidth]{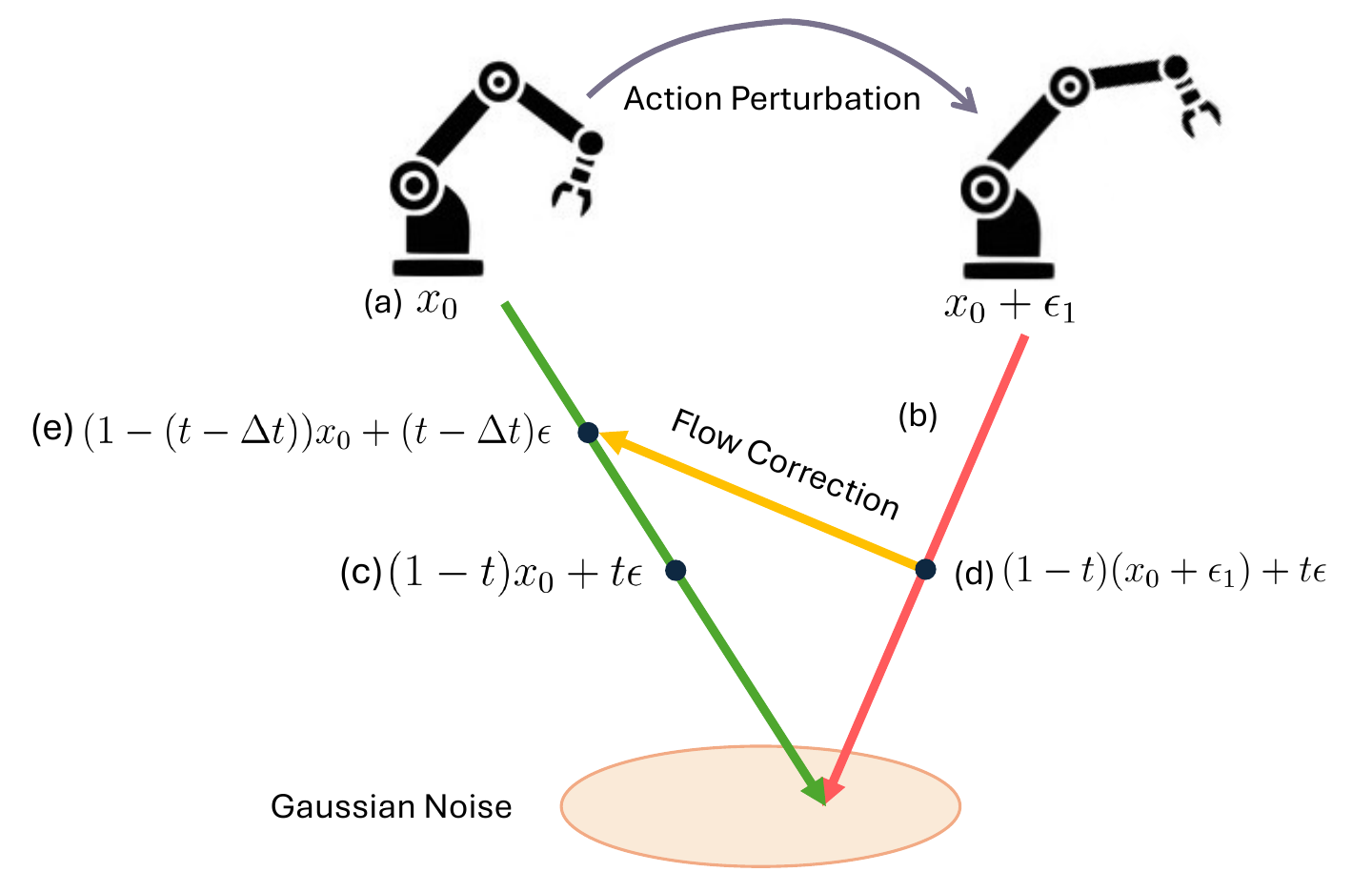}
  \caption{
    \textbf{Flow Matching with Correction (FMC)}.
    A clean action $x_0$ (green path) and its perturbed counterpart $x_0 + \epsilon_1$ (red path) are noised to the same time $t$. The model is trained to transport the perturbed sample back onto the clean trajectory within $\Delta t$ (yellow arrow), while the clean branch follows standard flow matching. This requires no extra conditioning or model capacity and leaves inference unchanged.
  }
  \label{fig:flow_correction}
  \vspace{0.1in}
\end{wrapfigure}

Given a clean action $x_0$ (cf.~\cref{fig:flow_correction}(a)), we first construct a perturbed action $x^p_0 := x_0 + \epsilon_1$  where $\epsilon_1 \sim \mathcal{N}(0,sI)$, where
$s$ is the noise scale for perturbation (cf.~\cref{algo:flow_correction}).
The goal of flow correction is to avoid denoising to the perturbed action $x^p_0$ and instead denoise to the correct action $x_0$. Using the same noise sample $\epsilon$, the corresponding point at time $t$ is
\[
    x^p_t := (1-t)x^p_0 + t\epsilon,
    \]
 which lies off the nominal denoising trajectory. Applying the standard flow-matching target at this point would continue denoising without explicitly removing this deviation. Instead, FMC chooses a target velocity such that one reverse step of size $\Delta t$ from $x_t^{\mathrm p}$ returns to the nominal trajectory at $x_{t-\Delta t}=
x_t^{\mathrm p}-\Delta t v_{\mathrm{corr}}$.
Solving for the correction target yields
\begin{equation*}
v_{\mathrm{corr}}
:=
\frac{x_t^{\mathrm p}-x_{t-\Delta t}}{\Delta t}
=
(\epsilon-x_0)
+
\frac{1-t}{\Delta t}\epsilon_1.
\end{equation*}

The expanded form makes the mechanism explicit: $v_{corr}$ is the standard flow-matching velocity $\epsilon - x_0$ plus a correction term that cancels the perturbation $\epsilon_1$, and it reduces exactly to the standard target when $\epsilon_1 = 0$.

 \begin{algorithm}[h]   

  \caption{Flow Matching with Correction}
  \label{algo:flow_correction}
  \begin{algorithmic}[1]
  \Require Clean sample $x_0$, model $v_\theta$, correction prob $p = 0.5$, step $\Delta t$, Noise Scale for Perturbation $s=0.2$
  \State Sample $t \sim \mathcal{U}(0, 1)$, \quad $\epsilon \sim \mathcal{N}(0, I)$
  \If{(($\text{rand}() < p$) \&\& ($t \geq 0.5$))} \Comment{Correction branch}
      \State Sample noise $\epsilon_1 \sim \mathcal{N}(0, sI)$
      \State $x^{\text{p}}_t \gets (1 - t)(x_0 + \epsilon_1) + t\, \epsilon$ \Comment{Perturbed action}
      \State $x_{t - \Delta t} \gets (1 - (t - \Delta t))\, x_0 + (t - \Delta t)\, \epsilon$
      \State $v_{\text{target}} \gets \dfrac{x^{\text{p}}_t - x_{t - \Delta t}}{\Delta t}$ \Comment{Correction Velocity}
      \State $x_{\text{in}} \gets x^{\text{p}}_t$
  \Else \Comment{Standard Flow Matching Branch}
      \State $x_{\text{in}} \gets (1 - t)\, x_0 + t\, \epsilon$
      \State $v_{\text{target}} \gets \epsilon - x_0$
  \EndIf
  \State $\mathcal{L} \gets \| v_\theta(x_{\text{in}}, t) - v_{\text{target}} \|_1$
  \end{algorithmic}
  \end{algorithm}
  \vspace{-0.1in}

Since the correction term scales with $(1-t)/ \Delta t$, we apply the correction branch only for $t \geq 0.5$, ensuring the sample is close to the noise manifold and the correction magnitude remains bounded; at smaller $t$ the injected noise $\epsilon_1$ dominates and correction provides no meaningful signal.
The procedure is detailed in~\cref{algo:flow_correction}.
FMC teaches the policy to  both denoise and self-correct in tandem.
Since the corrected velocity field coincides with the standard flow-matching field on the clean path, inference is unchanged: no additional conditioning, sampling steps, or model capacity is required.
This results in more precise actions and improves the policy's success rate (\cref{tab:ablation}).

FMC is related in spirit to corrective imitation approaches such as DAgger-style residual correction~\cite{Ross2010ARO}, yet operates entirely within the denoising process and does not require expert relabeling. It is also related to consistency-style objectives, but differs in its goal: rather than accelerating sampling, FMC trains recovery from local action-space perturbations around the ground-truth trajectory. A hyperparameter sweep over the perturbation scale, drift strength, and correction threshold on a 2D robot-trajectory benchmark is given in~\cref{app:fmc_sweep}.

\subsection{Architecture and Training Details}
\label{sec:arch}

\parahead{Architecture}
\begin{wrapfigure}{r}{0.5\textwidth}
    \vspace{-0.4in}
  \centering
    \includegraphics[width=0.48\textwidth]{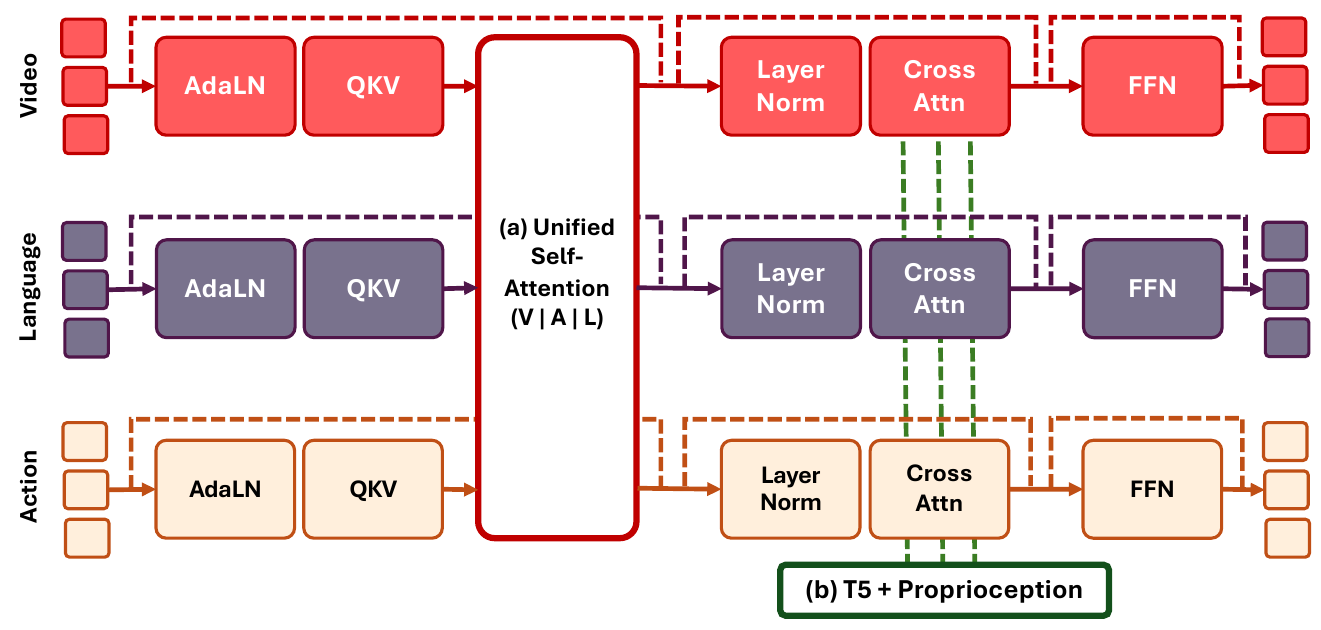}
  \caption{
    \textbf{Mixture of Transformers Layer}.
    Each modality is processed by its own transformer that performs adaptive layer norm followed by Q,K,V projection.
    The \textbf{(a) unified self-attention} enables feature sharing between different modalities.
    Each modality transformer then cross attends to the  \textbf{(b) task instruction and proprioception} for conditional generation of video, language, and action in a unified architecture.
  }
  \label{fig:arch_layer}
  \vspace{-0.2in}
\end{wrapfigure}

\coolname's architecture consists of a video prediction backbone~\cite{wan2025} and expands it to add action and language DiTs.
Our action and language DiT models are designed to match the video DiT and have $30$ layers each.
We reduce the size of the architecture by lowering the hidden dimension resulting in 1B action DiT (hidden dim $= 1024$) and $0.5$B language DiT (hidden dim $= 512$).
\cref{fig:arch_layer} displays one instance of the MoT layer.
Each modality in the MoT expert layer has its own adaptive layer norm for time conditioning followed by $Q$, $K$, $V$ projection.
These query, key, and value tokens from video, action, and language are concatenated along the token dimension to perform a unified self-attention.
We use RoPE position encoding on the query and key for unified self attention.
The unified self-attention is followed by a cross attention operation that attends video, action, and language (queries) to the task instruction and proprioception (keys).

\parahead{Implementation Details}
\label{method:implementation}
Our approach initializes from a Wan2.2 pretrained model and attaches an action and language DiT expert.
We fine-tune the video backbone with LoRA (rank 256)~\cite{LoRA}, restricting its update to a low-rank subspace.
This reduces trainable parameters and makes optimization easier and more stable on limited data.
During training, we randomly drop proprioception with a probability of 10\%, and use a single image as input.
We ensure that the reasoning tokens only attend to the visual features and input prompt and allow the action tokens to attend to the language tokens during generation (\cref{fig:arch_reasonvam}).
The model is trained with 8-bit AdamW optimizer~\cite{AdamW, 8bitOptimizers} on 12 H200 nodes with 8 gpus each for 44 hours.
We use an image resolution of 224 x 448 for simulator (two views) and 224 x 672 (three views) for real world experiments.
The language and action DiT are initialized from scratch.
We turn off video generation during inference, similar to FastWAM, and the language DiT runs synchronously with the action DiT.
Unoptimized \coolname without language generation runs at 76 ms and with the language expert runs at 123 ms to generate an action chunk of 32 with 10 denoising steps.
Our model allows interleaved language generation by caching language features from the last step to choose the frequency of language generation.

Full training configuration: action chunk $N = 32$, video horizon $H = 5$, 20 action and language denoising steps at inference, batch size $3840$, learning rate $4\mathrm{e}{-4}$, 15k steps on the 221K-trajectory MolmoSpaces mixture (the ablation in~\cref{tab:ablation} uses a $38K$ $Pick\&Place$ subset).
Our model generates 64 language tokens for each diffusion forward process.

\parahead{Data Flexibility} \coolname by construction has a modular architecture and training objective that supports co-training (by masking attention) on heterogeneous data (e.g., video-only, video-action, video-VQA, and video-action-VQA) within one representation.
We validate the design on in-domain video-action and reasoning data and leave large-scale multi-source co-training to future work.

\section{Experiments}
\label{sec:experiments}
\begin{figure*}[t]
  \centering
  \vspace{-0.1in}
  \includegraphics[width=\textwidth]{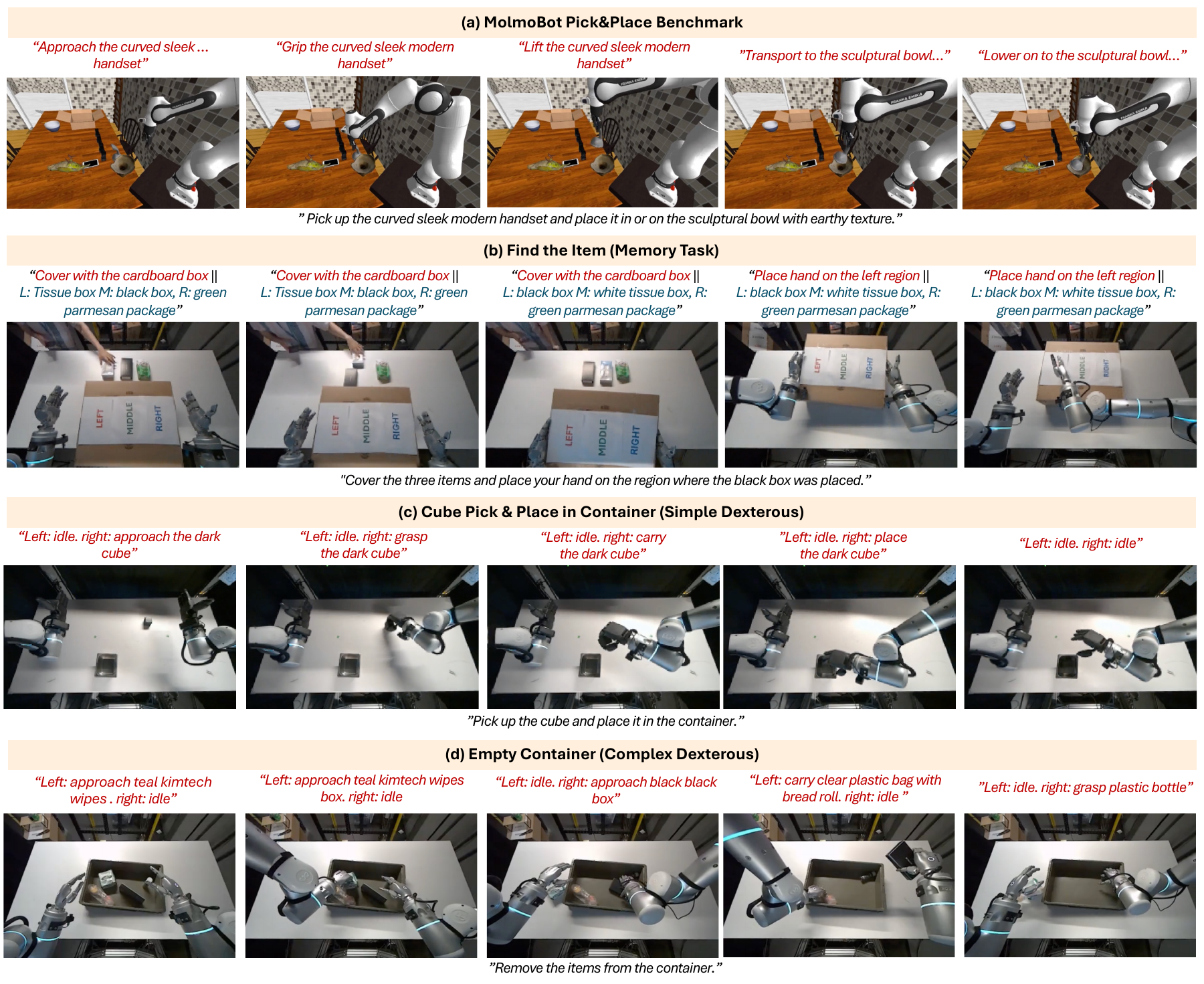}
  \vspace{-0.3in}
  \caption{
    \textbf{Qualitative Results of \coolname}. 
  \label{fig:qualitative_reasonwam}
    Our work can perform various tasks including: \textbf{(a) pick and place tasks, (b) find the item memory task, (c) simple high-dexterous task, and (d) complex dexterous bi-manual task involving difficult items to grasp}.
    For each demonstration, the top row displays the {\color{cb-maroon}subtasks} and {\color{cb-dark-blue} memory} predicted by the model, and the task instruction is shown below the images.
    Note, in \textbf{(b)} we make the task difficult by shifting items mid way during roll out for the policy to update its memory. 
    In (d) \coolname reasons about both the hands and retries to remove objects with varying geometries.
  }
  \vspace{-0.2in}
\end{figure*}

We evaluate \coolname{} along two axes: \textbf{Q1}: does jointly generating language enrich the shared representation of a video-action model and improve manipulation? \textbf{Q2}: can a policy use its own generated language as working memory on memory-dependent tasks?
The evaluation setup involves five simulation tasks on MolmoSpaces, the memory-dependent $M(n)$ subset of RMBench, and three real-world dexterous tasks (one of which requires memory).
\textbf{For fairness to baselines, we enable the anchor frame (visual memory) and the task-completion signal ($\beta$) only on RMBench $M(n)$}; on MolmoSpaces and all real-world experiments (including Find-the-Item) \coolname{} runs with neither mechanism, so the manipulation and real-world comparisons use the same inputs as the baselines and the entire memory signal on Find-the-Item comes from \coolname{}'s own re-consumed language.
\coolname{} and FastWAM~\cite{FastWAM} share the same Wan2.2 video backbone and action-expert size; \coolname{} adds language generation, FMC, LoRA finetuning, and (only for RMBench) the anchor frame.

\parahead{MolmoSpaces (sim, manipulation)} MolmoSpaces~\cite{MolmoSpaces} contains 221K trajectories across Pick, Pick-RandCam, Pick\&Place, Pick\&Place-NextTo, and Pick\&Place-Color on a DROID-style~\cite{droid_2024} Franka setup; we evaluate on 1000 OOD episodes per task using the Filament renderer, comparing against MolmoBot~\cite{Molmobot}, $\pi_{0.5}$~\cite{pi0.5}, and FastWAM~\cite{FastWAM}.

\parahead{RMBench $M(n)$ (sim, memory)} RMBench~\cite{rmbench} organizes memory-dependent manipulation by the number $n$ of task-relevant past observations required for optimal action; we evaluate on all four $M(n)$ tasks (Press Button, Blocks Ranking Try, Cover Blocks, Battery Try) on a dual-arm AgileX tabletop, with 50 demos / 100 rollout episodes per task, against same-backbone FastWAM, Mem-0~\cite{rmbench} (with and without privileged keyframes), and concurrent WLA-0~\cite{WLA0}.

\parahead{Real-World (dexterous, bi-manual)} Two Flexiv Rizon arms with dexterous hands (Sharpa Wave or Paxini) over three tasks (\cref{fig:qualitative_reasonwam}): \emph{Cube Pick \& Place} (simple, 44-DoF), \emph{Empty Container} (bi-manual, varied items), and \emph{Find the Item} (memory; items are covered mid-episode and queried). We evaluate on 20 held-out episodes per dexterous task and 18 for the memory task against $\pi_{0.5}$~\cite{pi0.5}, GROOT-N1.7~\cite{Nvidia2025GR00TNA}, FastWAM~\cite{FastWAM}, and MolmoBot~\cite{Molmobot}. Full environment and per-task details are deferred to \cref{app:eval}.

\parahead{Q1: Language generation improves manipulation} On MolmoSpaces (\cref{tab:molmobot}), \coolname{} surpasses MolmoBot on all Pick\&Place variants by \textbf{+7\%} on average without video history, despite using neither video history nor a large-scale pretrained vision-language generator (only the Wan2.2 video backbone is pretrained; language and action DiTs are trained from scratch). That \coolname{} still beats grounding- and language-pretrained baselines on compositional placement indicates the gains stem from the unified representation rather than inherited VLM priors.
FastWAM shares the same Wan2.2 backbone and action DiT structure, and is effectively the \emph{frame-level world model} ablation of our co-training objective (video + action, no task-level language stream); against it, \coolname{} improves success by \textbf{14.8\%} on average, isolating the benefit of adding task-level predictive structure.
\coolname{} underperforms on the two short-horizon grasp-only tasks (Pick, Pick-RandCam), which are bounded by a 303-step budget that penalizes our more deliberate trajectories; our gains grow with task compositionality (highest on Pick\&Place-NextTo and Pick\&Place-Color), indicating the unified representation contributes most where reasoning and placement precision, not raw reaching speed, are the bottleneck.
\begin{table*}[ht]
    \centering
    \caption{\textbf{Success rate on the MolmoB0t v2 benchmark} (1000 episodes per task, Filament renderer, 95\% confidence intervals). \colorbox{cb-cream-yellow}{\textbf{Best}} / \colorbox{silvercell}{\textbf{second-best}} per column.}
    \adjustbox{max width=\textwidth}{
    \begin{tabular}{l|c|ccccc}
    \toprule
     & \textbf{Pre-Training Data} & \textbf{Pick}{\color{blue}  $\uparrow$} & \textbf{Pick-RandCam} {\color{blue} $\uparrow$}  & \multicolumn{1}{c}{\textbf{Pick\&Place}} {\color{blue} $\uparrow$} & \multicolumn{1}{c}{\textbf{Pick\&Place-Color} {\color{blue} $\uparrow$}}  & \textbf{Pick\&Place-NextTo} {\color{blue} $\uparrow$}  \\
    \midrule
    \textbf{$\pi_{0.5}$-Finetuned}~\cite{pi0.5} & Grounding + Language & 
    25.8 $\pm$ 2.71 & 
    29.7 $\pm$ 2.83 & 
    43.5 $\pm$ 3.07 & 
    48.3 $\pm$ 3.10 & 
    28.7 $\pm$ 2.80 \\
    
    \textbf{Molmobot-Pi0}~\cite{Molmobot} & Grounding + Language & 
    33.3 $\pm$ 2.92& 
    39.8 $\pm$ 3.03& 
    46.6 $\pm$ 3.09 & 
    46.2 $\pm$ 3.09 & 
    24.5 $\pm$ 2.66 \\
    
    \textbf{FastWAM}~\cite{FastWAM} & Video & 
    30.1 $\pm$ 2.84 & 
    36.3 $\pm$ 2.98 & 
    51.1 $\pm$ 3.10 & 
    54.4 $\pm$ 3.09 & 
    \cellcolor{silvercell} \textbf{32.0} $\pm$ \textbf{2.89} \\
    
    \textbf{Molmobot-Img}~\cite{Molmobot} & Video + Grounding + Language  & 
    \cellcolor{silvercell} \textbf{61.4} $\pm$ \textbf{3.02} & 
    \cellcolor{silvercell} \textbf{62.1} $\pm$ \textbf{3.01} & 
    62.9 $\pm$ 2.99 & 
    \cellcolor{silvercell} \textbf{67.8} $\pm$ \textbf{2.89} & 
    21.0 $\pm$ 2.52 \\
    
    \textbf{Molmobot (With history)}~\cite{Molmobot} & Video + Grounding + Language & 
    \cellcolor{cb-cream-yellow} \textbf{64.0} $\pm$ \textbf{2.98} & 
    \cellcolor{cb-cream-yellow} \textbf{63.7} $\pm$ \textbf{2.98} & 
    \cellcolor{silvercell} \textbf{66.5} $\pm$ \textbf{2.92} & 
    \cellcolor{silvercell}\textbf{67.8} $\pm$ \textbf{2.89} & 
    26.4 $\pm$ 2.73 \\
    
    \textbf{\coolname (Ours)} & Video &  49.2 $\pm$ 3.10 & 51.0 $\pm$ 3.10 & \cellcolor{cb-cream-yellow} \textbf{68.0} $\pm$ \textbf{2.89} & \cellcolor{cb-cream-yellow} \textbf{70.8} $\pm$ \textbf{2.81}  & \cellcolor{cb-cream-yellow} \textbf{38.8} $\pm$ \textbf{3.02} \\
    \bottomrule
\end{tabular}}
    \label{tab:molmobot}
\end{table*}

\begin{table*}[h]
    \centering
    \caption{\textbf{Real-world evaluation.} Success rate (\%) on three dexterous bi-manual tasks — simple dexterity (Cube Pick\&Place), complex item removal (Empty Container), and the Find the Item memory task; 95\% CI [low, high]. 20 rollouts per dexterous task and 18 for the memory task (58 rollouts in total). \colorbox{cb-cream-yellow}{\textbf{Best}} / \colorbox{silvercell}{\textbf{second-best}} per column.}
    \adjustbox{max width=\textwidth}{
    \begin{tabular}{l|cccc}
    \toprule
     & \textbf{Cube Pick\&Place} & \textbf{Empty Container} & \textbf{Find the Item} & \textbf{Average}\\
    \midrule
     & \textbf{Dexterous - Simple} & \textbf{Dexterous - Complex} & \textbf{Memory} & \\
    \midrule
    
    \textbf{$\pi_{0.5}$}~\cite{pi0.5} &
    40.0 [20, 60] & 
    \cellcolor{silvercell} \textbf{20.0} \textbf{[0, 40]} &
    5.6 [0, 15.6] &
    21.9 [11.2, 32.5] 
    \\
    
    \textbf{Groot-N1.7}~\cite{Nvidia2025GR00TNA} & 
    \cellcolor{silvercell} \textbf{80.0} [60, 100] & 
    10.0 [0, 15] & 
    11.1 [0, 25.6]
    & 
    33.7 [21.5, 45.9] 
    \\

    \textbf{FastWAM}~\cite{FastWAM} & 
    70.0 [50, 90] & 
    \cellcolor{cb-cream-yellow} \textbf{50.0 [30, 70]} 
    & 
    11.1 [0, 25.6]
    &
    43.7 [30.9, 56.5] 
    \\

    \textbf{MolmoB0t}~\cite{Molmobot} & 
    \cellcolor{silvercell} \textbf{80.0} [60, 100] & 
    \cellcolor{silvercell} \textbf{20.0 [0, 40]} & 
    \cellcolor{silvercell} \textbf{33.3} \textbf{[11.6, 55]} 
    &
    \cellcolor{silvercell} \textbf{44.4 [31.7, 57.2]} 
    \\ 
    
    \textbf{\coolname (Ours)} & 
    \cellcolor{cb-cream-yellow} \textbf{95.0} \textbf{[85, 100]}& 
    \cellcolor{cb-cream-yellow} \textbf{50.0 [30, 70]} & 
    \cellcolor{cb-cream-yellow} \textbf{61.1} \textbf{[39.6, 82.6]} & 
    \cellcolor{cb-cream-yellow} \textbf{68.7 [56.8, 80.6]} 
    \\
    \bottomrule
\end{tabular}
}
\label{tab:real_evaluation}
\end{table*}

On the two non-memory real-world tasks (\cref{tab:real_evaluation}), \coolname{} reaches 95.0\% on Cube Pick\&Place (vs.\ 80.0\% for the next-best baseline) and ties FastWAM at 50.0\% on Empty Container. FastWAM matches us on Empty Container as that task requires no task-level reasoning, consistent with reasoning helping specifically where it is needed.
Per-component attribution of the manipulation gain is in \cref{sec:ablations}.

\begin{figure*}[h]
  \centering
  \setlength{\tabcolsep}{2pt}
  \renewcommand{\arraystretch}{1.05}
  \begin{tabular}{@{}c c c c c@{}}
    \noalign{\vspace{3ex}}
    \multicolumn{5}{@{}l@{}}{\textbf{Press Button (target: 4 left, 1 middle, then confirm)}}\\[0.5ex]
    \begin{minipage}[t]{0.19\textwidth}
      \centering
      \includegraphics[width=\linewidth]{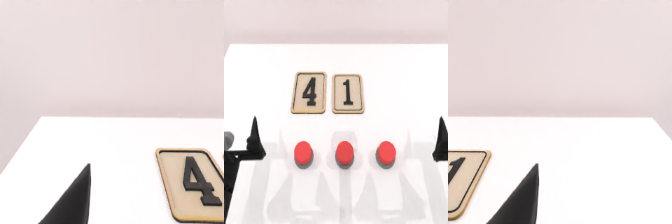}\\[0.3ex]
      \raggedright
      {\footnotesize\color{cb-maroon}\textbf{next subtask:}~left button.\par}
      {\footnotesize\color{cb-dark-blue}\textbf{memory:}~visible 4 1 \textbar{}\textbar{} left button 0/4 middle button 0/1 \textbar{}\textbar{} left button.\par}
    \end{minipage} &
    \begin{minipage}[t]{0.19\textwidth}
      \centering
      \includegraphics[width=\linewidth]{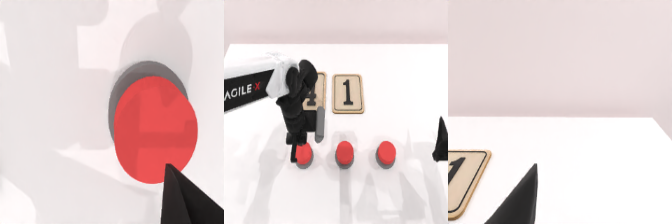}\\[0.3ex]
      \raggedright
      {\footnotesize\color{cb-maroon}\textbf{next subtask:}~left button.\par}
      {\footnotesize\color{cb-dark-blue}\textbf{memory:}~visible 4 1 \textbar{}\textbar{} left button 1/4 middle button 0/1 \textbar{}\textbar{} left button.\par}
    \end{minipage} &
    \begin{minipage}[t]{0.19\textwidth}
      \centering
      \includegraphics[width=\linewidth]{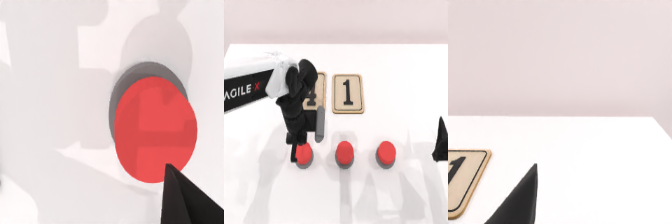}\\[0.3ex]
      \raggedright
      {\footnotesize\color{cb-maroon}\textbf{next subtask:}~left button.\par}
      {\footnotesize\color{cb-dark-blue}\textbf{memory:}~visible 4 1 \textbar{}\textbar{} left button 2/4 middle button 0/1 \textbar{}\textbar{} left button.\par}
    \end{minipage} &
    \begin{minipage}[t]{0.19\textwidth}
      \centering
      \includegraphics[width=\linewidth]{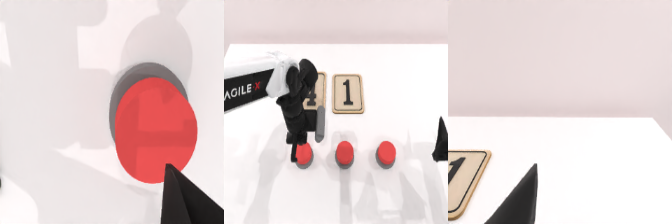}\\[0.3ex]
      \raggedright
      {\footnotesize\color{cb-maroon}\textbf{next subtask:}~left button.\par}
      {\footnotesize\color{cb-dark-blue}\textbf{memory:}~visible 4 1 \textbar{}\textbar{} left button 3/4 middle button 0/1 \textbar{}\textbar{} left button. next: middle button.\par}
    \end{minipage} &
    \begin{minipage}[t]{0.19\textwidth}
      \centering
      \includegraphics[width=\linewidth]{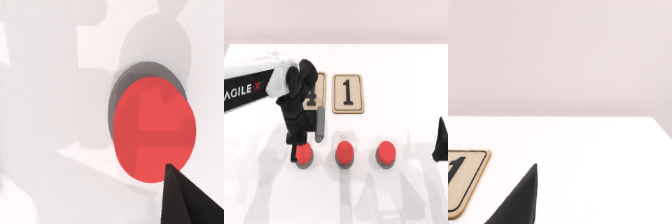}\\[0.3ex]
      \raggedright
      {\footnotesize\color{cb-maroon}\textbf{next subtask:}~middle button.\par}
      {\footnotesize\color{cb-dark-blue}\textbf{memory:}~visible 4 1 \textbar{}\textbar{} left button 4/4 middle button 0/1 \textbar{}\textbar{} middle button. next: confirm.\par}
    \end{minipage}\\
    \noalign{\vspace{3ex}}
    \multicolumn{5}{@{}l@{}}{\textbf{Battery Try (find working (battery, slot, orientation) pairs)}}\\[0.5ex]
    \begin{minipage}[t]{0.19\textwidth}
      \centering
      \includegraphics[width=\linewidth]{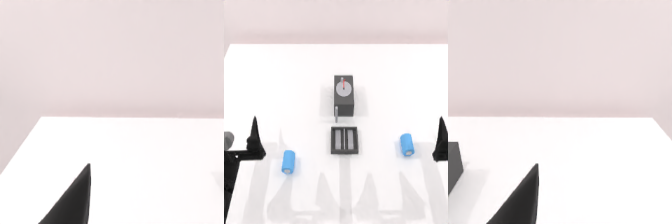}\\[0.3ex]
      \raggedright
      {\footnotesize\color{cb-maroon}\textbf{next subtask:}~dual positive.\par}
      {\footnotesize\color{cb-dark-blue}\textbf{memory:}~left none right none \textbar{}\textbar{} dual positive. next: left negative.\par}
    \end{minipage} &
    \begin{minipage}[t]{0.19\textwidth}
      \centering
      \includegraphics[width=\linewidth]{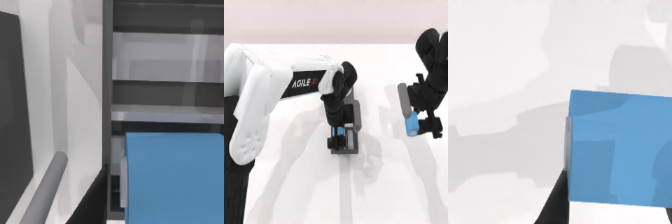}\\[0.3ex]
      \raggedright
      {\footnotesize\color{cb-maroon}\textbf{next subtask:}~dual positive.\par}
      {\footnotesize\color{cb-dark-blue}\textbf{memory:}~left none right none \textbar{}\textbar{} dual positive. next: left negative.\par}
    \end{minipage} &
    \begin{minipage}[t]{0.19\textwidth}
      \centering
      \includegraphics[width=\linewidth]{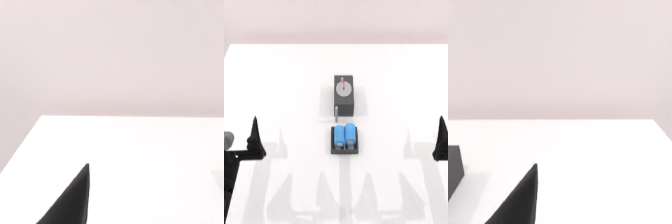}\\[0.3ex]
      \raggedright
      {\footnotesize\color{cb-maroon}\textbf{next subtask:}~left negative.\par}
      {\footnotesize\color{cb-dark-blue}\textbf{memory:}~left pos right pos \textbar{}\textbar{} left negative. next: done.\par}
    \end{minipage} &
    \begin{minipage}[t]{0.19\textwidth}
      \centering
      \includegraphics[width=\linewidth]{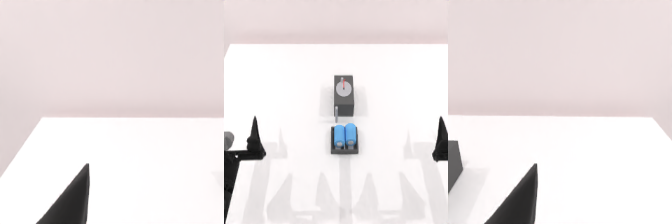}\\[0.3ex]
      \raggedright
      {\footnotesize\color{cb-maroon}\textbf{next subtask:}~right negative.\par}
      {\footnotesize\color{cb-dark-blue}\textbf{memory:}~left neg right pos \textbar{}\textbar{} right negative. next: left.\par}
    \end{minipage} &
    \begin{minipage}[t]{0.19\textwidth}
      \centering
      \includegraphics[width=\linewidth]{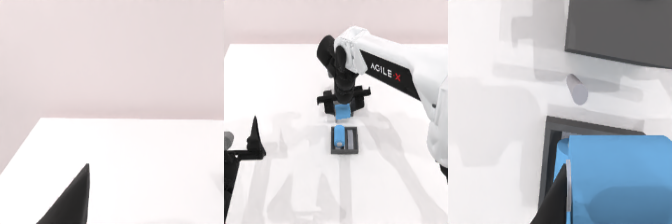}\\[0.3ex]
      \raggedright
      {\footnotesize\color{cb-maroon}\textbf{next subtask:}~right negative.\par}
      {\footnotesize\color{cb-dark-blue}\textbf{memory:}~left neg right pos \textbar{}\textbar{} right negative. next: done.\par}
    \end{minipage}\\
    \noalign{\vspace{3ex}}
    \multicolumn{5}{@{}l@{}}{\textbf{Blocks Ranking Try (find target permutation by trial and press to confirm)}}\\[0.5ex]
    \begin{minipage}[t]{0.19\textwidth}
      \centering
      \includegraphics[width=\linewidth]{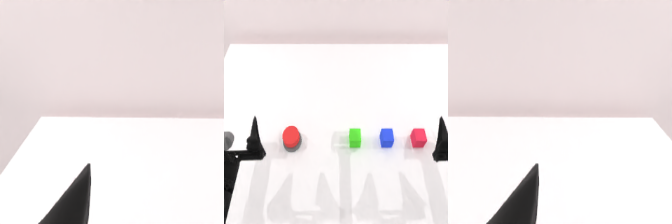}\\[0.3ex]
      \raggedright
      {\footnotesize\color{cb-maroon}\textbf{next subtask:}~press.\par}
      {\footnotesize\color{cb-dark-blue}\textbf{memory:}~press 0 swap 0 checking \textbar{}\textbar{} press. next: swap middle right.\par}
    \end{minipage} &
    \begin{minipage}[t]{0.19\textwidth}
      \centering
      \includegraphics[width=\linewidth]{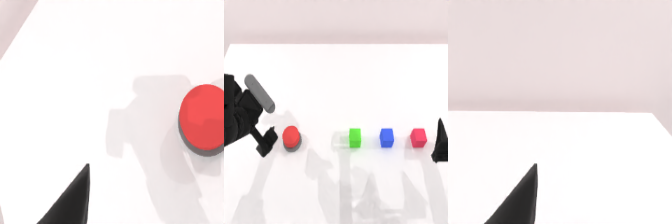}\\[0.3ex]
      \raggedright
      {\footnotesize\color{cb-maroon}\textbf{next subtask:}~swap middle right.\par}
      {\footnotesize\color{cb-dark-blue}\textbf{memory:}~press 1 swap 0 swapping \textbar{}\textbar{} swap middle right. next: press.\par}
    \end{minipage} &
    \begin{minipage}[t]{0.19\textwidth}
      \centering
      \includegraphics[width=\linewidth]{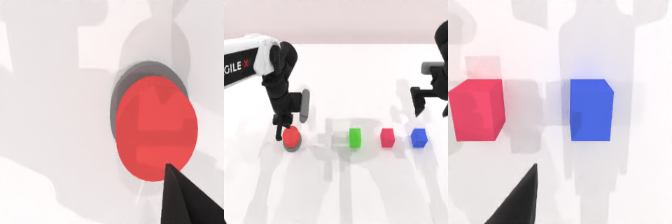}\\[0.3ex]
      \raggedright
      {\footnotesize\color{cb-maroon}\textbf{next subtask:}~press.\par}
      {\footnotesize\color{cb-dark-blue}\textbf{memory:}~press 1 swap 1 checking \textbar{}\textbar{} press. next: swap left right.\par}
    \end{minipage} &
    \begin{minipage}[t]{0.19\textwidth}
      \centering
      \includegraphics[width=\linewidth]{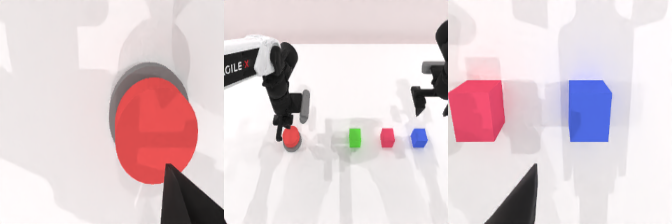}\\[0.3ex]
      \raggedright
      {\footnotesize\color{cb-maroon}\textbf{next subtask:}~swap left right.\par}
      {\footnotesize\color{cb-dark-blue}\textbf{memory:}~press 2 swap 1 swapping \textbar{}\textbar{} swap left right. next: press.\par}
    \end{minipage} &
    \begin{minipage}[t]{0.19\textwidth}
      \centering
      \includegraphics[width=\linewidth]{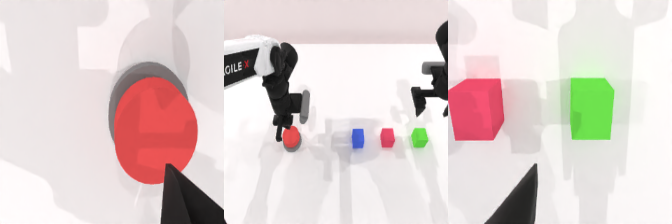}\\[0.3ex]
      \raggedright
      {\footnotesize\color{cb-maroon}\textbf{next subtask:}~press.\par}
      {\footnotesize\color{cb-dark-blue}\textbf{memory:}~press 2 swap 1 checking \textbar{}\textbar{} press. next: done.\par}
    \end{minipage}\\
    \noalign{\vspace{3ex}}
    \multicolumn{5}{@{}l@{}}{\textbf{Cover Blocks (cover blocks then uncover in red -- green -- blue order)}}\\[0.5ex]
    \begin{minipage}[t]{0.19\textwidth}
      \centering
      \includegraphics[width=\linewidth]{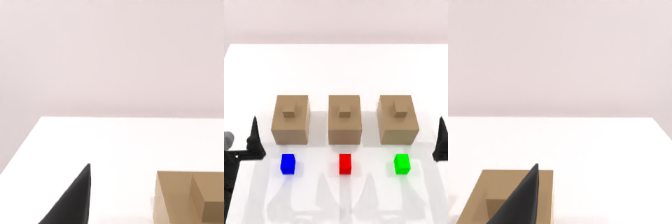}\\[0.3ex]
      \raggedright
      {\footnotesize\color{cb-maroon}\textbf{next subtask:}~cover left.\par}
      {\footnotesize\color{cb-dark-blue}\textbf{memory:}~covered none \textbar{}\textbar{} cover left. next: cover middle.\par}
    \end{minipage} &
    \begin{minipage}[t]{0.19\textwidth}
      \centering
      \includegraphics[width=\linewidth]{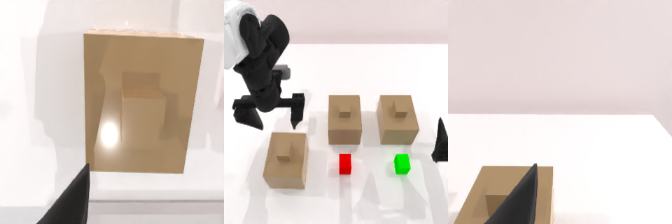}\\[0.3ex]
      \raggedright
      {\footnotesize\color{cb-maroon}\textbf{next subtask:}~cover middle.\par}
      {\footnotesize\color{cb-dark-blue}\textbf{memory:}~covered left \textbar{}\textbar{} cover middle. next: cover right.\par}
    \end{minipage} &
    \begin{minipage}[t]{0.19\textwidth}
      \centering
      \includegraphics[width=\linewidth]{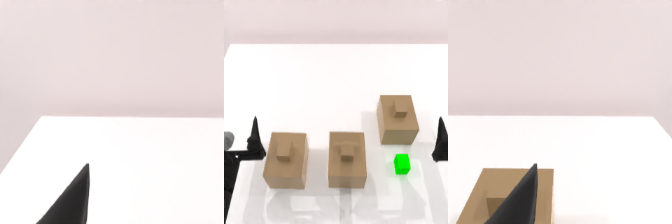}\\[0.3ex]
      \raggedright
      {\footnotesize\color{cb-maroon}\textbf{next subtask:}~cover right.\par}
      {\footnotesize\color{cb-dark-blue}\textbf{memory:}~covered left middle \textbar{}\textbar{} cover right. next: open middle.\par}
    \end{minipage} &
    \begin{minipage}[t]{0.19\textwidth}
      \centering
      \includegraphics[width=\linewidth]{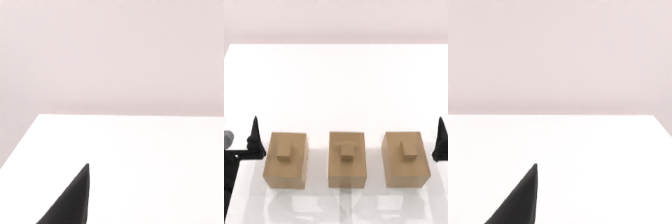}\\[0.3ex]
      \raggedright
      {\footnotesize\color{cb-maroon}\textbf{next subtask:}~open middle.\par}
      {\footnotesize\color{cb-dark-blue}\textbf{memory:}~covered left middle right \textbar{}\textbar{} open middle. next: open right.\par}
    \end{minipage} &
    \begin{minipage}[t]{0.19\textwidth}
      \centering
      \includegraphics[width=\linewidth]{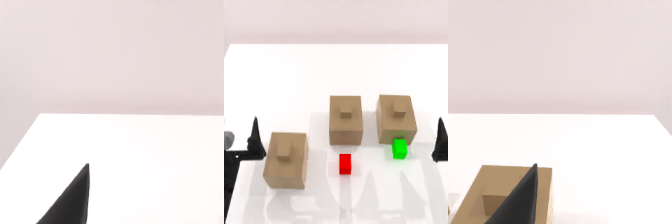}\\[0.3ex]
      \raggedright
      {\footnotesize\color{cb-maroon}\textbf{next subtask:}~open left.\par}
      {\footnotesize\color{cb-dark-blue}\textbf{memory:}~covered left \textbar{}\textbar{} open left. next: done.\par}
    \end{minipage}\\
  \end{tabular}
  \caption{\textbf{RMBench $M(n)$ qualitative rollouts.} Each row is a successful \coolname{} rollout on one $M(n)$ task; we display a subsampled sequence of inference steps for clarity. Below each panel we show the predicted \textcolor{cb-maroon}{next subtask} and the generated \textcolor{cb-dark-blue}{textual memory} that is re-consumed as the prompt on the next control step; within the memory string, ``\texttt{\textbar\textbar}'' separates memory components (task state, progress counters, and the next subtask). Reading left-to-right shows how the language-memory loop accumulates task-relevant state (button counts, battery orientations, tried permutations, cover progress) that memoryless baselines lose across steps.}
  \label{fig:qualitative_rmbench}
\end{figure*}

\begin{table*}[h]
    \centering
    \caption{\textbf{RMBench evaluation on the $M(n)$ memory subset.} Success rate (\%) per task on the four $M(n)$ RMBench~\cite{rmbench} tasks. Baselines and Mem-0 w/o Key operate without keyframes; \coolname{} uses a single initial anchor frame plus a generated textual memory. \colorbox{cb-cream-yellow}{\textbf{Best}} / \colorbox{silvercell}{\textbf{second-best}} per column in the without-keyframes block.}
    \adjustbox{max width=\textwidth}{
    \begin{tabular}{l|cccc|c}
    \toprule
    \textbf{Method} & \textbf{Battery Try} & \textbf{Blocks Ranking Try} & \textbf{Cover Blocks} & \textbf{Press Button} & \textbf{Average} \\
    \midrule
    \multicolumn{6}{l}{\emph{Without keyframes}} \\
    \midrule
    DP~\cite{diffusion_policy}            & 10\% & 10\% & 0\%  & 0\%  & 5.0\%  \\
    ACT~\cite{zhao2023act}                & 19\% & 0\%  & 0\%  & 0\%  & 4.8\%  \\
    $\pi_{0.5}$~\cite{pi0.5}              & 16\% & 6\%  & 0\%  & 0\%  & 5.5\%  \\
    X-VLA~\cite{Zheng2025XVLAST}          & \cellcolor{cb-cream-yellow}\textbf{26\%} & 1\%  & 2\%  & 0\%  & 7.3\%  \\
    Mem-0~\cite{rmbench} (w/o Key)        & 13\% & 1\%  & \cellcolor{silvercell}\textbf{5\%}  & 0\%  & 4.8\%  \\
    FastWAM~\cite{FastWAM}                & 16\% & \cellcolor{silvercell}\textbf{37\%} & 0\%  & 0\%  & \cellcolor{silvercell}\textbf{13.3\%} \\
    \textbf{\coolname{} (Ours)}           & \cellcolor{silvercell}\textbf{25\%} & \cellcolor{cb-cream-yellow}\textbf{49\%} & \cellcolor{cb-cream-yellow}\textbf{13\%} & \cellcolor{cb-cream-yellow}\textbf{51\%} & \cellcolor{cb-cream-yellow}\textbf{34.5\% \textcolor{green!60!black}{(+21.2\%)}} \\
    \midrule
    \multicolumn{6}{l}{\emph{With keyframes}} \\
    \midrule
    Mem-0~\cite{rmbench}                  & 28\% & 18\% & 68\% & 0\%  & 28.5\% \\
    WLA-0~\cite{WLA0}                     & 45\% & 23\% & 84\% & 74\% & 56.5\% \\
    \bottomrule
\end{tabular}
}
\label{tab:rmbench}
\vspace{-0.1in}
\end{table*}

\parahead{Q2: Generated language serves as working memory} On RMBench's memory-dependent $M(n)$ subset (\cref{tab:rmbench}, qualitative rollouts in \cref{fig:qualitative_rmbench}), \coolname{} succeeds \textbf{34.5\%} on average \emph{without any pretrained language or reasoning model}, versus 13.3\% for same-backbone FastWAM and 5--13\% for other memoryless baselines (Mem-0 with its key memory ablated collapses to 4.8\%); it also exceeds Mem-0's keyframe-equipped setup (28.5\%) using only a single anchor frame rather than per-task privileged keyframes. On RMBench we match the Mem-0 and WLA-0 input setup (visual memory + task-completion channel + textual memory), while using less information: no history keyframes. FastWAM shares our backbone and action-expert size but has no language stream; same-backbone FastWAM reaches only 13.3\%, so the \textbf{+21.2-point RMBench gap} is attributable to the language-memory loop and the anchor-frame visual memory, with per-mechanism attribution in \cref{tab:rmbench_ablation}. Concurrent WLA-0~\cite{WLA0} scores higher but is a pipeline of a pretrained VLM plus a pretrained image-generation model linked via meta-queries; it inherits web-scale language and image-generation priors that \coolname{} does not use, and does not form a unified representation. On the real-world Find-the-Item task (\cref{tab:real_evaluation}, rollouts in \cref{fig:qualitative_real_memory}), we use \emph{textual memory only} (no anchor frame, no task-completion channel) for \emph{fairness to the baselines} ($\pi_{0.5}$, GR00T-N1.7, FastWAM, MolmoBot), which have no anchor-frame mechanism; the entire memory signal therefore comes from \coolname{}'s own re-consumed language string. Objects are covered mid-episode, so any policy that re-plans from the current frame loses their identities; \coolname{}'s re-consumed textual memory retains them across steps, raising success from \textbf{11.1\%} (FastWAM, same video backbone, no language) to \textbf{61.1\%}, while VLA and WAM baselines stay at or below chance ($\leq 33.3\%$). This gap indicates that generating language reasoning, not action quality alone, drives memory-dependent performance. Per-mechanism attribution of the memory gain is in \cref{sec:ablations}; a Press Button failure case is in \cref{app:eval}.

\subsection{Ablations}
\label{sec:ablations}

\parahead{Manipulation (Pick\&Place)}
\cref{tab:ablation} ablates each module on manipulation. For each row we train on a single Pick\&Place split (38K trajectories) and evaluate on the MolmoSpaces Pick\&Place OOD test set (1000 episodes, Filament renderer). 
Language generation (\cref{method:reasoning}) improves OOD success by 4.9 points (15.8\% relative); Flow Correction (\cref{method:fc}) adds 3.0 points (9.7\% relative), producing more precise actions; LoRA (\cref{method:implementation}) improves over full fine-tuning while drastically reducing trainable parameters.

\begin{wraptable}{r}{0.55\textwidth}
    \centering
    \vspace{-0.2in}
    \caption{Ablation on Pick\&Place task. We ablate design choices of our policy based on \textbf{Reasoning (R), Flow Matching with Correction (FMC)}, and \textbf{LoRA}. \colorbox{cb-cream-yellow}{\textbf{Best}} / \colorbox{silvercell}{\textbf{second-best}}}
    \adjustbox{max width=0.53\textwidth}{
    \begin{tabular}{l|rrr}
    \toprule
     & Success Rate (\%)\\
    \midrule
    \textbf{\coolname (w/o R w/o FMC w/o LoRA) $\simeq$ FastWAM} & 31.00 $\pm$ 2.87\\
    \textbf{\coolname (R (\ref{method:reasoning}))} & 35.90 $\pm$ 2.97\\
    \textbf{\coolname (R (\ref{method:reasoning}) + FMC (\ref{method:fc}))} & \cellcolor{silvercell} \textbf{38.90} $\pm$ \textbf{3.02} \\
    \textbf{\coolname (Ours) (R (\ref{method:reasoning}) + FMC (\ref{method:fc}) + LoRA (\ref{method:implementation}))} & \cellcolor{cb-cream-yellow} \textbf{41.00} $\pm$ \textbf{3.04} \\
    \bottomrule
\end{tabular}
}
\label{tab:ablation}
\vspace{-0.1in}
\end{wraptable}
Overall, \coolname{} improves the base WAM from 31.0\% to 41.0\% (+10 points, 32\% relative) on this split, and circa 14.8 points (37\% relative) on average across the full benchmark over FastWAM. These gains come from merely adding a 0.5B-parameter language expert.
The language decoding paradigm (AR vs.~diffusion) is orthogonal to our contribution: these ablations isolate the effect of language co-training within a shared diffusion representation, not the decoding paradigm.

\begin{wraptable}{r}{0.65\textwidth}
    \centering
    \vspace{-0.1in}
    \caption{\textbf{RMBench Memory Ablation on  Press Button Task} (\cref{subsec:memory}). Effect of language generation, visual memory (anchor frame), and training length. \coolname{} without both language generation and visual memory is architecturally identical to FastWAM ($\simeq$FastWAM). Visual memory alone (no language generation) reaches only 1\%, while language generation alone (no visual memory) reaches 15\% at 30k --- language generation carries most of the memory signal. Combining both (30k) gives 18\%, and continuing training to 100k reaches 51\%.}
    \vspace{0.05in}
    \adjustbox{max width=0.64\textwidth}{
    \begin{tabular}{l|c|c}
    \toprule
    \textbf{Method} & \textbf{Steps} & \textbf{Press Button} \\
    \midrule
    \textbf{\coolname{} w/o language generation, w/o visual memory ($\simeq$FastWAM)~\cite{FastWAM}} & 100k & 0\%  \\
    \textbf{\coolname{} w/o language generation (textual memory (\ref{para:textual_memory}))}                                                    & 30k  & 1\%  \\
    \textbf{\coolname{} w/o visual memory (\ref{para:visual_memory})}                                                          & 30k  & 15\% \\
    \textbf{\coolname{} (full)}                                                                     & 30k  & 18\% \\
    \textbf{\coolname{} (full)}                                                                     & 100k & \cellcolor{cb-cream-yellow}\textbf{51\%} \\
    \bottomrule
    \end{tabular}
    }
    \label{tab:rmbench_ablation}
    \vspace{-0.1in}
\end{wraptable}

\parahead{Memory (Press Button)}
A per-mechanism ablation on RMBench \emph{Press Button} (\cref{tab:rmbench_ablation}) confirms language generation carries the memory signal. Under a limited training budget, visual memory alone (anchor frame only, no language generation) reaches only 1\%, while language generation alone (no anchor frame) reaches 15\%; the fully trained \coolname{} with both visual memory and language generation reaches 51\%. Both mechanisms are needed, with generated language carrying most of the signal: it encodes the symbolic state the policy needs to decide whether to press a given button, while the visual anchor stabilizes which buttons remain targets as frames advance.

\begin{figure*}[t]
  \centering
  \setlength{\tabcolsep}{2pt}
  \renewcommand{\arraystretch}{1.05}
  \begin{tabular}{@{}c c c c c c@{}}
    \multicolumn{6}{@{}l@{}}{\textbf{Press Button, target: 3 left, 4 middle, then confirm.}}\\[0.3ex]
    \begin{minipage}[t]{0.165\textwidth}
      \centering
      \includegraphics[width=\linewidth]{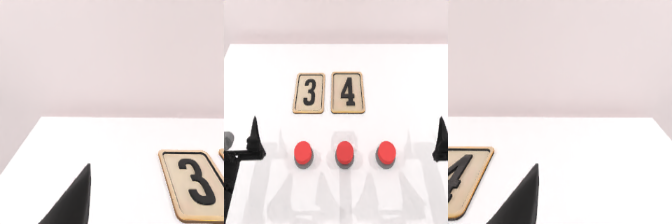}\\[0.3ex]
      \raggedright
      {\footnotesize\color{cb-maroon}\textbf{next subtask:}~left button.\par}
      {\footnotesize\color{cb-dark-blue}\textbf{memory:}~visible 3 4 \textbar{}\textbar{} left 0/3 mid 0/4 \textbar{}\textbar{} left button.\par}
    \end{minipage} &
    \begin{minipage}[t]{0.165\textwidth}
      \centering
      \includegraphics[width=\linewidth]{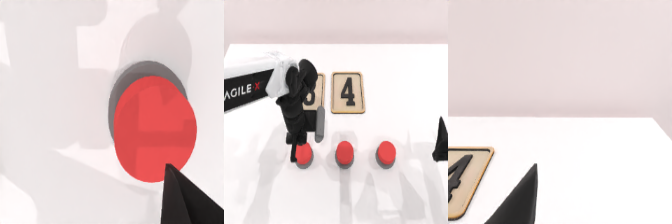}\\[0.3ex]
      \raggedright
      {\footnotesize\color{cb-maroon}\textbf{next subtask:}~left button.\par}
      {\footnotesize\color{cb-dark-blue}\textbf{memory:}~visible 3 4 \textbar{}\textbar{} left 2/3 mid 0/4 \textbar{}\textbar{} left button. next: middle.\par}
    \end{minipage} &
    \begin{minipage}[t]{0.165\textwidth}
      \centering
      \includegraphics[width=\linewidth]{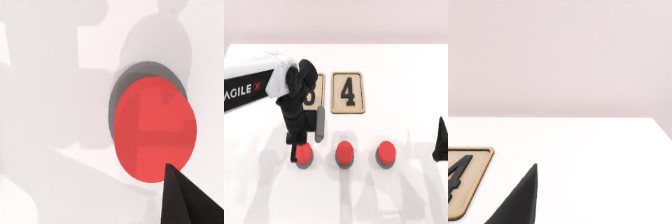}\\[0.3ex]
      \raggedright
      {\footnotesize\color{cb-maroon}\textbf{next subtask:}~middle button.\par}
      {\footnotesize\color{cb-dark-blue}\textbf{memory:}~visible 3 4 \textbar{}\textbar{} left 3/3 mid 0/4 \textbar{}\textbar{} middle button.\par}
    \end{minipage} &
    \begin{minipage}[t]{0.165\textwidth}
      \centering
      \includegraphics[width=\linewidth]{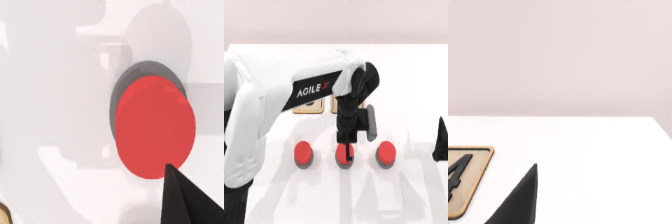}\\[0.3ex]
      \raggedright
      {\footnotesize\color{cb-maroon}\textbf{next subtask:}~middle button.\par}
      {\footnotesize\color{cb-dark-blue}\textbf{memory:}~visible 3 4 \textbar{}\textbar{} left 3/3 mid 1/4 \textbar{}\textbar{} middle button.\par}
    \end{minipage} &
    \begin{minipage}[t]{0.165\textwidth}
      \centering
      \includegraphics[width=\linewidth]{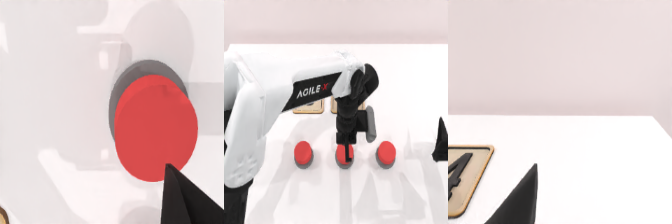}\\[0.3ex]
      \raggedright
      {\footnotesize\color{cb-maroon}\textbf{next subtask:}~middle button.\par}
      {\footnotesize\color{cb-dark-blue}\textbf{memory:}~visible 3 4 \textbar{}\textbar{} left 3/3 mid 3/4 \textbar{}\textbar{} middle button. next: confirm.\par}
    \end{minipage} &
    \begin{minipage}[t]{0.165\textwidth}
      \centering
      \includegraphics[width=\linewidth]{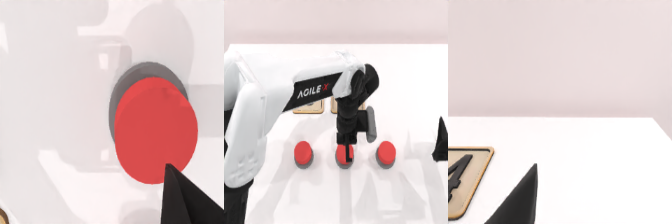}\\[0.3ex]
      \raggedright
      {\footnotesize\color{cb-maroon}\textbf{next subtask:}~confirm.\par}
      {\footnotesize\color{cb-dark-blue}\textbf{memory:}~visible 3 4 \textbar{}\textbar{} left 3/3 mid 4/4 \textbar{}\textbar{} confirm. next: done.\par}
    \end{minipage}\\
  \end{tabular}
  \caption{\textbf{Language-as-memory loop on Press Button.} At each step \coolname{} emits a \textcolor{cb-maroon}{next subtask} and \textcolor{cb-dark-blue}{textual memory} that is re-consumed as the next-step prompt; press counters transition cleanly (left $0/3\to3/3$; middle $1/4\to4/4$) until confirm. Intermediate frames skipped for clarity; see \cref{fig:qualitative_rmbench} for all four $M(n)$ tasks.}
  \label{fig:qualitative_teaser}
  \vspace{-0.2in}
\end{figure*}

\parahead{Qualitative Rollouts} We provide per-task rollout galleries that visualize how \coolname{}'s predicted {\color{cb-maroon}subtasks} and {\color{cb-dark-blue}textual memory} evolve across time: RMBench $M(n)$ (\cref{fig:qualitative_rmbench}), MolmoSpaces Pick \& Place (\cref{fig:qualitative_sim_tasks,fig:qualitative_sim_tasks_b}), real-world Empty Container (\cref{fig:qualitative_real_dexterous}), and real-world \emph{Find the Item} (\cref{fig:qualitative_real_memory}).

\parahead{Conclusion} We present \coolname{}, a single-representation \walm that jointly denoises video, action, and language, and re-consumes its generated language as working memory for the next control step.
Without any pretrained language or reasoning model, \coolname{} achieves state-of-the-art on MolmoSpaces Pick\&Place (+14.8 pts over FastWAM), lifts real-world Find-the-Item from 11.1\% to 61.1\%, and reaches 34.5\% on RMBench $M(n)$, exceeding both memoryless baselines and Mem-0~\cite{rmbench}'s keyframe-equipped setup while using only a single anchor frame.

\parahead{Limitations} Failures include grasping complex geometries and cluttered scenes. On RMBench we observe cases where the subtask and textual memory correctly track state but the action expert fails to execute it (e.g., subtask-boundary over/undershoot from task-completion timing lag; \cref{fig:failure_press_button}), and near-identical observations with different underlying memory occasionally confuse the model; the full list of failure modes is in \cref{sec:limitations}.

\parahead{Future Work} We aim to scale multimodal co-training (ego-video, VQA, grounding), extend FMC to learn from the model's own rollout errors, and extend visual memory to multiple keyframes; see \cref{sec:future_work} for extended directions.


\bibliography{example}  

@article{OminiControl,
  title={OminiControl: Minimal and Universal Control for Diffusion Transformer},
  author={Zhenxiong Tan and Song-Hua Liu and Xingyi Yang and Qiaochu Xue and Xinchao Wang},
  journal={2025 IEEE/CVF International Conference on Computer Vision (ICCV)},
  year={2024},
  pages={14940-14950}
  }

@inproceedings{diffusiongemma,
  title={{DiffusionGemma Technical Report}},
  author={{DiffusionGemma Team} Adrien Ali Ta{\"i}ga and James Assiene and Daniele Calandriello and Rahma Chaabouni and Jo{\~a}o Gante and Tamara von Glehn and Nate Keating and Chris Knutsen and Martin Kukla and Tianlin Liu and Ivan Lobov and Ofir Nabati and Jo{\~a}o Gabriel Oliveira and Nicolas Perez-Nieves and Nastasia Prutianova and Bobak Shahriari and Jean Tarbouriech and Pavel Tyletski and Çağlar {\"U}nl{\"u} and Cindy Wu and Glenn Cameron and Jerome T. Connor and Sertan Girgin and Maarten R. Grootendorst and Alon Levkovitch and Eliya Nachmani and Omar Sanseviero and Piotr Stańczyk and Quentin Berthet and Andrew Campbell and Cl{\'e}-ment Crepy and Valentin De Bortoli and Arnaud Doucet and Romuald {\'E}lie and Alexandre Galashov and Klaus Greff and Alexis Jacq and David Ruhe and Yu-Han Wu and Sebastian Flennerhag and Brendan O'Donoghue and George Scrivener and Shantanu Thakoor},
  year={2026},
  journal={arxiv:2608.00146}
}

@article{FastWAM,
  title={{Fast-WAM: Do World Action Models Need Test-time Future Imagination?}},
  author={Tianyuan Yuan and Zibin Dong and Yicheng Liu and Hang Zhao},
  journal={arxiv:2603.16666},
  year={2026},
}

@article{MolmoSpaces,
  title={MolmoSpaces: A Large-Scale Open Ecosystem for Robot Navigation and Manipulation},
  author={Yejin Kim and Wilbert Pumacay and Omar Rayyan and Max Argus and Winson Han and Eli VanderBilt and Jordi Salvador and Abhay Deshpande and Rose Hendrix and Snehal Jauhri and Shuo Liu and Nur Muhammad Shafiullah and Maya Guru and Ainaz Eftekhar and Karen Farley and Donovan Clay and Jiafei Duan and A A Guru and Piper Wolters and Alvaro Herrasti and Ying-Chun Lee and Georgia Chalvatzaki and Yuchen Cui and Ali Farhadi and Dieter Fox and Ranjay Krishna},
  journal={arxiv:2602.11337},
  year={2026},
}

@article{Molmobot,
  title={{MolmoB0T: Large-Scale Simulation Enables Zero-Shot Manipulation}},
  author={Abhay Deshpande and Maya Guru and Rose Hendrix and Snehal Jauhri and Haoquan Fang and Wilbert Pumacay and Yejin Kim and Quinn Pfeifer and Ying-Chun Lee and Piper Wolters and Omar Rayyan and Mingtong Zhang and Karen Farley and Winson Han and Eli VanderBilt and Dieter Fox and Ali Farhadi and Georgia Chalvatzaki and Dhruv Shah and Ranjay Krishna},
  journal={ArXiv},
  year={2026},
  volume={abs/2603.16861},
}

@article{LoRA,
  title={{LoRA: Low-Rank Adaptation of Large Language Models}},
  author={Edward J. Hu and Yelong Shen and Phillip Wallis and Zeyuan Allen-Zhu and Yuanzhi Li and Shean Wang and Weizhu Chen},
  journal={arxiv:2106.09685},
  year={2021},
}

@article{AdamW,
  title={{Fixing Weight Decay Regularization in Adam}},
  author={Ilya Loshchilov and Frank Hutter},
  journal={arxiv:1711.05101},
  year={2017},
}

@article{8bitOptimizers,
  title={8-bit Optimizers via Block-wise Quantization},
  author={Tim Dettmers and Mike Lewis and Sam Shleifer and Luke Zettlemoyer},
  journal={arxiv:2110.02861},
  year={2021},
}

@article{Motus,
  title={Motus: A Unified Latent Action World Model},
  author={Hong-Zhe Bi and Hengkai Tan and Shenghao Xie and Zeyu Wang and Shuhe Huang and Hai-Tian Liu and Ruowen Zhao and Yao Feng and Chendong Xiang and Yinze Rong and Hongyang Zhao and Han-Yu Liu and Zhi-Zhong Su and Lei Ma and Hang Su and Jun Zhu},
  journal={ArXiv},
  year={2025},
  volume={abs/2512.13030},
}

@article{Nvidia2025GR00TNA,
  title={GR00T N1: An Open Foundation Model for Generalist Humanoid Robots},
  author={Nvidia and Johan Bjorck and Fernando Casta{\~n}eda and Nikita Cherniadev and Xingye Da and Runyu Ding and LinxiJimFan and Yu Fang and Dieter Fox and Fengyuan Hu and Spencer Huang and Joel Jang and Zhenyuan Jiang and Jan Kautz and Kaushil Kundalia and Lawrence Lao and Zhiqi Li and Zongyu Lin and Kevin Lin and Guilin Liu and Edith Llontop and Loic Magne and Ajay Mandlekar and Avnish Narayan and Soroush Nasiriany and Scott Reed and You Liang Tan and Guanzhi Wang and Zu Wang and Jing Wang and Qi Wang and Jiannan Xiang and Yuqi Xie and Yinzhen Xu and Zhen-Teng Xu and Seonghyeon Ye and Zhiding Yu and Ao Zhang and Hao Zhang and Yizhou Zhao and Ruijie Zheng and Yuke Zhu},
  journal={ArXiv},
  year={2025},
  volume={abs/2503.14734},
}

@article{pi0.5,
  title={$\pi$0.5: a Vision-Language-Action Model with Open-World Generalization},
  author={{Physical Intelligence} and Kevin Black and Noah Brown and James Darpinian and Karan Dhabalia and Danny Driess and Adnan Esmail and Michael Equi and Chelsea Finn and Niccolo Fusai and Manuel Y. Galliker and Dibya Ghosh and Lachy Groom and Karol Hausman and Brian Ichter and Szymon Jakubczak and Tim Jones and Liyiming Ke and Devin LeBlanc and Sergey Levine and Adrian Li-Bell and Mohith Mothukuri and Suraj Nair and Karl Pertsch and Allen Z. Ren and Lucy Xiaoyang Shi and Laura Smith and Jost Tobias Springenberg and Kyle Stachowicz and James Tanner and Quan Vuong and Homer Rich Walke and Anna Walling and Hao-Huan Wang and Lili Yu and Ury Zhilinsky},
  journal={arxiv:2504.16054},
  year={2025},
}

@article{droid_2024,
    title={{DROID: A Large-Scale In-The-Wild Robot Manipulation Dataset}},
    author  = {Alexander Khazatsky and Karl Pertsch and Suraj Nair and Ashwin Balakrishna and Sudeep Dasari and Siddharth Karamcheti and Soroush Nasiriany and Mohan Kumar Srirama and Lawrence Yunliang Chen and Kirsty Ellis and Peter David Fagan and Joey Hejna and Masha Itkina and Marion Lepert and Yecheng Jason Ma and Patrick Tree Miller and Jimmy Wu and Suneel Belkhale and Shivin Dass and Huy Ha and Arhan Jain and Abraham Lee and Youngwoon Lee and Marius Memmel and Sungjae Park and Ilija Radosavovic and Kaiyuan Wang and Albert Zhan and Kevin Black and Cheng Chi and Kyle Beltran Hatch and Shan Lin and Jingpei Lu and Jean Mercat and Abdul Rehman and Pannag R Sanketi and Archit Sharma and Cody Simpson and Quan Vuong and Homer Rich Walke and Blake Wulfe and Ted Xiao and Jonathan Heewon Yang and Arefeh Yavary and Tony Z. Zhao and Christopher Agia and Rohan Baijal and Mateo Guaman Castro and Daphne Chen and Qiuyu Chen and Trinity Chung and Jaimyn Drake and Ethan Paul Foster and Jensen Gao and Vitor Guizilini and David Antonio Herrera and Minho Heo and Kyle Hsu and Jiaheng Hu and Muhammad Zubair Irshad and Donovon Jackson and Charlotte Le and Yunshuang Li and Kevin Lin and Roy Lin and Zehan Ma and Abhiram Maddukuri and Suvir Mirchandani and Daniel Morton and Tony Nguyen and Abigail O'Neill and Rosario Scalise and Derick Seale and Victor Son and Stephen Tian and Emi Tran and Andrew E. Wang and Yilin Wu and Annie Xie and Jingyun Yang and Patrick Yin and Yunchu Zhang and Osbert Bastani and Glen Berseth and Jeannette Bohg and Ken Goldberg and Abhinav Gupta and Abhishek Gupta and Dinesh Jayaraman and Joseph J Lim and Jitendra Malik and Roberto Martín-Martín and Subramanian Ramamoorthy and Dorsa Sadigh and Shuran Song and Jiajun Wu and Michael C. Yip and Yuke Zhu and Thomas Kollar and Sergey Levine and Chelsea Finn},
    journal={arxiv:2403.12945v1},
    year = {2024},
}

@article{dreamzero,
      title={{World Action Models are Zero-shot Policies}}, 
      author={Seonghyeon Ye and Yunhao Ge and Kaiyuan Zheng and Shenyuan Gao and Sihyun Yu and George Kurian and Suneel Indupuru and You Liang Tan and Chuning Zhu and Jiannan Xiang and Ayaan Malik and Kyungmin Lee and William Liang and Nadun Ranawaka and Jiasheng Gu and Yinzhen Xu and Guanzhi Wang and Fengyuan Hu and Avnish Narayan and Johan Bjorck and Jing Wang and Gwanghyun Kim and Dantong Niu and Ruijie Zheng and Yuqi Xie and Jimmy Wu and Qi Wang and Ryan Julian and Danfei Xu and Yilun Du and Yevgen Chebotar and Scott Reed and Jan Kautz and Yuke Zhu and Linxi "Jim" Fan and Joel Jang},
      year={2026},
      journal={arxiv:2602.15922},
      archivePrefix={arXiv},
      primaryClass={cs.RO}, 
}

@article{cosmospolicy,
      title={Cosmos Policy: Fine-Tuning Video Models for Visuomotor Control and Planning}, 
      author={Moo Jin Kim and Yihuai Gao and Tsung-Yi Lin and Yen-Chen Lin and Yunhao Ge and Grace Lam and Percy Liang and Shuran Song and Ming-Yu Liu and Chelsea Finn and Jinwei Gu},
      year={2026},
      journal={arxiv:2601.16163},
      archivePrefix={arXiv},
      primaryClass={cs.AI},
}

@article{wan2025,
      title={Wan: Open and Advanced Large-Scale Video Generative Models}, 
      author={Team Wan and Ang Wang and Baole Ai and Bin Wen and Chaojie Mao and Chen-Wei Xie and Di Chen and Feiwu Yu and Haiming Zhao and Jianxiao Yang and Jianyuan Zeng and Jiayu Wang and Jingfeng Zhang and Jingren Zhou and Jinkai Wang and Jixuan Chen and Kai Zhu and Kang Zhao and Keyu Yan and Lianghua Huang and Mengyang Feng and Ningyi Zhang and Pandeng Li and Pingyu Wu and Ruihang Chu and Ruili Feng and Shiwei Zhang and Siyang Sun and Tao Fang and Tianxing Wang and Tianyi Gui and Tingyu Weng and Tong Shen and Wei Lin and Wei Wang and Wei Wang and Wenmeng Zhou and Wente Wang and Wenting Shen and Wenyuan Yu and Xianzhong Shi and Xiaoming Huang and Xin Xu and Yan Kou and Yangyu Lv and Yifei Li and Yijing Liu and Yiming Wang and Yingya Zhang and Yitong Huang and Yong Li and You Wu and Yu Liu and Yulin Pan and Yun Zheng and Yuntao Hong and Yupeng Shi and Yutong Feng and Zeyinzi Jiang and Zhen Han and Zhi-Fan Wu and Ziyu Liu},
      journal = {arxiv:2503.20314},
      year={2025}
}

@article{flow_matching,
  title={Flow Matching for Generative Modeling},
  author={Yaron Lipman and Ricky T. Q. Chen and Heli Ben-Hamu and Maximilian Nickel and Matt Le},
  journal={ArXiv},
  year={2022},
  volume={abs/2210.02747},
}

@inproceedings{Esser2024ScalingRF,
  title={Scaling Rectified Flow Transformers for High-Resolution Image Synthesis},
  author={Patrick Esser and Sumith Kulal and A. Blattmann and Rahim Entezari and Jonas Muller and Harry Saini and Yam Levi and Dominik Lorenz and Axel Sauer and Frederic Boesel and Dustin Podell and Tim Dockhorn and Zion English and Kyle Lacey and Alex Goodwin and Yannik Marek and Robin Rombach},
  booktitle={ICML},
  year={2024},
}

@article{pi0,
  title={$\pi$0: A Vision-Language-Action Flow Model for General Robot Control},
  author={Kevin Black and Noah Brown and Danny Driess and Adnan Esmail and Michael Equi and Chelsea Finn and Niccolo Fusai and Lachy Groom and Karol Hausman and Brian Ichter and Szymon Jakubczak and Tim Jones and Liyiming Ke and Sergey Levine and Adrian Li-Bell and Mohith Mothukuri and Suraj Nair and Karl Pertsch and Lucy Xiaoyang Shi and James Tanner and Quan Vuong and Anna Walling and Hao-Huan Wang and Ury Zhilinsky},
  journal={ArXiv},
  year={2024},
  volume={abs/2410.24164},
}

@article{Zheng2025XVLAST,
  title={X-VLA: Soft-Prompted Transformer as Scalable Cross-Embodiment Vision-Language-Action Model},
  author={Jinliang Zheng and Jianxiong Li and Zhihao Wang and Dong-Xiu Liu and Xirui Kang and Yuchun Feng and Yinan Zheng and Jia-Yi Zou and Yi-Lun Chen and Jia Zeng and Ya-Qin Zhang and Jiang-Miao Pang and Jingjing Liu and Tai Wang and Xianyuan Zhan},
  journal={ArXiv},
  year={2025},
  volume={abs/2510.10274}
}

@article{Torne2026MEMME,
  title={MEM: Multi-Scale Embodied Memory for Vision Language Action Models},
  author={Marcel Torne and Karl Pertsch and Homer Rich Walke and Kyle Vedder and Suraj Nair and Brian Ichter and Allen Z. Ren and Hao-Huan Wang and Jiaming Tang and Kyle Stachowicz and Karan Dhabalia and Michael Robert Equi and Quan Ho Vuong and Jost Tobias Springenberg and Sergey Levine and Chelsea Finn and Danny Driess},
  journal={ArXiv},
  year={2026},
  volume={abs/2603.03596}
}

@article{WLA0,
  title={World-Language-Action Model for Unified World Modeling, Language Reasoning, and Action Synthesis},
  author={Yi Yang and Zhihong Liu and Siqi Kou and Yiyang Chen and Yanzhe Hu and Jianbo Zhou and Bo Zhao and Zhijie Wei and Xiaowu Xia and Xueqi Li and Pengfei Liu and Zhijie Deng},
  journal={ArXiv},
  year={2026},
  volume={abs/2606.05979}
  }

@inproceedings{unipi,
 author = {Du, Yilun and Yang, Sherry and Dai, Bo and Dai, Hanjun and Nachum, Ofir and Tenenbaum, Josh and Schuurmans, Dale and Abbeel, Pieter},
 booktitle = {NeurIPS},
 doi = {10.52202/075280-0403},
 title = {Learning Universal Policies via Text-Guided Video Generation},
 year = {2023}
}

@article{qwenvla,
  title={Qwen-VLA: Unifying Vision-Language-Action Modeling across Tasks, Environments, and Robot Embodiments},
  author={Qiuyue Wang and Ming-Sheng Li and Jian Guan and Ji-lu Ye and Sicheng Xie and Yitao Liu and Junhao Chen and Zhixuan Liang and Jie Zhang and Xintong Hu and Xuhong Huang and Pei Lin and Junyang Lin and Dayiheng Liu and Shuai Bai and Jing Zhou and Jiazhao Zhang and Haoqi Yuan and Gengze Zhou and Hang Yin and Yebin Wang and Yiyang Huang and Zixing Lei and Wujian Peng and Delin Chen and Yingming Zheng and Jingyang Fan and Xianwei Zhuang and Xin Zhou and Haoyang Li and An-Jen Chen and Tong Zhang and Xue-Jing Liu and Yuchong Sun and Rui Chen and Zhao Li and Chenxu Lu and Zhibo Yang and Tao Yu and Xiong-Hui Chen},
  journal={ArXiv},
  year={2026},
  volume={abs/2605.30280}
}

@inproceedings{qwen3,
  title={Qwen3 Technical Report},
  author={An Yang and Anfeng Li and Baosong Yang and Beichen Zhang and Binyuan Hui and Bo Zheng and Bowen Yu and Chang Gao and Chengen Huang and Chenxu Lv and Chujie Zheng and Dayiheng Liu and Fan Zhou and Fei Huang and Feng Hu and Hao Ge and Haoran Wei and Huan Lin and Jialong Tang and Jian Yang and Jianhong Tu and Jianwei Zhang and Jianxin Yang and Jiaxin Yang and Jingren Zhou and Jingren Zhou and Junyan Lin and Kai Dang and Keqin Bao and Ke‐Pei Yang and Le Yu and Li-Chun Deng and Mei Li and Min Xue and Mingze Li and Pei Zhang and Peng Wang and Qin Zhu and Rui Men and Ruize Gao and Shixuan Liu and Shuang Luo and Tianhao Li and Tianyi Tang and Wenbiao Yin and Xingzhang Ren and Xinyu Wang and Xinyu Zhang and Xuancheng Ren and Yang Fan and Yang Su and Yi-Chao Zhang and Ying-Er Zhang and Yu Wan and Yuqiong Liu and Zekun Wang and Zeyu Cui and Zhenru Zhang and Zhipeng Zhou and Zihan Qiu},
  year={2025}
}

@article{molmo2,
  title={{Molmo2: Open Weights and Data for Vision-Language Models with Video Understanding and Grounding}},
  author={Christopher Clark and Jieyu Zhang and Zixian Ma and Jae Sung Park and Mohammadreza Salehi and Rohun Tripathi and Sangho Lee and Zhongzheng Ren and Chris Dongjoo Kim and Yinuo Yang and Vincent Shao and Yue Yang and Weikai Huang and Ziqi Gao and Taira Anderson and Jianrui Zhang and Jitesh Jain and George Stoica and Winson Han and Ali Farhadi and Ranjay Krishna},
  journal={arxiv:2601.10611},
  year={2026},
}

@article{diffusion_policy,
  title={Diffusion policy: Visuomotor policy learning via action diffusion},
  author={Cheng Chi and Siyuan Feng and Yilun Du and Zhenjia Xu and Eric Cousineau and Benjamin Burchfiel and Shuran Song},
  journal={Inter.~J.~Robo.~Res.},
  year={2023},
  volume={44},
  pages={1684 - 1704}
}

@inproceedings{eagle2.5,
    title={Eagle 2.5: Boosting Long-Context Post-Training for Frontier Vision-Language Models},
    author={Guo Chen and Zhiqi Li and Shihao Wang and Jindong Jiang and Yicheng Liu and Lidong Lu and De-An Huang and Wonmin Byeon and Matthieu Le and Max Ehrlich and Tong Lu and Limin Wang and Bryan Catanzaro and Jan Kautz and Andrew Tao and Zhiding Yu and Guilin Liu},
    booktitle={NeurIPS},
    year={2025}
}

@inproceedings{shi2025eagle,
    title = {Eagle: Exploring The Design Space for Multimodal LLMs with Mixture of Encoders}, 
    author={Min Shi and Fuxiao Liu and Shihao Wang and Shijia Liao and Subhashree Radhakrishnan and De-An Huang and Hongxu Yin and Karan Sapra and Yaser Yacoob and Humphrey Shi and Bryan Catanzaro and Andrew Tao and Jan Kautz and Zhiding Yu and Guilin Liu},
    booktitle={ICLR},
    year={2025}
}

@article{lingbotva,
  title={Causal World Modeling for Robot Control},
  author={Lin Li and Qihang Zhang and Yiming Luo and Shuai Yang and Ruilin Wang and Fei Han and Mingrui Yu and Zelin Gao and Nan Xue and Xing Zhu and Yujun Shen and Yinghao Xu},
  journal={arxiv:2601.21998},
  year={2026},
}

@inproceedings{rt22023arxiv,
    title={{RT-2: Vision-Language-Action Models Transfer Web Knowledge to Robotic Control}},
    author={Anthony Brohan and Noah Brown and Justice Carbajal and Yevgen Chebotar and Xi Chen and Krzysztof Choromanski and Tianli Ding and Danny Driess and Avinava Dubey and Chelsea Finn and Pete Florence and Chuyuan Fu and Montse Gonzalez Arenas and Keerthana Gopalakrishnan and Kehang Han and Karol Hausman and Alex Herzog and Jasmine Hsu and Brian Ichter and Alex Irpan and Nikhil Joshi and Ryan Julian and Dmitry Kalashnikov and Yuheng Kuang and Isabel Leal  and Lisa Lee and Tsang-Wei Edward Lee and Sergey Levine and Yao Lu and Henryk Michalewski and Igor Mordatch and Karl Pertsch and Kanishka Rao and Krista Reymann and Michael Ryoo and Grecia Salazar and Pannag Sanketi and Pierre Sermanet and Jaspiar Singh and Anikait Singh and Radu Soricut and Huong Tran and Vincent Vanhoucke and Quan Vuong and Ayzaan Wahid and Stefan Welker and Paul Wohlhart and  Jialin Wu and Fei Xia and Ted Xiao and Peng Xu and Sichun Xu and Tianhe Yu and Brianna Zitkovich},
    booktitle={arxiv:2307.15818},
    year={2023}
}

@inproceedings{zhao2023act,
  title={Learning Fine-Grained Bimanual Manipulation with Low-Cost Hardware},
  author={Zhao, Tony Z and Kim, Victoria and Bohg, Jiayuan and Finn, Chelsea},
  booktitle={RSS},
  year={2023}
}

@article{leWorldModel,
  title={LeWorldModel: Stable End-to-End Joint-Embedding Predictive Architecture from Pixels},
  author={Lucas Maes and Quentin Le Lidec and Damien Scieur and Yann LeCun and Randall Balestriero},
  journal={arxiv:2603.19312},
  year={2026},
}

@article{dino,
  title={Emerging Properties in Self-Supervised Vision Transformers},
  author={Mathilde Caron and Hugo Touvron and Ishan Misra and Herv{\'e} J{\'e}gou and Julien Mairal and Piotr Bojanowski and Armand Joulin},
  journal={ICCV},
  year={2021},
  pages={9630-9640}
}

@inproceedings{Lin2026JEPAWAMLV,
  title={{JEPA-WAM: Learning Vision-Language-Action Policies with Joint-Embedding World Modeling}},
  author={Yihan Lin and Jiawei He and Shifeng Bao and Chen Zhao and Yang Li and Xiaobo Wang and Yan Wang and Cheng Chi and Jing Zhang},
  year={2026}
}

@article{vjepa2,
  title={V-JEPA 2.1: Unlocking Dense Features in Video Self-Supervised Learning},
  author={Lorenzo Mur-Labadia and Matthew Muckley and Amir Bar and Mido Assran and Koustuv Sinha and Michael Rabbat and Yann LeCun and Nicolas Ballas and Adrien Bardes},
  journal={arxiv:2603.14482},
  year={2026},
}

@article{pi07,
  title={$\pi$0.7: a Steerable Generalist Robotic Foundation Model with Emergent Capabilities},
  author={{Physical Intelligence} and Bo Ai and Ali Amin and Raichelle J. Aniceto and Ashwin Balakrishna and Gregor Balke and Kevin Black and George Bokinsky and Shi-Jie Cao and Thomas Charbonnier and Vedant Choudhary and Foster Collins and Ken Conley and Grace Connors and James Darpinian and Karan Dhabalia and Maitrayee Dhaka and Jared DiCarlo and Danny Driess and Michael Equi and Adnan Esmail and Yunhao Fang and Chelsea Finn and Catherine Glossop and Thomas Godden and Ivan Goryachev and Lachlan Groom and Haroun Habeeb and Hunter Hancock and Karol Hausman and Gashon Hussein and Victor Hwang and Brian Ichter and Connor Jacobsen and Szymon Jakubczak and Rowan Jen and Tim Jones and Gregg Kammerer and Ben Katz and Liyiming Ke and Mairbek Khadikov and Chandra Kuchi and Marinda Lamb and Devin LeBlanc and Brendon LeCount and Sergey Levine and Xinyu Li and Adrian Li-Bell and Vladislav Lialin and Zhong-Zhou Liang and W. K. Lim and Yao Lu and Enyu Luo and Vishnu Mano and Nandan Marwaha and A. Mongush and Liam Murphy and Suraj Nair and Tyler M Patterson and Karl Pertsch and Allen Z. Ren and Gavin Schelske and Charvi Sharma and Baifeng Shi and Lucy Xiaoyang Shi and Laura Smith and Jost Tobias Springenberg and Kyle Stachowicz and Will Stoeckle and Jiaming Tang and James Tanner and Shalom Tekeste and Marcel Torne and Kyle Vedder and Quan Vuong and Anna Walling and Hao-Huan Wang and Jason Wang and Xudong Wang and Christopher C. Whalen and Sir John Whitmore and Blake Williams and Charles Xu and Sukwon Yoo and Lili Yu and Wu Zhang and Zhuoyang Zhang and Ury Zhilinsky},
  journal={arxiv:2604.15483},
  year={2026},
}

@article{mem,
  title={MEM: Multi-Scale Embodied Memory for Vision Language Action Models},
  author={Marcel Torne and Karl Pertsch and Homer Rich Walke and Kyle Vedder and Suraj Nair and Brian Ichter and Allen Z. Ren and Hao-Huan Wang and Jiaming Tang and Kyle Stachowicz and Karan Dhabalia and Michael Equi and Quan Ho Vuong and Jost Tobias Springenberg and Sergey Levine and Chelsea Finn and Danny Driess},
  journal={arxiv:2603.03596},
  year={2026},
}

@article{tau0,
  title={$\tau_0$-VLA: a Hierarchical Robot Foundation Model with World-Model-Guided Test-Time Computation},
  author={Xiaowei Cai and Yunuo Cai and Bin Chen and Jingxiao Chen and Zhi Chen and Siyuan Feng and Teng Hou and Jingshun Huang and Han Jiang and Runkun Ju and Dong Li and Mingxi Li and Shaowei Li and Xinchen Li and Yifan Li and Yi Liu and Zhongyuan Liu and Jianlan Luo and Junwen Miao and Ruiqi Ni and Buqing Nie and Mingjie Pan and Xinlin Ren and Jia-Yi Song and Jiaxu Wang and Peiqi Wang and Sen Wang and Xiaoyan Wang and Dafeng Wei and Dongming Wu and Pengwei Xie and Pu Yang and Hang Ye and Xiangyu Yue and Jin-Yu Zhang and Qingli Zhang and Xue Zhao and Pengfei Zhou and Yue Zhou},
  year={2026},
  journal={arxiv:2608.16885}
}

@article{rmbench,
  title={RMBench: Memory-Dependent Robotic Manipulation Benchmark with Insights into Policy Design},
  author={Tianxing Chen and Yuran Wang and Mingleyang Li and Yanjiao Qin and Haochen Shi and Zixuan Li and Yifan Hu and Yingsheng J Zhang and Kaixuan Wang and Yue Chen and Hongchen Wang and Ren-Jing Xu and Ruihai Wu and Yao Mu and Yaodong Yang and Hao Dong and Ping Luo},
  journal={arxiv:2603.01229},
  year={2026},
}

@article{memoryvla,
  title={{MemoryVLA}: Perceptual-Cognitive Memory in Vision-Language-Action Models for Robotic Manipulation},
  author={Shi, Hao and Xie, Bin and Liu, Yingfei and Sun, Lin and Liu, Fengrong and Wang, Tiancai and Zhou, Erjin and Fan, Haoqiang and Zhang, Xiangyu and Huang, Gao},
  journal={arxiv:2508.19236},
  year={2025}
}

@inproceedings{octo_2023,
    title={Octo: An Open-Source Generalist Robot Policy},
    author = {{Octo Model Team} and Dibya Ghosh and Homer Walke and Karl Pertsch and Kevin Black and Oier Mees and Sudeep Dasari and Joey Hejna and Charles Xu and Jianlan Luo and Tobias Kreiman and {You Liang} Tan and Lawrence Yunliang Chen and Pannag Sanketi and Quan Vuong and Ted Xiao and Dorsa Sadigh and Chelsea Finn and Sergey Levine},
    booktitle = {RSS},
    address  = {Delft, Netherlands},
    year = {2024},
}

@article{zadbood2017transmit,
  author = {Zadbood, Asieh and Chen, Janice and Leong, Yuan Chang and Norman, Kenneth A. and Hasson, Uri},
  title = {How We Transmit Memories to Other Brains: Constructing Shared Neural Representations Via Communication},
  journal = {Cerebral Cortex},
  year = {2017}, volume = {27}, number = {10}, pages = {4988--5000},
  doi = {10.1093/cercor/bhx202}
}

@article{lin2026cotraininglbm,
  title={A Systematic Study of Data Modalities and Strategies for Co-training Large Behavior Models for Robot Manipulation},
  author={Lin, Fanqi and Arora, Kushal and Mercat, Jean and Nishimura, Haruki and Shah, Paarth and Xu, Chen and Zhang, Mengchao and Zolotas, Mark and Angeles, Maya and Pfannenstiehl, Owen and Beaulieu, Andrew and Barreiros, Jose},
  journal={arXiv:2602.01067},
  year={2026},
}

@article{lbmtri2025,
  title={A Careful Examination of Large Behavior Models for Multitask Dexterous Manipulation},
  author={{TRI LBM Team} and Jose Barreiros and Andrew Beaulieu and Aditya Bhat and Rick Cory and Eric Cousineau and Hongkai Dai and Ching-Hsin Fang and Kunimatsu Hashimoto and Muhammad Zubair Irshad and Masha Itkina and Naveen Kuppuswamy and Kuan-Hui Lee and Katherine Liu and Dale McConachie and Ian McMahon and Haruki Nishimura and Calder Phillips-Grafflin and Charles Richter and Paarth Shah and Krishnan Srinivasan and Blake Wulfe and Chen Xu and Mengchao Zhang and Alex Alspach and Maya Angeles and Kushal Arora and Vitor Campagnolo Guizilini and Alejandro Castro and Dian Chen and Ting-Sheng Chu and Sam Creasey and Sean Curtis and Richard Denitto and Emma Dixon and Eric Dusel and Matthew Ferreira and Aimee Goncalves and Grant Gould and Damrong Guoy and Swati Gupta and Xuchen Han and Kyle Hatch and Brendan Hathaway and Allison Henry and Hillel Hochsztein and Phoebe Horgan and Shun Iwase and Donovon Jackson and Siddharth Karamcheti and Sedrick Keh and Joseph Masterjohn and Jean Mercat and Patrick Miller and Paul Mitiguy and Tony Nguyen and Jeremy Nimmer and Yuki Noguchi and Reko Ong and Aykut Onol and Owen Pfannenstiehl and Richard Poyner and Leticia Priebe Mendes Rocha and Gordon Richardson and Christopher Rodriguez and Derick Seale and Michael Sherman and Mariah Smith-Jones and David Tago and Pavel Tokmakov and Matthew Tran and Basile Van Hoorick and Igor Vasiljevic and Sergey Zakharov and Mark Zolotas and Rares Ambrus and Kerri Fetzer-Borelli and Benjamin Burchfiel and Hadas Kress-Gazit and Siyuan Feng and Stacie Ford and Russ Tedrake},
  journal={arXiv:2507.05331},
  year={2025},
}

@article{flexpi,
  title={Flex-$\pi$: A Multi-Stream World-Action Model with Compute Flexibility},
  author={Ge Yan and Jinghao Liu and Yuzhi Fan and Lei Cai and Min-Wen Liao and Jesse Zhang and Dieter Fox},
  journal={arxiv:2608.10860},
  year={2026}
}

@inproceedings{massively-multimodal,
title = {4M: Massively Multimodal Masked Modeling},
booktitle = {NeurIPS},
author = {David Mizrahi* and Roman Bachmann* and Oguzhan Fatih Kar and Teresa Yeo and Mingfei Gao and Afshin Dehghan and Amir Zamir},
year = {2023},
}

@article{cosmos3,
  title={Cosmos 3: Omnimodal World Models for Physical AI},
  author={Aditi and Niket Agarwal and Arslan Ali and Jon Allen and Martin Antolini and Adeline Aubame and Alisson G. Azzolini and Junjie Bai and Maciej Bala and Yogesh Balaji and Josh Bapst and Aarti Basant and Mukesh Beladiya and Mohammad Bhat and Zaid Pervaiz Bhat and Daniel F. Blick and Vanni Brighella and Han Cai and Tiffany Cai and Eric Cameracci and Jiaxin Cao and Yulong Cao and Mark T. Carlson and Carlos Casanova and Ting-Yun Chang and Yan Chang and Yu-Wei Chao and Prithvijit Chattopadhyay and Roshan Chaudhari and Chieh-yun Chen and Jun-Yu Chen and Kejiang Chen and Qizhi Chen and Wen-Kai Chen and Xiaotong Chen and Yu Chen and An-Chieh Cheng and Click Cheng and Xiu Li Chia and Jeana Choi and Chaeyeon Chung and Wenyan Cong and Yin Cui and Magdalena Dadela and Nalin Dadhich and Wenliang Dai and Joyjit Daw and Alperen Değirmenci and Rodrigo Vieira Del Monte and Robert Thomas Denomm{\'e} and Sameer Dharur and Marco Di Lucca and Ke Ding and Wenhao Ding and Yifan Ding and Yuzhu Dong and Nicole Drumheller and Yi-Lun Du and Aigul Dzhumamuratova and Aleksandr Efitorov and Hamid Eghbalzadeh and Naomi Eigbe and Imad El Hanafi and Hassan Eslami and Ben Falk and Jiaojiao Fan and Jim Fan and Amol Fasale and Sergiy Fefilatyev and Liang Feng and Francesco Ferroni and Sanja Fidler and Xiaojin Fu and Vikram Fugro and Prashant Tatyarao Gaikwad and TJ Galda and Katelyn Gao and Yihuai Gao and Wen-Hang Ge and Sreyan Ghosh and Arushi Goel and Vivek Goel and Akash Gokul and Rama K. Govindaraju and Jinwei Gu and Miguel Guerrero and Elfie Guo and Aryaman Gupta and Siddharth Gururani and Hugo Hadfield and Song Han and Ankur Handa and Zekun Hao and Mohammad Harrim and Ali Hassani and Nathan Hayes-Roth and Yufan He and Christopher S. Helvig and Cyrus Hogg and Ma-Lan Huang and Michael Huang and Sophia Huang and Yufan Huang and Jacob Samuel Huffman and DeLesley S. Hutchins and Suneel Indupuru and B. Ivanovic and Arihant Jain and Joel Jang and Ryan Ji and Yanan Jian and Dongfu Jiang and Jingyi Jin and Atharva Vinay Joshi and Nikhil Yogesh Joshi and Pranjali Joshi and Jaehun Jung and Weiwei Kang and Scott Kassekert and Jan Kautz and Ashna Khetan and J Kiczka and Slawomir Kierat and Gwanghyun Kim and Kuno Kim and Sunny Kim and Kezhi Kong and Xinke Kong and Zhifeng Kong and Tomasz Kornuta and Egor Krivov and Hui Kuang and Saurav Kumar and Chia-Wen Kuo and George Kurian and Wojciech Kutak and JF Lafleche and Himangshu Lahkar and Omar Laymoun and Jayjun Lee and Sang-Gyoung Lee and Gabriele Leone and Boyi Li and Freya Li and Jiajun Li and Jinfeng Li and Ling Li and Pengcheng Li and Shangru Li and Ting Li and Xiao-Long Li and Xuan Li and Zhao Li and Zhiqi Li and Haojie Liang and Maosheng Liao and Cheng-Hua Lin and Tsung-Yi Lin and Mingyang Liu and Sifei Liu and Zihan Liu and Haiying Lu and Xiangyu Lu and Alice Luo and Ruipu Luo and Wen-Jie Luo and Jiangran Lyu and Martin Ding Ma and Nic Ma and Qianli Ma and Dawid Majchrowski and Louis-Alexandre Marcoux and Miguel Martin and Qing Miao and Ashkan Mirzaei and Shreyas Misra and Kaichun Mo and Dr. Durdana Mohsin and Hye-Jeong Moon and Pawel M. Morkisz and Saeid Motiian and Kirill Motkov and Seungjun Nah and Yashraj S. Narang and Deepak Narayanan and Thabang Ngazimbi and Julian Ouyang and David Page and Yatian Pang and Sehwi Park and Mahesh Kishore Patekar and Mostofa Patwary and Marco Pavone and Trung Xuan Pham and Wei Ping and Soha Pouya and Shrimai Prabhumoye and Varun Praveen and Delin Qu and Hesam Rabeti and Morteza Ramezanali and Margaret Ingeborg Reeb and Xuanchi Ren and Kristen Rumley and Wojciech Rymer and Jun Saito and Yeongho Seol and John Shao and Piyush Shekdar and Tianwei Shen and Humphrey Shi and Min Shi and Stella Shi and Kevin J. Shih and Mohammad Shoeybi and Mateusz Sieniawski and Shuran Song and Alexander Sotelo and Amir Mojtaba Sotoodeh and Sunil Srinivasa and Vignesh Srinivasakumar and Bartosz Stefaniak and Rahul Steiger and Shangkun Sun and Jiaxiang Tang and Shitao Tang and Yan Tang and Yue Tang and Tolou Tavakkoli and Kayley Ting and Krzysztof Tomala and Wei-Cheng Tseng and Jibin Yesudas Varghese and S. F. Vasil'ev and Thomas Volk and Raju Wagwani and Roger Waleffe and Andrew Z. Wang and Boxiang Wang and Haoxiang Wang and Qiaolin Wang and Shihao Wang and Shijie Wang and Tingchang Wang and Yan Wang and Yu Wang and David Wehr and Fang-Yin Wei and Xinshuo Weng and Jay Zhangjie Wu and Kedi Wu and Hongchi Xia and S. Xiao and Tianjun Xiao and Kevin Xie and Daguang Xu and Jiashu Xu and Mengya Xu and Ruqing Xu and Xingqian Xu and Yao Xu and Dinghao Yang and Dong Yang and Han Yang and Xiaodong Yang and Xuning Yang and Yichu Yang and Yurong You and Zhiding Yu and Haobo Yuan and S. Chung Kim Yuen and Xiaohui Zeng and Pengcuo Zeren and Cindy Zha and Haotian Zhang and Jenny Zhang and Jing Zhang and Liang Zhang and Paris Zhang and Shun Zhang and Xuan Zhang and Zhizheng Zhang and Annabeth Xinyu Zhao and Yilin Zhao and Yuliya Zhautouskaya and Charles Zhou and Fengzhe Zhou and Shilin Zhu and Yuke Zhu and Dima Zhylko and Artur Zolkowski},
  journal={ArXiv},
  year={2026},
  volume={abs/2606.02800}
}

@article{zhan2025anygptunifiedmultimodalllm,
    title={AnyGPT: Unified Multimodal LLM with Discrete Sequence Modeling}, 
    author={Jun Zhan and Junqi Dai and Jiasheng Ye and Yunhua Zhou and Dong Zhang and Zhigeng Liu and Xin Zhang and Ruibin Yuan and Ge Zhang and Linyang Li and Hang Yan and Jie Fu and Tao Gui and Tianxiang Sun and Yu-Gang Jiang and Xipeng Qiu},
    journal={arxiv:2402.12226},
    year={2025},
}

@article{wu2024nextgptanytoanymultimodalllm,
    title={NExT-GPT: Any-to-Any Multimodal LLM}, 
    author={Shengqiong Wu and Hao Fei and Leigang Qu and Wei Ji and Tat-Seng Chua},
    journal={ArXiv},
    year={2024},
    volume={abs/2309.05519},
}

@article{Andersen2003Sensorimotor,
  author    = {Andersen, Richard A. and Buneo, Christopher A.},
  title     = {Sensorimotor integration in posterior parietal cortex},
  journal   = {Advances in Neurology},
  volume    = {93},
  pages     = {159--177},
  year      = {2003},
  pmid      = {12894407}
}

@article{Goodale1992Separate,
  author    = {Goodale, M. A. and Milner, A. D.},
  title     = {Separate visual pathways for perception and action},
  journal   = {Trends Neurosci.},
  volume    = {15},
  pages     = {20--25},
  year      = {1992},
  publisher = {Elsevier},
}

@article{Cisek2007,
  author = {Cisek, Paul},
  title = {Cortical mechanisms of action selection: the affordance competition hypothesis},
  journal = {Philosophical Trans.~Royal Soc.~B: Biological Sci.},
  volume = {362},
  number = {1485},
  pages = {1585--1599},
  year = {2007},
}

@book{eagleman2015brain,
  title={The Brain: The Story of You},
  author={Eagleman, D.},
  isbn={9781782116592},
  url={https://books.google.com/books?id=gUTGBwAAQBAJ},
  year={2015},
  publisher={Canongate Books}
}

@article{innermonologue,
  title={Inner Monologue: Embodied Reasoning through Planning with Language Models},
  author={Huang, Wenlong and Xia, Fei and Xiao, Ted and Chan, Harris and Liang, Jacky and Florence, Pete and Zeng, Andy and Tompson, Jonathan and Mordatch, Igor and Chebotar, Yevgen and Sermanet, Pierre and Brown, Noah and Jackson, Tomas and Luu, Linda and Levine, Sergey and Hausman, Karol and Ichter, Brian},
  journal={arXiv preprint arXiv:2207.05608},
  year={2022}
}

@article{ecot,
  title={Robotic Control via Embodied Chain-of-Thought Reasoning},
  author={Zawalski, Micha{\l} and Chen, William and Pertsch, Karl and Mees, Oier and Finn, Chelsea and Levine, Sergey},
  journal={arXiv preprint arXiv:2407.08693},
  year={2024}
}

@article{rth,
  title={RT-H: Action Hierarchies Using Language},
  author={Belkhale, Suneel and Ding, Tianli and Xiao, Ted and Sermanet, Pierre and Vuong, Quan and Tompson, Jonathan and Chebotar, Yevgen and Dwibedi, Debidatta and Sadigh, Dorsa},
  journal={arXiv preprint arXiv:2403.01823},
  year={2024}
}

@article{cotvla,
  title={CoT-VLA: Visual Chain-of-Thought Reasoning for Vision-Language-Action Models},
  author={Zhao, Qingqing and Lu, Yao and Kim, Moo Jin and Fu, Zipeng and Zhang, Zhuoyang and Wu, Yecheng and Li, Zhaoshuo and Ma, Qianli and Han, Song and Finn, Chelsea and Handa, Ankur and Liu, Ming-Yu and Xiang, Donglai and Wetzstein, Gordon and Lin, Tsung-Yi},
  journal={arXiv preprint arXiv:2503.22020},
  year={2025}
}

@inproceedings{Ross2010ARO,
  title={A Reduction of Imitation Learning and Structured Prediction to No-Regret Online Learning},
  author={St{\'e}phane Ross and Geoffrey J. Gordon and J. Andrew Bagnell},
  booktitle={International Conference on Artificial Intelligence and Statistics},
  year={2010}
}

\appendix
\etocsettocdepth.toc{subsection}
\begingroup
\renewcommand*\contentsname{Appendix Contents}
\setcounter{tocdepth}{2}
\tableofcontents
\endgroup
\bigskip

\section{Related Works}

\parahead{Generalist Visuomotor Policies} Generalist robot policies~\cite{pi0.5, pi0, Molmobot, Nvidia2025GR00TNA, qwenvla, rt22023arxiv, octo_2023, lbmtri2025} improve over single-task models~\cite{diffusion_policy, zhao2023act} by initializing from vision-language backbones~\cite{qwen3, molmo2, eagle2.5, shi2025eagle, dino, vjepa2} and training for action. These policies map observations to actions but do not model world dynamics or reason over long-horizon subtasks.

\parahead{World-Action Models} A recent line of works~\cite{FastWAM, dreamzero, cosmospolicy, Motus, unipi, leWorldModel, Lin2026JEPAWAMLV} adds a video-prediction objective, jointly modeling world dynamics and action. FastWAM~\cite{FastWAM}, closest to our work, couples a pretrained Wan2.2~\cite{wan2025} video DiT with a parallel action DiT through unified attention. These models treat language purely as a conditioning input, never as a generation target and provide no mechanism for subtask reasoning or memory.

\parahead{Reasoning, Memory, and Hybrid Policies} Long-horizon and memory-dependent tasks have motivated explicit reasoning and memory~\cite{rmbench, memoryvla}, and hybrid systems~\cite{pi07, mem, tau0, lingbotva, flexpi} that combine a high-level VLM planner, a low-level policy, and a video model to imagine action effects. Because these components are trained separately on heterogeneous data, they lack a shared representation and execute planning and control asynchronously.

\parahead{Unified and Omnimodal Models} Any-to-any modeling originates in multimodal LLMs~\cite{zhan2025anygptunifiedmultimodalllm, wu2024nextgptanytoanymultimodalllm} which omit generating actions, and  vision works like 4M~\cite{massively-multimodal}, which unify many modalities via masked token modeling but target perception rather than control and do not reason. 
Concurrent work~\cite{flexpi} additionally generates actions, DINO features, semantics, and video, but cannot reason or preserve memory during manipulation. 
Concurrent omnimodal models such as Cosmos~3~\cite{cosmos3} extend any-to-any modeling to action within a single framework, but decode text autoregressively in one tower and continuous modalities by diffusion in another, running reasoning and generation as separate stages in a 64B model. 
\coolname instead generates language, video, and action \emph{entirely by diffusion in one shared representation}, denoising language and action \emph{synchronously} so that reasoning and control co-refine in a single unified model.

\label{app:related_work}

\parahead{World-Action Models (WAM background)} World-Action Models (WAMs)~\cite{FastWAM, dreamzero, cosmospolicy} extend video-generation backbones with a parallel action head so that action prediction can leverage the priors learned by large-scale video pretraining. FastWAM~\cite{FastWAM} --- the WAM we compare against throughout --- initializes from a pretrained Wan2.2~\cite{wan2025} video DiT and attaches a matched action DiT whose tokens attend to the video tokens through a per-layer unified self-attention, followed by a cross-attention over proprioception and the task prompt. Language is consumed only as fixed conditioning; there is no language expert and no memory over time. \coolname{} keeps the same video backbone and unified-attention mechanism but adds (i) a language DiT that co-denoises subtasks and textual memory alongside action, (ii) an anchor-frame visual memory, and (iii) the FMC training objective; the FastWAM-to-\coolname{} delta therefore isolates these three additions under matched backbone.

\parahead{Multi-scale memory VLAs ($\pi$ lineage)} MEM~\cite{Torne2026MEMME} equips a pretrained VLA with mixed-modal memory: a short-horizon stream of recent video frames compressed by a video encoder, and a long-horizon stream of language capturing abstracted semantic progress (e.g., which recipe stages are complete). $\pi_{0.7}$~\cite{pi07} builds on the $\pi_{0.6}$-MEM architecture and inherits the same memory design. Both approaches thus combine visual and textual memory, similar to \coolname{}. Two key differences: (i) MEM and $\pi_{0.7}$ attach memory to a pretrained VLM/VLA backbone (autoregressive), whereas \coolname{}'s language and action experts are trained from scratch on top of a video-only backbone; and (ii) \coolname{} co-denoises language, action, and video \emph{synchronously} in a single unified diffusion representation, whereas MEM's long-horizon text summary is produced autoregressively by the pretrained VLM.

\parahead{Language-generating policies (CoT lineage)} A closely related line of work generates intermediate language during control: Inner Monologue~\cite{innermonologue}, RT-2's chain-of-thought~\cite{rt22023arxiv}, Embodied CoT (ECoT)~\cite{ecot}, RT-H language action hierarchies~\cite{rth}, and CoT-VLA~\cite{cotvla} predict rationales, subtasks, or ``language actions'' \emph{autoregressively} and consume them as conditioning for the next action within the same control step. \coolname{} differs on three axes. First, language is \emph{co-denoised} synchronously with action in a single unified diffusion representation rather than produced autoregressively before conditioning. Second, the generated string is \emph{re-consumed as prompt} on the following control step, closing an explicit memory loop across time rather than serving as within-step reasoning. Third, \coolname{} supports \emph{both} visual memory via anchor frames and textual memory via generated language, whereas the CoT lineage above uses only language-side reasoning. This distinction matters for memory-dependent tasks such as Find-the-Item: policies that reason once per step cannot carry information across step boundaries, whereas \coolname{}'s readout persists as retrievable state.

\section{Additional Experimental Details}
\label{app:eval}

\subsection{Qualitative Results}
\label{sec:qualitative_results}
\begin{figure*}[t]
  \centering
  \setlength{\tabcolsep}{2pt}
  \renewcommand{\arraystretch}{0.95}
  \begin{tabular}{@{} c c c c c @{}}
    \multicolumn{5}{@{}l@{}}{\textbf{Empty Container (real, bi-manual item removal)}}\\[0.4ex]
    \begin{minipage}[t]{0.19\textwidth}
      \centering
      \includegraphics[width=\linewidth]{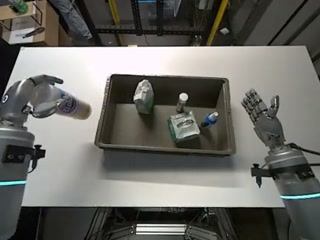}\\[-0.2ex]
      \raggedright
      {\footnotesize\color{cb-maroon}\textbf{subtask:}~left: carry Clorox wipes canister. right: idle\par}
      \vspace{0.6ex}
    \end{minipage} &
    \begin{minipage}[t]{0.19\textwidth}
      \centering
      \includegraphics[width=\linewidth]{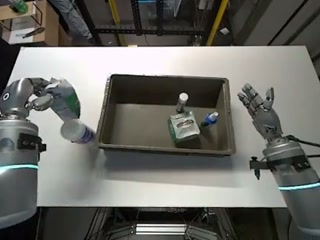}\\[-0.2ex]
      \raggedright
      {\footnotesize\color{cb-maroon}\textbf{subtask:}~left: carry green andand-white snack bag. right: idle\par}
      \vspace{0.6ex}
    \end{minipage} &
    \begin{minipage}[t]{0.19\textwidth}
      \centering
      \includegraphics[width=\linewidth]{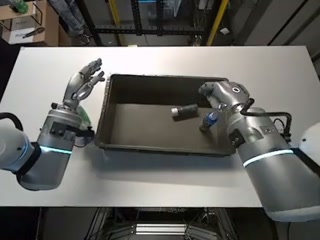}\\[-0.2ex]
      \raggedright
      {\footnotesize\color{cb-maroon}\textbf{subtask:}~left: idle. right: carry small white bottle\par}
      \vspace{0.6ex}
    \end{minipage} &
    \begin{minipage}[t]{0.19\textwidth}
      \centering
      \includegraphics[width=\linewidth]{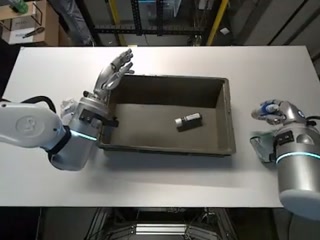}\\[-0.2ex]
      \raggedright
      {\footnotesize\color{cb-maroon}\textbf{subtask:}~left: idle. right: carry\par}
      \vspace{0.6ex}
    \end{minipage} &
    \begin{minipage}[t]{0.19\textwidth}
      \centering
      \includegraphics[width=\linewidth]{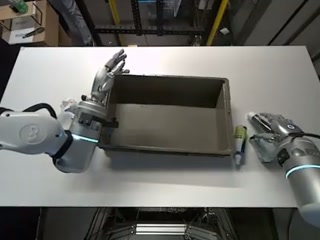}\\[-0.2ex]
      \raggedright
      {\footnotesize\color{cb-maroon}\textbf{subtask:}~left: idle. right: carry\par}
      \vspace{0.6ex}
    \end{minipage}\\[0.5ex]
    \begin{minipage}[t]{0.19\textwidth}
      \centering
      \includegraphics[width=\linewidth]{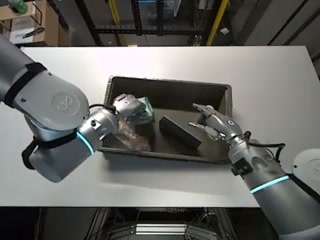}\\[-0.2ex]
      \raggedright
      {\footnotesize\color{cb-maroon}\textbf{subtask:}~left: carry green Kimtech wipes bag. right: idle\par}
      \vspace{0.6ex}
    \end{minipage} &
    \begin{minipage}[t]{0.19\textwidth}
      \centering
      \includegraphics[width=\linewidth]{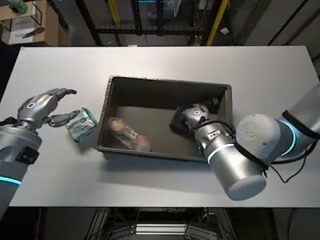}\\[-0.2ex]
      \raggedright
      {\footnotesize\color{cb-maroon}\textbf{subtask:}~left: idle. right: grasp\par}
      \vspace{0.6ex}
    \end{minipage} &
    \begin{minipage}[t]{0.19\textwidth}
      \centering
      \includegraphics[width=\linewidth]{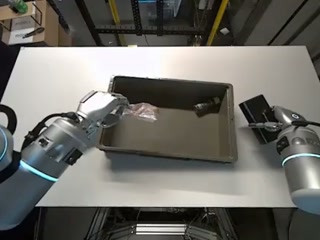}\\[-0.2ex]
      \raggedright
      {\footnotesize\color{cb-maroon}\textbf{subtask:}~left: carry clear plastic bag with bread roll. right: idle\par}
      \vspace{0.6ex}
    \end{minipage} &
    \begin{minipage}[t]{0.19\textwidth}
      \centering
      \includegraphics[width=\linewidth]{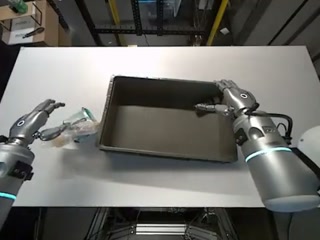}\\[-0.2ex]
      \raggedright
      {\footnotesize\color{cb-maroon}\textbf{subtask:}~left: place clear plastic bag of bread roll. right: idle\par}
      \vspace{0.6ex}
    \end{minipage} &
    \begin{minipage}[t]{0.19\textwidth}
      \centering
      \includegraphics[width=\linewidth]{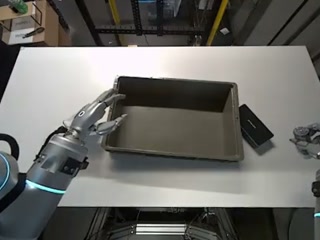}\\[-0.2ex]
      \raggedright
      {\footnotesize\color{cb-maroon}\textbf{subtask:}~left: idle. right: carry blue spray bottle\par}
      \vspace{0.6ex}
    \end{minipage}\\[0.5ex]
  \end{tabular}
  \caption{\textbf{Empty Container (real, bi-manual) rollouts.} Each row is a successful \coolname{} rollout. The {\color{cb-maroon}subtask} under each frame is \emph{generated by the language DiT in tandem with the action} (co-denoised in the same forward pass).}
  \label{fig:qualitative_real_dexterous}
\end{figure*}

\begin{figure*}[t]
  \centering
  \setlength{\tabcolsep}{2pt}
  \renewcommand{\arraystretch}{0.95}
  \begin{tabular}{@{} c c c c c @{}}
    \multicolumn{5}{@{}l@{}}{\textbf{Find the Item (real, memory)}}\\[0.4ex]
    \begin{minipage}[t]{0.19\textwidth}
      \centering
      \includegraphics[width=\linewidth]{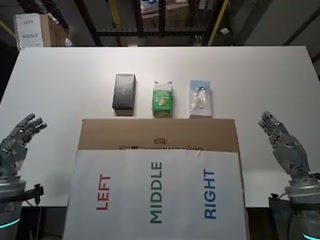}\\[-0.2ex]
      \raggedright
      {\footnotesize\color{cb-maroon}\textbf{subtask:}~left: grasp the cardboard box. right: grasp the cardboard box\par}
      {\footnotesize\color{cb-dark-blue}\textbf{memory:}~L:black box M:green parmesan package R:white tissue box\par}
      \vspace{0.6ex}
    \end{minipage} &
    \begin{minipage}[t]{0.19\textwidth}
      \centering
      \includegraphics[width=\linewidth]{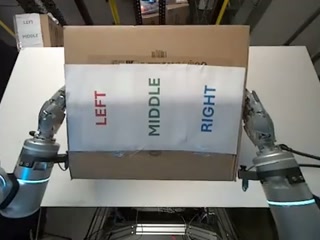}\\[-0.2ex]
      \raggedright
      {\footnotesize\color{cb-maroon}\textbf{subtask:}~left: cover the cardboard box. right: cover the cardboard box\par}
      {\footnotesize\color{cb-dark-blue}\textbf{memory:}~L:black box M:green par boxan package R:white tissue box\par}
      \vspace{0.6ex}
    \end{minipage} &
    \begin{minipage}[t]{0.19\textwidth}
      \centering
      \includegraphics[width=\linewidth]{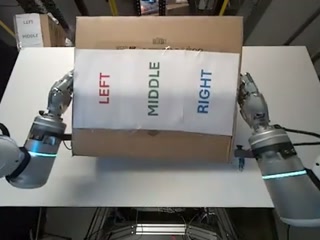}\\[-0.2ex]
      \raggedright
      {\footnotesize\color{cb-maroon}\textbf{subtask:}~place hand on the region\par}
      {\footnotesize\color{cb-dark-blue}\textbf{memory:}~L:black box M:green parmesan package R:white tissue box\par}
      \vspace{0.6ex}
    \end{minipage} &
    \begin{minipage}[t]{0.19\textwidth}
      \centering
      \includegraphics[width=\linewidth]{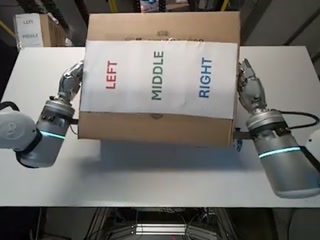}\\[-0.2ex]
      \raggedright
      {\footnotesize\color{cb-maroon}\textbf{subtask:}~place hand on left region\par}
      {\footnotesize\color{cb-dark-blue}\textbf{memory:}~L:black box M:green parmesan package R:white tissue box\par}
      \vspace{0.6ex}
    \end{minipage} &
    \begin{minipage}[t]{0.19\textwidth}
      \centering
      \includegraphics[width=\linewidth]{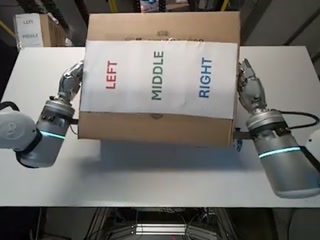}\\[-0.2ex]
      \raggedright
      {\footnotesize\color{cb-maroon}\textbf{subtask:}~place hand on left region\par}
      {\footnotesize\color{cb-dark-blue}\textbf{memory:}~L:black box M:green parmesan package R:white tissue box\par}
      \vspace{0.6ex}
    \end{minipage}\\[0.5ex]
    \begin{minipage}[t]{0.19\textwidth}
      \centering
      \includegraphics[width=\linewidth]{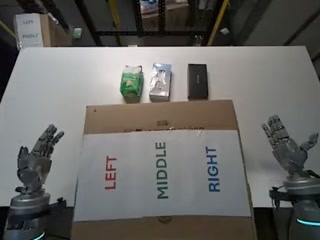}\\[-0.2ex]
      \raggedright
      {\footnotesize\color{cb-maroon}\textbf{subtask:}~left: grasp the cardboard box. right: grasp the cardboard box\par}
      {\footnotesize\color{cb-dark-blue}\textbf{memory:}~L:green parmesan package M:white tissue box R:black box\par}
      \vspace{0.6ex}
    \end{minipage} &
    \begin{minipage}[t]{0.19\textwidth}
      \centering
      \includegraphics[width=\linewidth]{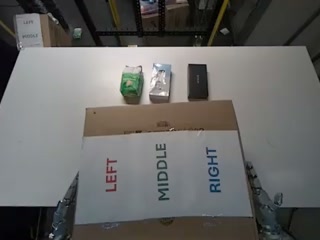}\\[-0.2ex]
      \raggedright
      {\footnotesize\color{cb-maroon}\textbf{subtask:}~left: grasp the cardboard box. right: grasp the cardboard box\par}
      {\footnotesize\color{cb-dark-blue}\textbf{memory:}~L:green parmesan package M:white tissue box R:black box\par}
      \vspace{0.6ex}
    \end{minipage} &
    \begin{minipage}[t]{0.19\textwidth}
      \centering
      \includegraphics[width=\linewidth]{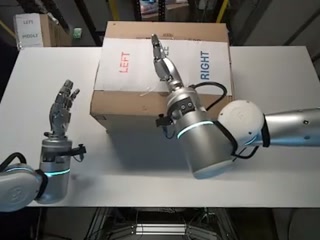}\\[-0.2ex]
      \raggedright
      {\footnotesize\color{cb-maroon}\textbf{subtask:}~left: cover the cardboard box. right: cover the cardboard box\par}
      {\footnotesize\color{cb-dark-blue}\textbf{memory:}~L:green parmesan package M:white tissue box R:black box\par}
      \vspace{0.6ex}
    \end{minipage} &
    \begin{minipage}[t]{0.19\textwidth}
      \centering
      \includegraphics[width=\linewidth]{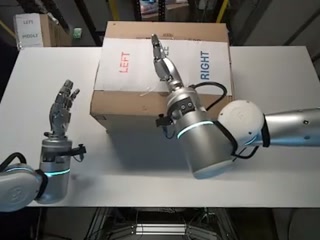}\\[-0.2ex]
      \raggedright
      {\footnotesize\color{cb-maroon}\textbf{subtask:}~left: cover the cardboard box. right: cover the cardboard box\par}
      {\footnotesize\color{cb-dark-blue}\textbf{memory:}~L:green parmesan package M:white tissue box R:black box\par}
      \vspace{0.6ex}
    \end{minipage} &
    \begin{minipage}[t]{0.19\textwidth}
      \centering
      \includegraphics[width=\linewidth]{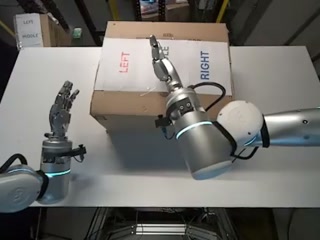}\\[-0.2ex]
      \raggedright
      {\footnotesize\color{cb-maroon}\textbf{subtask:}~place hand on middle region\par}
      {\footnotesize\color{cb-dark-blue}\textbf{memory:}~L:green parmesan package M:white tissue box R:black box\par}
      \vspace{0.6ex}
    \end{minipage}\\[0.5ex]
    \begin{minipage}[t]{0.19\textwidth}
      \centering
      \includegraphics[width=\linewidth]{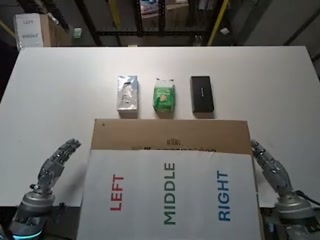}\\[-0.2ex]
      \raggedright
      {\footnotesize\color{cb-maroon}\textbf{subtask:}~left: grasp the cardboard box. right: grasp the cardboard box\par}
      {\footnotesize\color{cb-dark-blue}\textbf{memory:}~L:white tissue box M:green parmesan package R:black box\par}
      \vspace{0.6ex}
    \end{minipage} &
    \begin{minipage}[t]{0.19\textwidth}
      \centering
      \includegraphics[width=\linewidth]{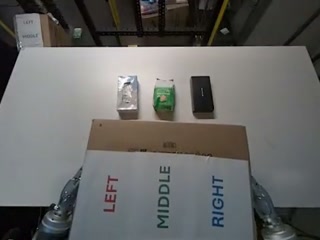}\\[-0.2ex]
      \raggedright
      {\footnotesize\color{cb-maroon}\textbf{subtask:}~left: grasp the cardboard box. right: grasp the cardboard box\par}
      {\footnotesize\color{cb-dark-blue}\textbf{memory:}~L:white tissue box M:green parmesan package R:black box\par}
      \vspace{0.6ex}
    \end{minipage} &
    \begin{minipage}[t]{0.19\textwidth}
      \centering
      \includegraphics[width=\linewidth]{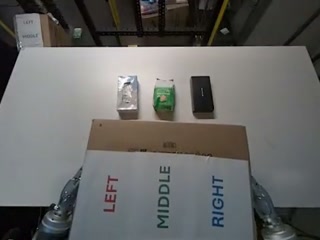}\\[-0.2ex]
      \raggedright
      {\footnotesize\color{cb-maroon}\textbf{subtask:}~left: grasp the cardboard box. right: grasp the cardboard box\par}
      {\footnotesize\color{cb-dark-blue}\textbf{memory:}~L:white tissue box M:green parmesan package R:black box\par}
      \vspace{0.6ex}
    \end{minipage} &
    \begin{minipage}[t]{0.19\textwidth}
      \centering
      \includegraphics[width=\linewidth]{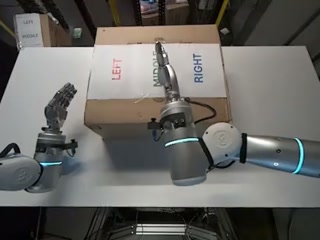}\\[-0.2ex]
      \raggedright
      {\footnotesize\color{cb-maroon}\textbf{subtask:}~left: cover the cardboard box. right: grasp the cardboard box\par}
      {\footnotesize\color{cb-dark-blue}\textbf{memory:}~L:white tissue box M:green parmesan package R:black box\par}
      \vspace{0.6ex}
    \end{minipage} &
    \begin{minipage}[t]{0.19\textwidth}
      \centering
      \includegraphics[width=\linewidth]{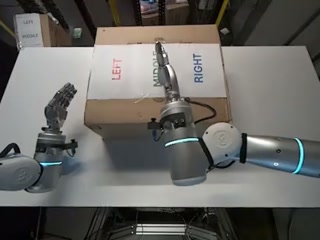}\\[-0.2ex]
      \raggedright
      {\footnotesize\color{cb-maroon}\textbf{subtask:}~left: cover the cardboard box. right: grasp the cardboard box\par}
      {\footnotesize\color{cb-dark-blue}\textbf{memory:}~L:white tissue box M:green parmesan package R:black box\par}
      \vspace{0.6ex}
    \end{minipage}\\[0.5ex]
  \end{tabular}
  \caption{\textbf{Real-world \emph{Find the Item} rollouts.} Row 1: \emph{black box}; Row 2: \emph{white tissue box}; Row 3: \emph{green parmesan package} (ends at the cover-complete moment). Items are hidden mid-episode under cardboard covers; the {\color{cb-maroon}subtask} and {\color{cb-dark-blue}textual memory} under each frame are \emph{generated by the language DiT in tandem with the action}, and the memory string is re-consumed as the next-step prompt to preserve the item--region (L/M/R) mapping across the occlusion.}
  \label{fig:qualitative_real_memory}
\end{figure*}

\begin{figure*}[t]
  \centering
  \setlength{\tabcolsep}{2pt}
  \renewcommand{\arraystretch}{0.95}
  \begin{tabular}{@{} c c c c c @{}}
    \multicolumn{5}{@{}l@{}}{\textbf{MolmoSpaces Pick \& Place}}\\[0.4ex]
    \begin{minipage}[t]{0.19\textwidth}
      \centering
      \includegraphics[width=\linewidth]{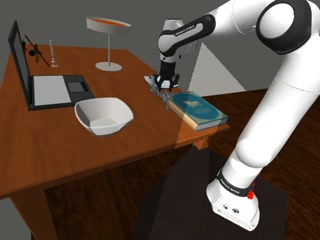}\\[-0.2ex]
      \raggedright
      {\footnotesize\color{cb-maroon}\textbf{subtask:}~Now: Grasp the metallic cursive sign. Next: Close gripper on the metallic cursive sign\par}
      \vspace{0.6ex}
    \end{minipage} &
    \begin{minipage}[t]{0.19\textwidth}
      \centering
      \includegraphics[width=\linewidth]{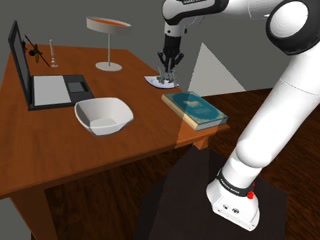}\\[-0.2ex]
      \raggedright
      {\footnotesize\color{cb-maroon}\textbf{subtask:}~Now: Lift the metallic cursive sign off the surface. Next: Transport to the smooth translucent glass bowl\par}
      \vspace{0.6ex}
    \end{minipage} &
    \begin{minipage}[t]{0.19\textwidth}
      \centering
      \includegraphics[width=\linewidth]{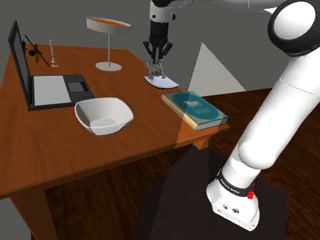}\\[-0.2ex]
      \raggedright
      {\footnotesize\color{cb-maroon}\textbf{subtask:}~Now: Carry to the smooth translucent glass bowl. Next: Release onto the smooth translucent glass bowl\par}
      \vspace{0.6ex}
    \end{minipage} &
    \begin{minipage}[t]{0.19\textwidth}
      \centering
      \includegraphics[width=\linewidth]{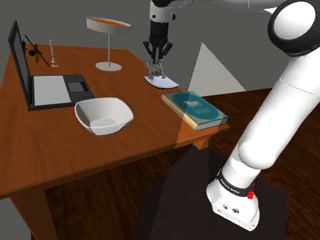}\\[-0.2ex]
      \raggedright
      {\footnotesize\color{cb-maroon}\textbf{subtask:}~Now: Carry to the smooth translucent glass bowl. Next: Release onto the smooth translucent glass bowl\par}
      \vspace{0.6ex}
    \end{minipage} &
    \begin{minipage}[t]{0.19\textwidth}
      \centering
      \includegraphics[width=\linewidth]{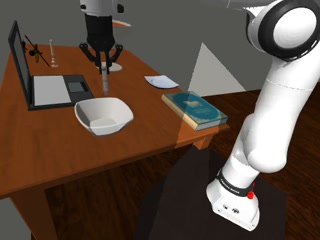}\\[-0.2ex]
      \raggedright
      {\footnotesize\color{cb-maroon}\textbf{subtask:}~Now: Set down on the smooth translucent glass bowl. Next: Open gripper\par}
      \vspace{0.6ex}
    \end{minipage}\\[0.5ex]
    \begin{minipage}[t]{0.19\textwidth}
      \centering
      \includegraphics[width=\linewidth]{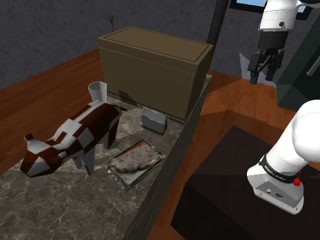}\\[-0.2ex]
      \raggedright
      {\footnotesize\color{cb-maroon}\textbf{subtask:}~Now: Open gripper. Next: Reach for the white calendar\par}
      \vspace{0.6ex}
    \end{minipage} &
    \begin{minipage}[t]{0.19\textwidth}
      \centering
      \includegraphics[width=\linewidth]{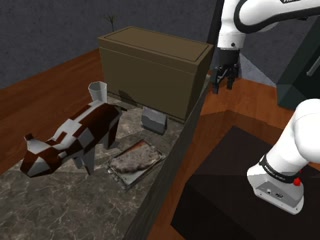}\\[-0.2ex]
      \raggedright
      {\footnotesize\color{cb-maroon}\textbf{subtask:}~Now: Move hand toward the white calendar. Next: Close fingers around the white calendar\par}
      \vspace{0.6ex}
    \end{minipage} &
    \begin{minipage}[t]{0.19\textwidth}
      \centering
      \includegraphics[width=\linewidth]{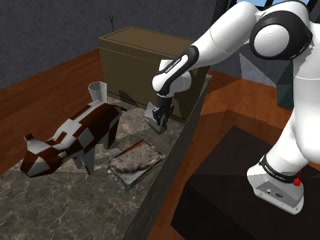}\\[-0.2ex]
      \raggedright
      {\footnotesize\color{cb-maroon}\textbf{subtask:}~Now: Firmly hold the white calendar. Next: Pick up the white calendar\par}
      \vspace{0.6ex}
    \end{minipage} &
    \begin{minipage}[t]{0.19\textwidth}
      \centering
      \includegraphics[width=\linewidth]{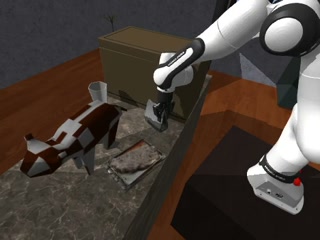}\\[-0.2ex]
      \raggedright
      {\footnotesize\color{cb-maroon}\textbf{subtask:}~Now: Firmly hold the white calendar. Next: Lift the white calendar\par}
      \vspace{0.6ex}
    \end{minipage} &
    \begin{minipage}[t]{0.19\textwidth}
      \centering
      \includegraphics[width=\linewidth]{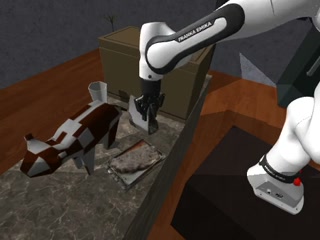}\\[-0.2ex]
      \raggedright
      {\footnotesize\color{cb-maroon}\textbf{subtask:}~Now: Carry to the rusty metal tray with meat. Next: Set down on the rusty metal tray with meat\par}
      \vspace{0.6ex}
    \end{minipage}\\[0.5ex]
    \begin{minipage}[t]{0.19\textwidth}
      \centering
      \includegraphics[width=\linewidth]{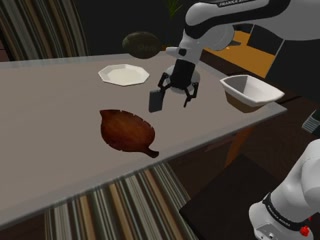}\\[-0.2ex]
      \raggedright
      {\footnotesize\color{cb-maroon}\textbf{subtask:}~Now: Grab the sleek ipod with screen and wheel. Next: Firmly hold the sleek ipod with screen and wheel\par}
      \vspace{0.6ex}
    \end{minipage} &
    \begin{minipage}[t]{0.19\textwidth}
      \centering
      \includegraphics[width=\linewidth]{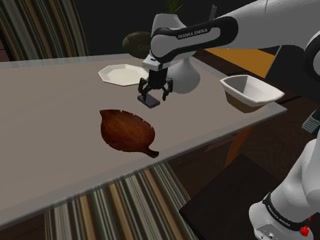}\\[-0.2ex]
      \raggedright
      {\footnotesize\color{cb-maroon}\textbf{subtask:}~Now: Wrap fingers around the sleek ipod with screen and wheel. Next: Firmly hold the sleek ipod with screen and wheel\par}
      \vspace{0.6ex}
    \end{minipage} &
    \begin{minipage}[t]{0.19\textwidth}
      \centering
      \includegraphics[width=\linewidth]{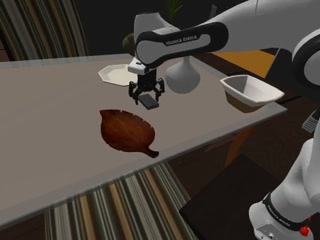}\\[-0.2ex]
      \raggedright
      {\footnotesize\color{cb-maroon}\textbf{subtask:}~Now: Grip the sleek ipod with screen and wheel. Next: Firmly hold the sleek ipod with screen and wheel\par}
      \vspace{0.6ex}
    \end{minipage} &
    \begin{minipage}[t]{0.19\textwidth}
      \centering
      \includegraphics[width=\linewidth]{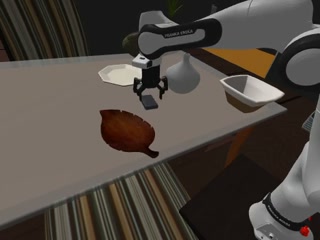}\\[-0.2ex]
      \raggedright
      {\footnotesize\color{cb-maroon}\textbf{subtask:}~Now: Close fingers around the sleek ipod with screen and wheel. Next: Firmly hold the sleek ipod with screen and wheel\par}
      \vspace{0.6ex}
    \end{minipage} &
    \begin{minipage}[t]{0.19\textwidth}
      \centering
      \includegraphics[width=\linewidth]{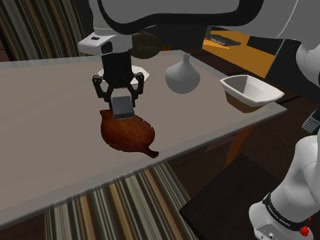}\\[-0.2ex]
      \raggedright
      {\footnotesize\color{cb-maroon}\textbf{subtask:}~Now: Open gripper. Next: Withdraw from the brown wooden leaf tray\par}
      \vspace{0.6ex}
    \end{minipage}\\[0.5ex]
  \end{tabular}
  \caption{\textbf{MolmoSpaces Pick \& Place rollouts (sim).} Each row is a successful \coolname{} rollout across diverse scenes. The {\color{cb-maroon}subtask} under each frame is \emph{generated by the language DiT in tandem with the action} (co-denoised in the same forward pass), not produced autoregressively before action.}
  \label{fig:qualitative_sim_tasks}
\end{figure*}

\begin{figure*}[t]
  \centering
  \setlength{\tabcolsep}{2pt}
  \renewcommand{\arraystretch}{0.95}
  \begin{tabular}{@{} c c c c c @{}}
    \multicolumn{5}{@{}l@{}}{\textbf{MolmoSpaces Pick \& Place}}\\[0.4ex]
    \begin{minipage}[t]{0.19\textwidth}
      \centering
      \includegraphics[width=\linewidth]{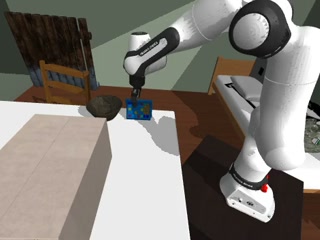}\\[-0.2ex]
      \raggedright
      {\footnotesize\color{cb-maroon}\textbf{subtask:}~Now: Close fingers around the calendar. Next: Secure grip on the calendar\par}
      \vspace{0.6ex}
    \end{minipage} &
    \begin{minipage}[t]{0.19\textwidth}
      \centering
      \includegraphics[width=\linewidth]{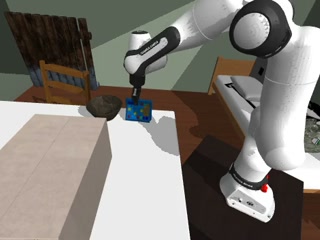}\\[-0.2ex]
      \raggedright
      {\footnotesize\color{cb-maroon}\textbf{subtask:}~Now: Close gripper on the calendar. Next: Lift the calendar off the surface\par}
      \vspace{0.6ex}
    \end{minipage} &
    \begin{minipage}[t]{0.19\textwidth}
      \centering
      \includegraphics[width=\linewidth]{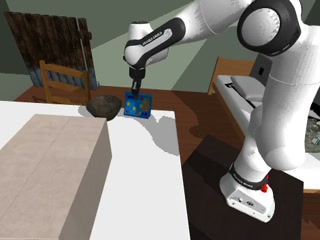}\\[-0.2ex]
      \raggedright
      {\footnotesize\color{cb-maroon}\textbf{subtask:}~Now: Close gripper on the calendar. Next: Raise the calendar\par}
      \vspace{0.6ex}
    \end{minipage} &
    \begin{minipage}[t]{0.19\textwidth}
      \centering
      \includegraphics[width=\linewidth]{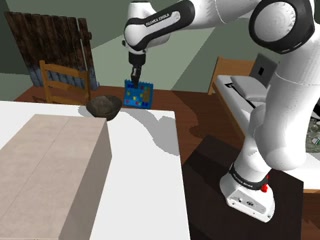}\\[-0.2ex]
      \raggedright
      {\footnotesize\color{cb-maroon}\textbf{subtask:}~Now: Raise the calendar. Next: Carry to the rustic gray bowl\par}
      \vspace{0.6ex}
    \end{minipage} &
    \begin{minipage}[t]{0.19\textwidth}
      \centering
      \includegraphics[width=\linewidth]{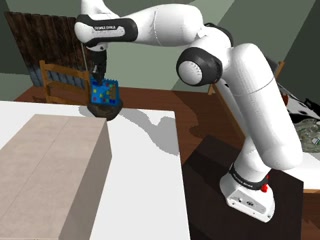}\\[-0.2ex]
      \raggedright
      {\footnotesize\color{cb-maroon}\textbf{subtask:}~Now: Release onto the rustic gray bowl. Next: Open gripper\par}
      \vspace{0.6ex}
    \end{minipage}\\[0.5ex]
    \begin{minipage}[t]{0.19\textwidth}
      \centering
      \includegraphics[width=\linewidth]{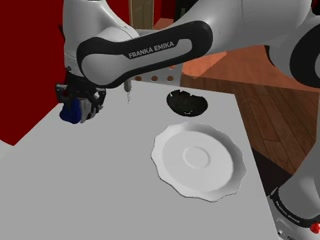}\\[-0.2ex]
      \raggedright
      {\footnotesize\color{cb-maroon}\textbf{subtask:}~Now: Wrap fingers around the gray cylindrical bacterium. Next: Firmly hold the gray cylindrical bacterium\par}
      \vspace{0.6ex}
    \end{minipage} &
    \begin{minipage}[t]{0.19\textwidth}
      \centering
      \includegraphics[width=\linewidth]{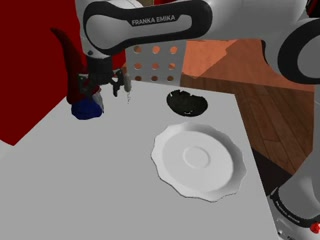}\\[-0.2ex]
      \raggedright
      {\footnotesize\color{cb-maroon}\textbf{subtask:}~Now: Failed to hold the gray cylindrical bacterium. Next: Move hand toward the gray cylindrical bacterium\par}
      \vspace{0.6ex}
    \end{minipage} &
    \begin{minipage}[t]{0.19\textwidth}
      \centering
      \includegraphics[width=\linewidth]{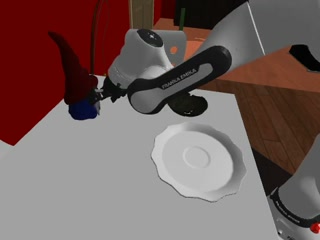}\\[-0.2ex]
      \raggedright
      {\footnotesize\color{cb-maroon}\textbf{subtask:}~Now: Move hand toward the gray cylindrical bacterium. Next: Close fingers around the gray cylindrical bacterium\par}
      \vspace{0.6ex}
    \end{minipage} &
    \begin{minipage}[t]{0.19\textwidth}
      \centering
      \includegraphics[width=\linewidth]{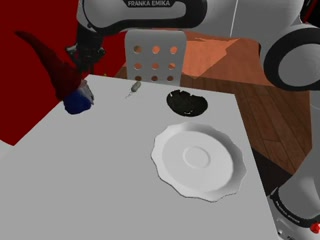}\\[-0.2ex]
      \raggedright
      {\footnotesize\color{cb-maroon}\textbf{subtask:}~Now: Wrap fingers around the gray cylindrical bacterium. Next: Firmly hold the gray cylindrical bacterium\par}
      \vspace{0.6ex}
    \end{minipage} &
    \begin{minipage}[t]{0.19\textwidth}
      \centering
      \includegraphics[width=\linewidth]{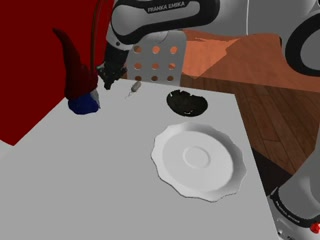}\\[-0.2ex]
      \raggedright
      {\footnotesize\color{cb-maroon}\textbf{subtask:}~Now: Wrap fingers around the gray cylindrical bacterium. Next: Close gripper on the gray cylindrical bacterium\par}
      \vspace{0.6ex}
    \end{minipage}\\[0.5ex]
    \begin{minipage}[t]{0.19\textwidth}
      \centering
      \includegraphics[width=\linewidth]{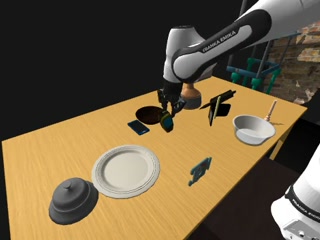}\\[-0.2ex]
      \raggedright
      {\footnotesize\color{cb-maroon}\textbf{subtask:}~Now: Lift the yellow handheld gps with antenna. Next: Carry to the shallow round wooden bowl with grain\par}
      \vspace{0.6ex}
    \end{minipage} &
    \begin{minipage}[t]{0.19\textwidth}
      \centering
      \includegraphics[width=\linewidth]{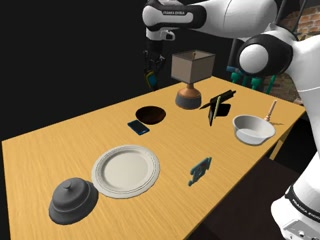}\\[-0.2ex]
      \raggedright
      {\footnotesize\color{cb-maroon}\textbf{subtask:}~Now: Carry to the shallow round wooden bowl with grain. Next: Put down on the shallow round wooden bowl with grain\par}
      \vspace{0.6ex}
    \end{minipage} &
    \begin{minipage}[t]{0.19\textwidth}
      \centering
      \includegraphics[width=\linewidth]{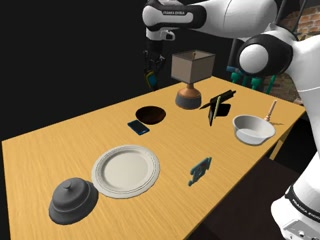}\\[-0.2ex]
      \raggedright
      {\footnotesize\color{cb-maroon}\textbf{subtask:}~Now: Carry to the shallow round wooden bowl with grain. Next: Put down on the shallow round wooden bowl with grain\par}
      \vspace{0.6ex}
    \end{minipage} &
    \begin{minipage}[t]{0.19\textwidth}
      \centering
      \includegraphics[width=\linewidth]{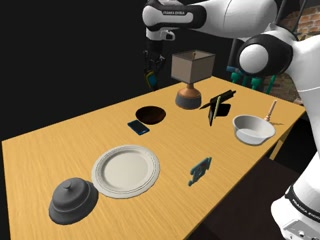}\\[-0.2ex]
      \raggedright
      {\footnotesize\color{cb-maroon}\textbf{subtask:}~Now: Carry to the shallow round wooden bowl with grain. Next: Put down on the shallow round wooden bowl with grain\par}
      \vspace{0.6ex}
    \end{minipage} &
    \begin{minipage}[t]{0.19\textwidth}
      \centering
      \includegraphics[width=\linewidth]{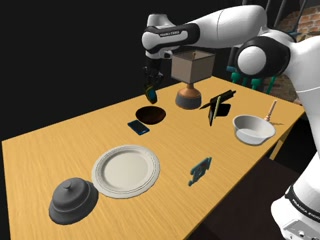}\\[-0.2ex]
      \raggedright
      {\footnotesize\color{cb-maroon}\textbf{subtask:}~Now: Place on the shallow round wooden bowl with grain. Next: Release gripper\par}
      \vspace{0.6ex}
    \end{minipage}\\[0.5ex]
  \end{tabular}
  \label{fig:qualitative_sim_tasks_b}
    \caption{\textbf{MolmoSpaces Pick \& Place rollouts (sim).} Each row is a successful \coolname{} rollout across diverse scenes. The {\color{cb-maroon}subtask} under each frame is \emph{generated by the language DiT in tandem with the action} (co-denoised in the same forward pass), not produced autoregressively before action.}
\end{figure*}

We provide qualitative rollout galleries that show how \coolname{}'s predicted {\color{cb-maroon}subtasks} and (for memory tasks) its {\color{cb-dark-blue}re-consumed textual memory} evolve across time. \Cref{fig:qualitative_rmbench} covers the four RMBench $M(n)$ tasks (Press Button, Battery Try, Blocks Ranking Try, Cover Blocks); \cref{fig:qualitative_sim_tasks,fig:qualitative_sim_tasks_b} show MolmoSpaces Pick \& Place rollouts across diverse scenes; \cref{fig:qualitative_real_dexterous} shows the real-world Empty Container task; and \cref{fig:qualitative_real_memory} shows the real-world \emph{Find the Item} memory task where the textual memory preserves the hidden item--region mapping across the occlusion.

\subsection{RMBench Environment and Tasks}
The simulation environment for memory tasks uses RMBench~\cite{rmbench}. RMBench organizes memory-dependent manipulation by \emph{task memory complexity}: $M(N)$ denotes the class of tasks whose optimal decisions depend on retaining $N$ task-relevant past observations, capturing non-local and multi-step temporal dependencies. We evaluate on the four $M(N)$ tasks --- (1) Press Button, (2) Blocks Ranking Try, (3) Cover Blocks, and (4) Battery Try --- which each require repeated exploration or trial-and-error and cannot be solved from the current observation alone. RMBench is built on the RoboTwin~2.0 platform running on the SAPIEN physics engine, with a fixed \emph{dual-arm bimanual} tabletop workspace using two AgileX arms. For each task we follow RMBench's protocol of 50 expert demonstrations for training and 100 rollout episodes for evaluation.
Because RMBench requires repeated-attempt, multi-step tracking, we enable the anchor frame and task-completion channel for this benchmark; the MolmoSpaces and real-world experiments do not require them and run without them. Their implementation details are given in \cref{sec:arch}.
The following are the task descriptions:
\begin{itemize}
    \item \textbf{Press Button}: press button involves pressing a button based on a count presented above the button. Given three buttons left, middle and right (confirm), the robot needs to read the number shown above left and middle button and press the respective button K times where K is the number displayed above the button.
    The robot then needs to press the confirm button.
    This task involves remembering the count of button presses followed pressing the confirm button.
    \item \textbf{Blocks Ranking Try}: The scene contains three colored blocks in a random initial arrangement and a confirm button. The task is to discover a specific target permutation by trial: the robot rearranges the blocks, hits the confirm button, receives feedback on whether the arrangement is correct, and repeats. Between attempts, the policy must retain which orderings have already been ruled out so it does not repeat a failed one.
    \item \textbf{Cover Blocks}: The robot faces three colored blocks and three covers. It first places a cover on each block from left to right, then must remove the covers in a fixed color order (red, then green, then blue) and return each cover to where it started. Because the blocks become visually hidden after covering, solving the second phase requires remembering which block was placed under each cover.
    \item \textbf{Battery Try}: A pair of batteries with random initial poses and a two-slot holder are staged on the table. The robot iteratively tries (battery, slot, orientation) combinations until both batteries seat correctly; combinations that previously failed must be recalled so the same attempt is not repeated.
\end{itemize}

\subsection{RMBench Training Configuration}
\label{app:rmbench_training}
For the RMBench $M(n)$ evaluation we train a separate \coolname{} on the RMBench-provided demonstrations for 100k optimizer steps with batch size 448, which takes approximately 36 hours on 10 H100 nodes. The language DiT emits 64 language tokens per diffusion forward pass; all other model settings match the Implementation Details in \cref{sec:arch}.

\subsection{Memory Architecture Details}
\label{app:memory_details}
The following architecture choices for \coolname{}'s memory channels apply across all benchmarks, independent of the specific evaluation setup.

\parahead{Visual Memory} \coolname{} conditions on one or more visual-memory keyframes captured during the episode. In our current experiments we use a single \emph{anchor frame} $\mathcal{F}$ taken at episode start, which provides a persistent visual reference of the initial scene layout (e.g., pre-occlusion object positions) that is otherwise lost as later observations advance. Each visual-memory frame is passed through the same VAE encoder as the current observation and concatenated along the token dimension into the video stream, so the language and action DiTs attend to it via the shared unified self-attention. To avoid position-encoding overlap with the current observation, we shift the RoPE~\cite{OminiControl} coordinates of each visual-memory frame by the frame width $W$ and vary the 0$^{th}$ RoPE coordinate based on video timestep. The design accepts an arbitrary number of frames (e.g., a bank of task-relevant keyframes selected online), but we defer multi-frame visual memory to future work. Because the visual-memory keyframes and the observation are on separate RoPE grids, adding or removing memory frames does not change the network's parameter count or interface; it only changes the number of memory tokens attended to.

\parahead{Task-Completion Supervision} Task-completion signals are sparse, so we supervise the appended 1-D channel as a step function within each action chunk: 0 before the subtask completes and 1 at and after the completion timestep. Being part of the action vector, the channel is trained by the same flow-matching objective $\mathcal{L}_{action}$, with no extra head or loss term.

\subsection{Flow Matching with Correction (FMC) Ablations on 2D Robot Trajectories}
\label{app:fmc_sweep}
\begin{figure}[t]
\centering
\setlength{\tabcolsep}{3pt}
\renewcommand{\arraystretch}{1.0}
\begin{tabular}{@{}cccc@{}}
\includegraphics[width=0.22\textwidth, height=0.22\textwidth, trim={0 0 1430 0}, clip]{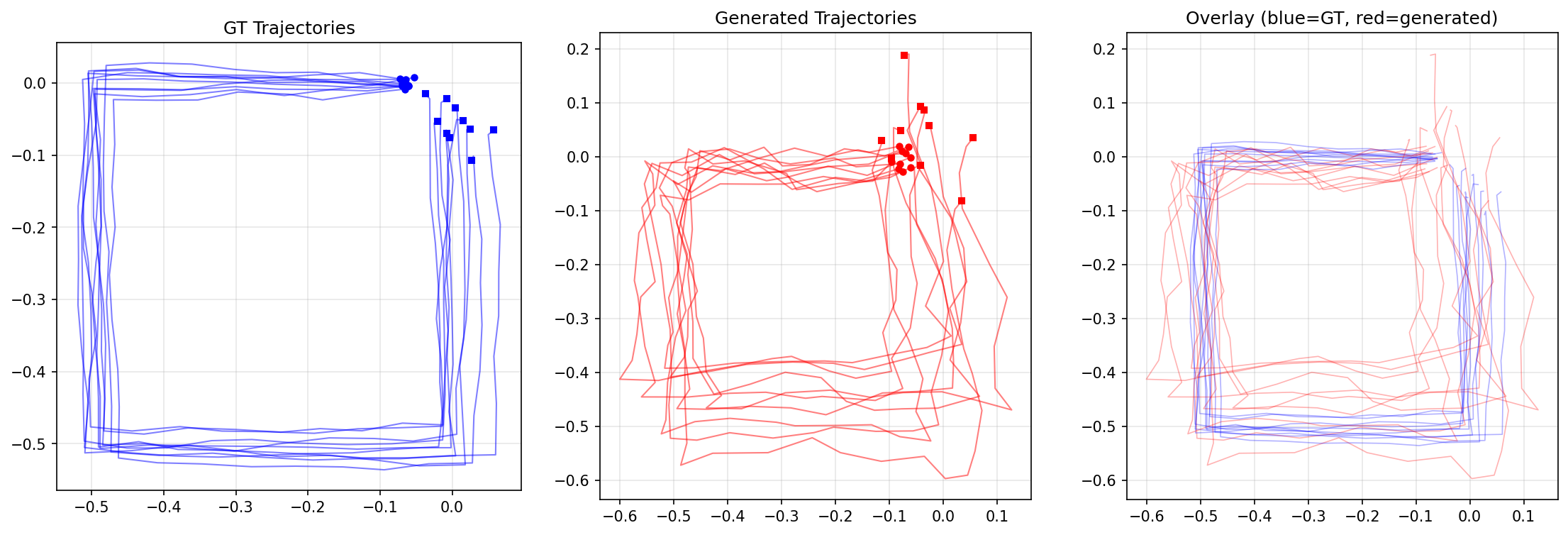} &
\includegraphics[width=0.22\textwidth, height=0.22\textwidth, trim={0 0 1500 0}, clip]{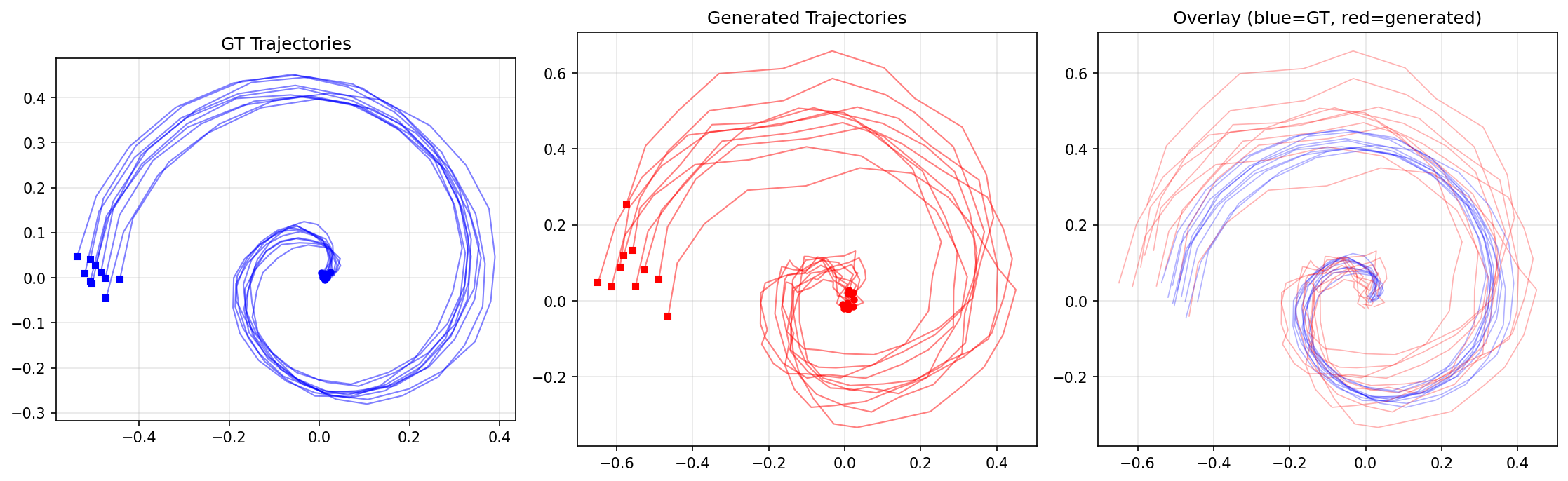} &
\includegraphics[width=0.22\textwidth, height=0.22\textwidth, trim={0 0 1490 0}, clip]{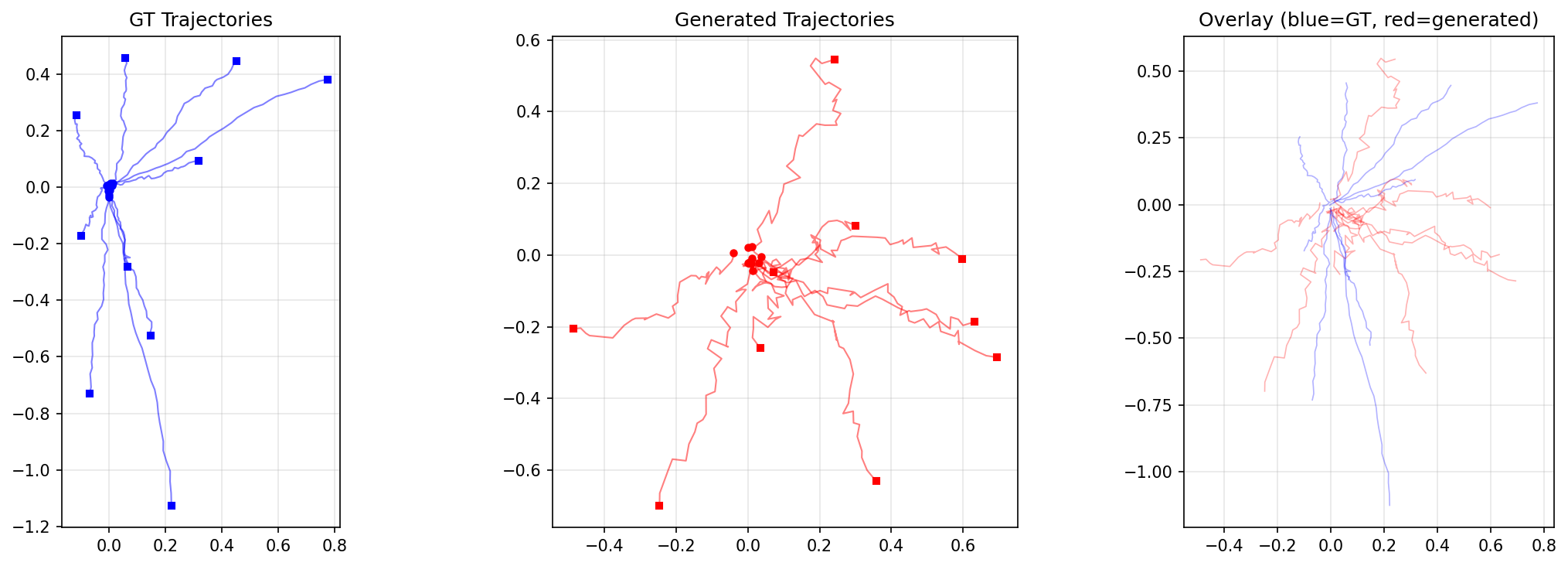} &
\includegraphics[width=0.22\textwidth, height=0.22\textwidth, trim={0 0 1480 0}, clip]{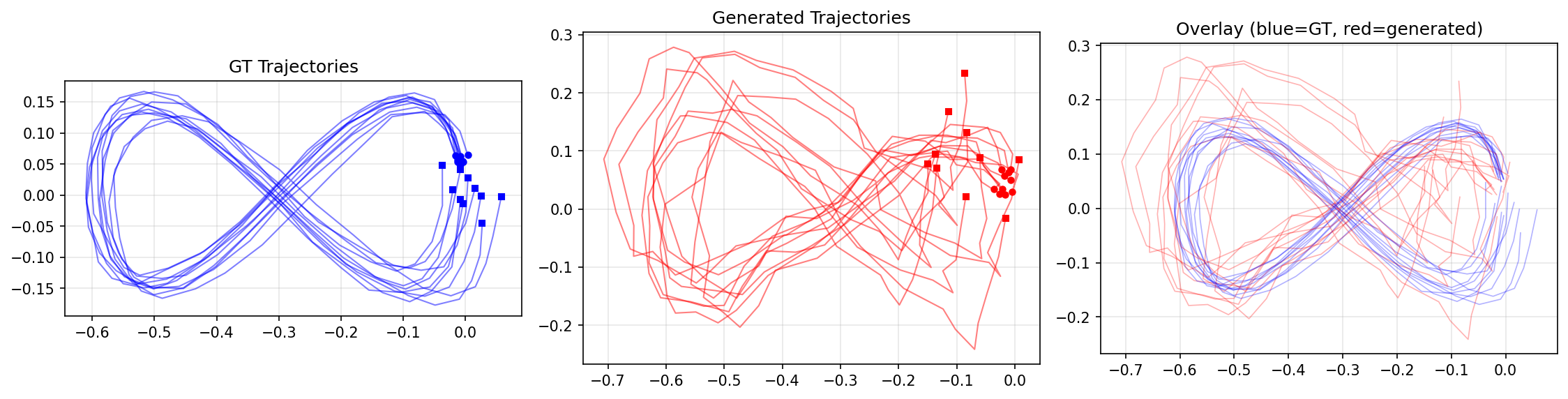} \\
\small (a) Square & \small (b) Spiral & \small (c) Line (radiating) & \small (d) Figure-8 \\
\end{tabular}
\caption{\textbf{2D robot-trajectory benchmark used for FMC ablation.} Each shape class contains multiple noisy demonstrations sharing a canonical motion pattern (blue curves show ground-truth trajectories). Shapes span repeating closed motion (Square, Figure-8), unbounded expansion (Spiral), and multi-modal starts (Line's radial variants). The policy is trained to generate trajectories matching the full distribution of each class.}
\label{fig:fmc_toy_dataset}
\end{figure}

To isolate the effect of FMC's perturbation hyperparameters without confounds from downstream task-success noise, we ablate FMC on a controlled 2D robot-trajectory benchmark. The dataset (\cref{fig:fmc_toy_dataset}) consists of noisy demonstrations grouped into seven canonical motion classes representative of common robot end-effector trajectories (circle, star, spiral, line, arc, figure-8, square). The policy must learn to generate trajectories that match the full distribution of each class.

The sweep varies three FMC hyperparameters:
\begin{itemize}[leftmargin=1.3em,itemsep=0.2ex,topsep=0.3ex]
    \item $s$ --- the \textbf{perturbation scale}: standard deviation of the Gaussian noise $\epsilon_1 \sim \mathcal{N}(0, sI)$ added to the clean action $x_0$ to produce the perturbed sample $x_0^p = x_0 + \epsilon_1$ (same $s$ as in \cref{sec:background_flow_matching}).
    \item $d$ --- the \textbf{drift strength}: magnitude of the temporally correlated drift (a normalized 2D random walk along the time axis) that structures how the perturbed trajectory accumulates offset from the clean one; iid noise corresponds to $d = 0$.
    \item $t_{\text{corrupt}}$ --- the \textbf{corruption time threshold}: minimum diffusion time at which the correction branch is active. For $t < t_{\text{corrupt}}$ the model trains with standard flow matching; for $t \geq t_{\text{corrupt}}$ it trains the correction target $v_{\text{corr}}$.
\end{itemize}

Each model is trained for 200k steps, and we evaluate 500 generated vs 1000 ground-truth trajectories per class using symmetric Chamfer distance in the flattened trajectory space. Chamfer is the appropriate signal here because FMC's correction induces a different noise-to-data bijection than the baseline, so paired endpoint drift is dominated by this change of mapping rather than trajectory quality. The configuration adopted in the main paper ($s{=}0.4, d{=}0.3, t_{\text{corrupt}}{\geq}0.5$) falls out as the winner of the sweep, with the strongest average Chamfer improvement and wins on 5 of 7 trajectory classes (\cref{tab:fmc_sweep}).

\begin{table}[h]
    \centering
    \caption{\textbf{Flow Matching with Correction (FMC) ablations on 2D robot trajectories.} We train each configuration for 200k steps on seven 2D trajectory classes representative of robot end-effector motions (circle, star, spiral, line, arc, figure-8, square) and evaluate 500 \emph{unconditionally generated} vs 1000 ground-truth trajectories per class using \emph{symmetric (two-way) Chamfer distance} --- averaging generated-to-GT and GT-to-generated nearest-neighbor distances (lower is better). We report the mean per-class relative change in Chamfer vs the linear flow-matching baseline, the number of classes on which the configuration beats the baseline, and the corresponding win rate. The last row is the configuration adopted in the main paper. \colorbox{cb-cream-yellow}{\textbf{Best}}.}
    \label{tab:fmc_sweep}
    \adjustbox{max width=\textwidth}{
    \begin{tabular}{l|rrr}
    \toprule
     Config & $\Delta$ Chamfer vs Base & Wins / 7 & Win Rate \\
    \midrule
    \textbf{Baseline (linear flow matching)}                                                 & $\pm 0.0\%$ & ---  & ---    \\
    \textbf{FMC ($s{=}0.08$, $d{=}0.2$, $t{\geq}0.25$)}                                      & $-0.1\%$    & 3/7  & 42.9\% \\
    \textbf{FMC ($s{=}0.2$,  $d{=}0.3$, $t{\geq}0.25$)}                                      & $-2.3\%$    & 4/7  & 57.1\% \\
    \textbf{FMC ($s{=}0.3$,  $d{=}0.3$, $t{\geq}0.5$)}                                       & $-1.2\%$    & 3/7  & 42.9\% \\
    \textbf{FMC ($s{=}0.4$,  $d{=}0.3$, $t{\geq}0.5$) \textit{(ours)}} & \cellcolor{cb-cream-yellow} $\mathbf{-4.0\%}$ & \cellcolor{cb-cream-yellow} \textbf{5/7} & \cellcolor{cb-cream-yellow} \textbf{71.4\%} \\
    \bottomrule
    \end{tabular}
    }
\end{table}

\begin{figure}[t]
\centering
\setlength{\tabcolsep}{3pt}
\renewcommand{\arraystretch}{1.0}
\begin{tabular}{@{}c@{\hspace{4pt}}ccc@{}}
 & \textbf{Spiral} & \textbf{Star} & \textbf{Rose} \\[0.4ex]
\rotatebox{90}{\footnotesize\textbf{Vanilla}} &
  \includegraphics[width=0.3\textwidth, height=0.3\textwidth, trim={1500 0 0 0}, clip]{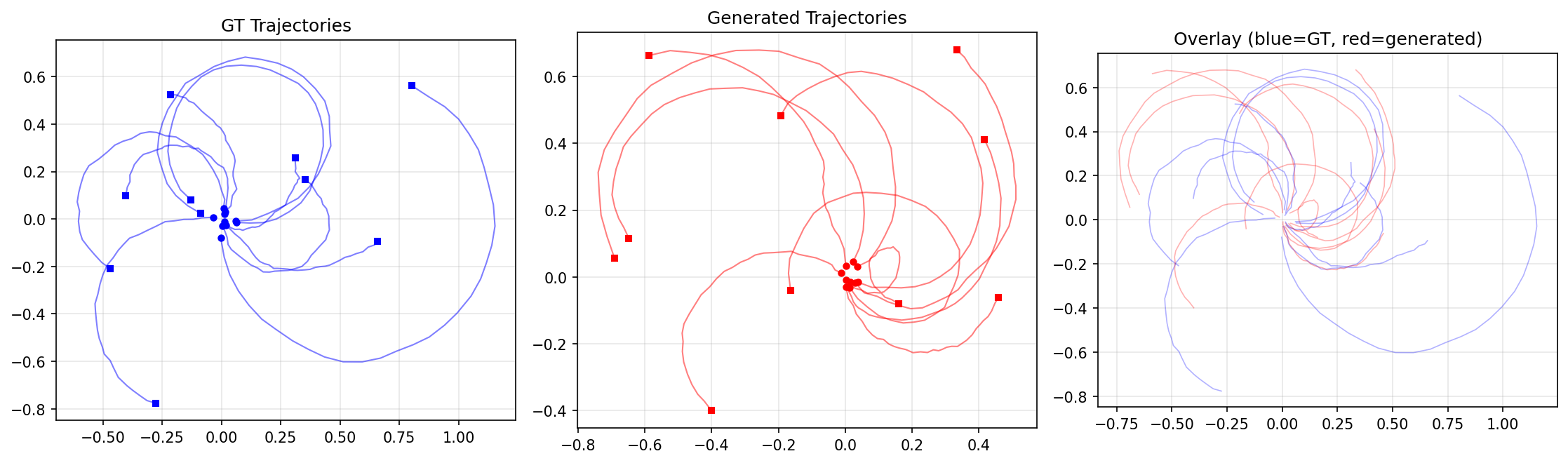} &
  \includegraphics[width=0.3\textwidth, height=0.3\textwidth, trim={1500 0 0 0}, clip]{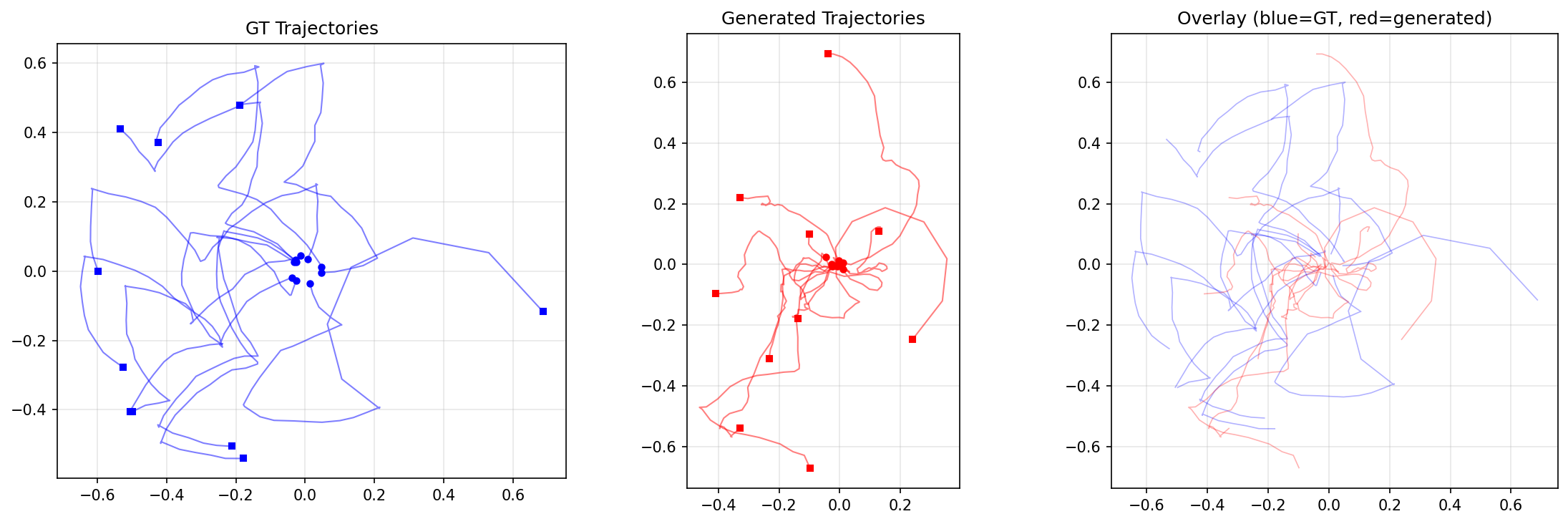} &
  \includegraphics[width=0.3\textwidth, height=0.3\textwidth, trim={1500 0 0 0}, clip]{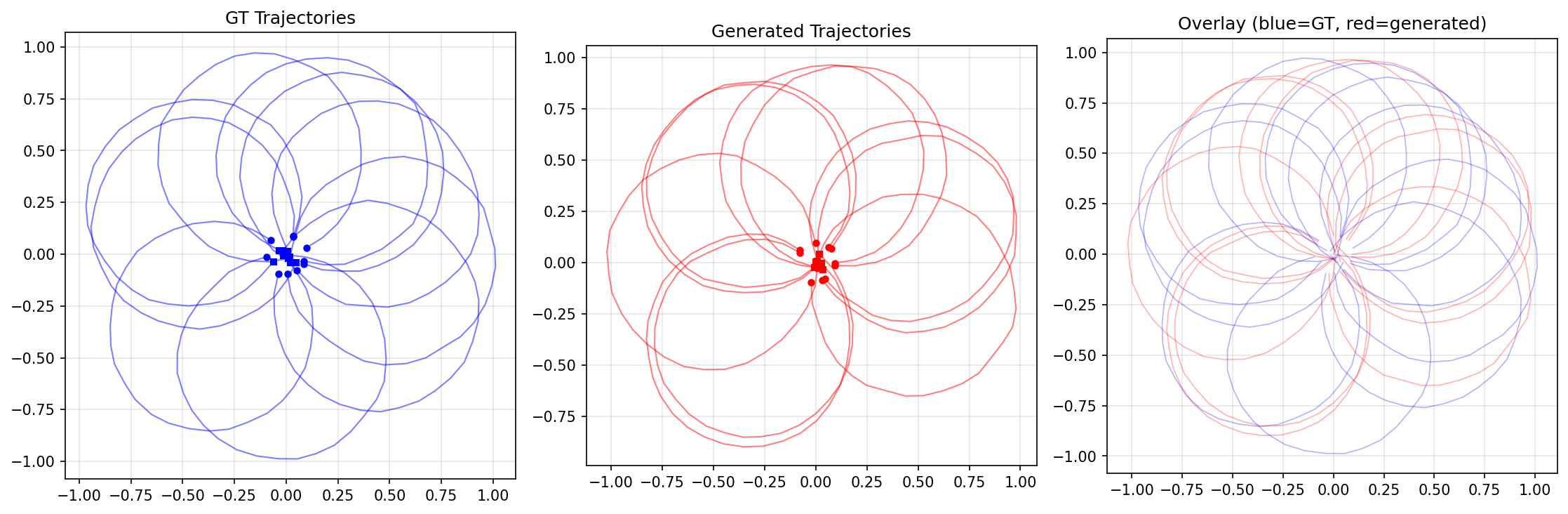} \\[0.3ex]
\rotatebox{90}{\footnotesize\textbf{FMC}} &
  \includegraphics[width=0.3\textwidth, height=0.3\textwidth, trim={1500 0 0 0}, clip]{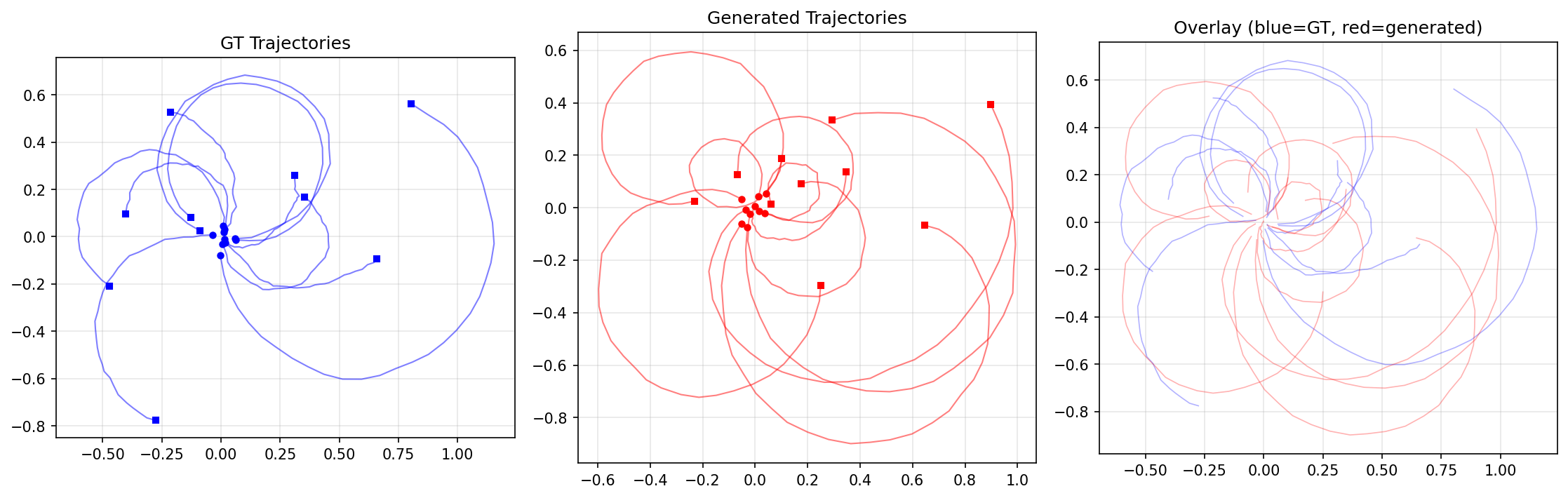} &
  \includegraphics[width=0.3\textwidth, height=0.3\textwidth, trim={1500 0 0 0}, clip]{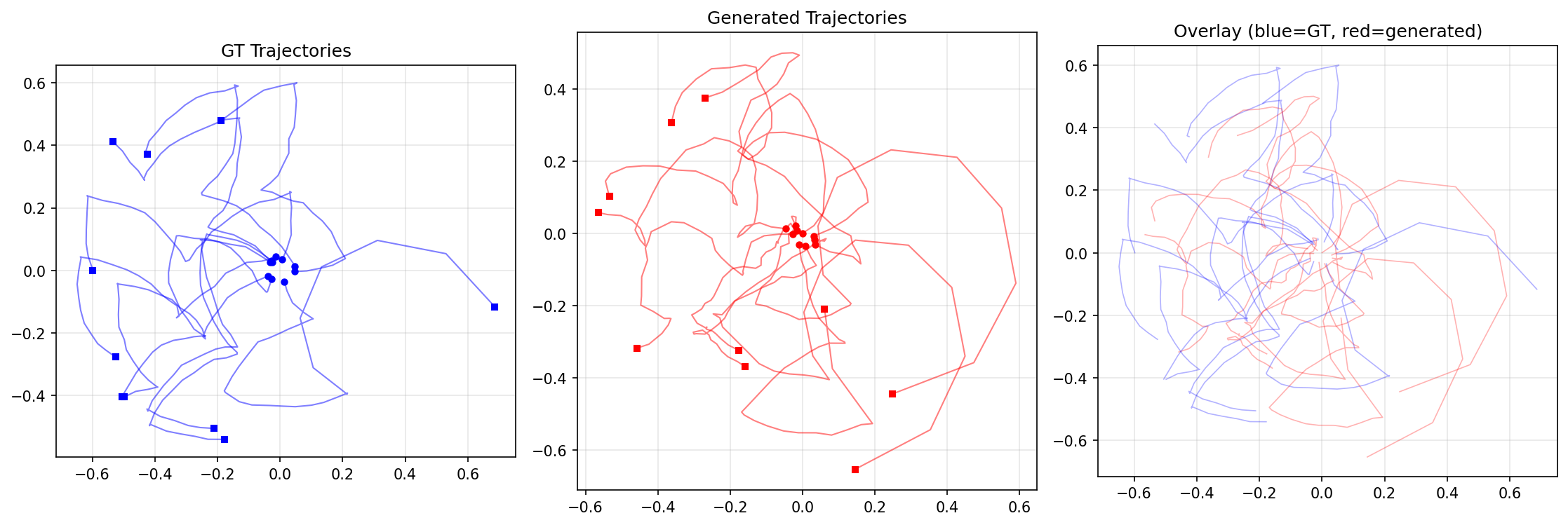} &
  \includegraphics[width=0.3\textwidth, height=0.3\textwidth, trim={1500 0 0 0}, clip]{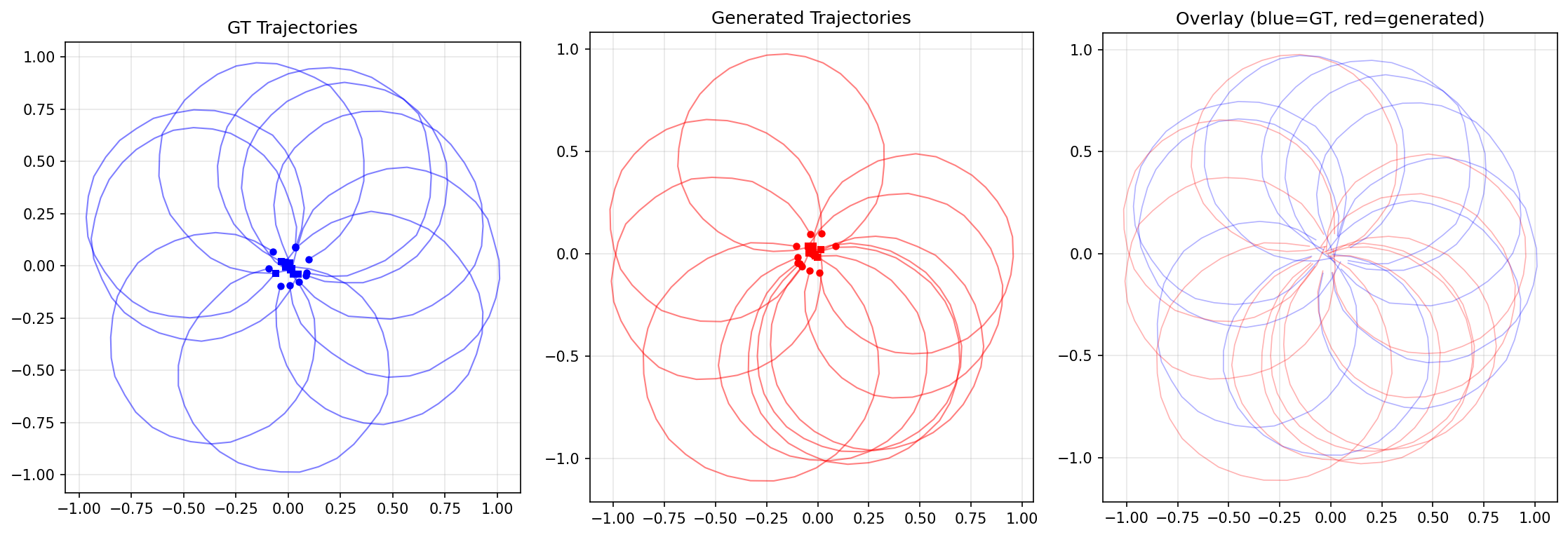} \\
\end{tabular}
\caption{\textbf{FMC vs vanilla flow matching on 2D robot trajectories (overlay view).} Blue curves are ground-truth demonstrations; red are model-generated trajectories, overlaid for direct comparison. Top row: vanilla flow matching frequently drifts off the demonstration manifold, truncates the shape, or collapses mode coverage. Bottom row: FMC tracks the ground-truth distribution more faithfully across all three shape classes, consistent with the quantitative Chamfer improvements in~\cref{tab:fmc_sweep}.}
\label{fig:fmc_toy_results}
\end{figure}

\subsection{Failure Case: Press Button}
\begin{figure*}[t]
  \centering
  \setlength{\tabcolsep}{2pt}
  \renewcommand{\arraystretch}{1.05}
  \begin{tabular}{@{}c c c c c@{}}
    \multicolumn{5}{@{}l@{}}{\textbf{Failure case: Press Button (target: 5 left, 2 middle, 1 confirm; actual: 5-\textcolor{cb-maroon}{3}-1)}}\\[0.5ex]
    \begin{minipage}[t]{0.192\textwidth}
      \centering
      \includegraphics[width=\linewidth]{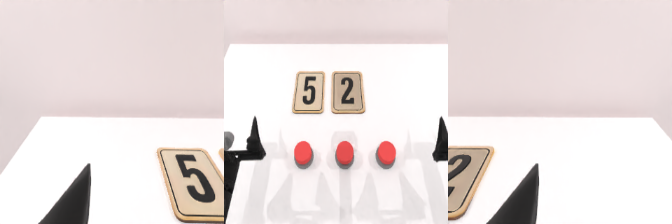}\\[0.3ex]
      \raggedright
      {\footnotesize\color{cb-maroon}\textbf{next subtask:}~left button.\par}
      {\footnotesize\color{cb-dark-blue}\textbf{memory:}~visible 5 2 \textbar{}\textbar{} left button 0/5 middle button 0/2 \textbar{}\textbar{} left button.\par}
    \end{minipage} &
    \begin{minipage}[t]{0.192\textwidth}
      \centering
      \includegraphics[width=\linewidth]{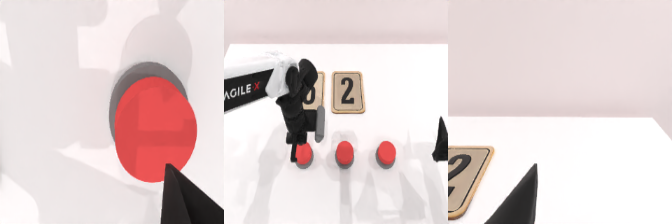}\\[0.3ex]
      \raggedright
      {\footnotesize\color{cb-maroon}\textbf{next subtask:}~middle button.\par}
      {\footnotesize\color{cb-dark-blue}\textbf{memory:}~visible 5 2 \textbar{}\textbar{} left button 5/5 middle button 0/2 \textbar{}\textbar{} middle button.\par}
    \end{minipage} &
    \begin{minipage}[t]{0.192\textwidth}
      \centering
      \includegraphics[width=\linewidth]{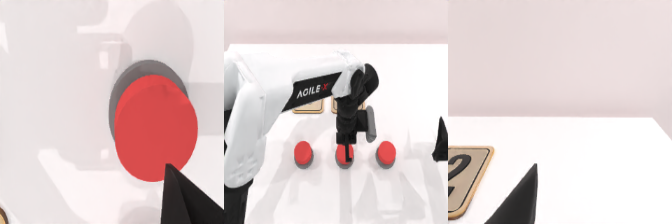}\\[0.3ex]
      \raggedright
      {\footnotesize\color{cb-maroon}\textbf{next subtask:}~middle button.\par}
      {\footnotesize\color{cb-dark-blue}\textbf{memory:}~visible 5 2 \textbar{}\textbar{} left button 5/5 middle button 1/2 \textbar{}\textbar{} middle button. next: confirm.\par}
    \end{minipage} &
    \begin{minipage}[t]{0.192\textwidth}
      \centering
      \includegraphics[width=\linewidth]{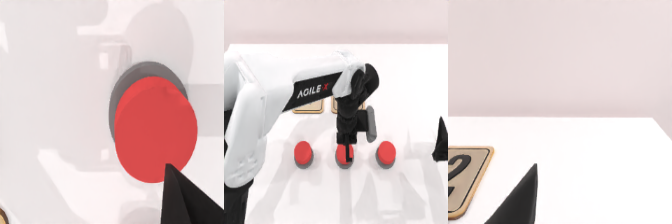}\\[0.3ex]
      \raggedright
      {\footnotesize\color{cb-maroon}\textbf{next subtask:}~confirm.\par}
      {\footnotesize\color{cb-dark-blue}\textbf{memory:}~visible 5 2 \textbar{}\textbar{} left button 5/5 middle button 2/2 \textbar{}\textbar{} confirm. next: done.\par}
    \end{minipage} &
    \begin{minipage}[t]{0.192\textwidth}
      \centering
      \includegraphics[width=\linewidth]{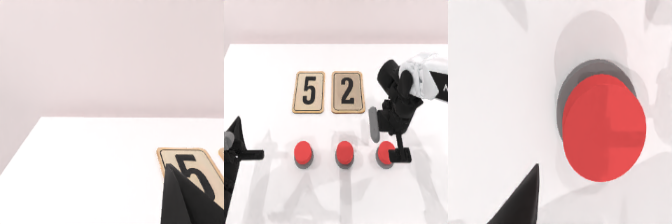}\\[0.3ex]
      \raggedright
      {\footnotesize\color{cb-maroon}\textbf{next subtask:}~confirm.\par}
      {\footnotesize\color{cb-dark-blue}\textbf{memory:}~visible 5 2 \textbar{}\textbar{} left button 5/5 middle button 2/2 \textbar{}\textbar{} confirm. next: done.\par}
    \end{minipage}\\
  \end{tabular}
  \caption{\textbf{Timing-lag failure on Press Button.} Target counts are $(\text{left}=5,\text{middle}=2,\text{confirm}=1)$. The \textcolor{cb-dark-blue}{memory} readout tracks the intended count correctly (\texttt{left 5/5, middle 2/2, confirm, next: done}), but the action executor over-shoots by one press on middle, yielding 3 physical middle presses instead of 2. The language-memory loop is correct; the task-completion channel fires one step late, so the action expert continues the current subtask for one extra press.}
  \label{fig:failure_press_button}
\end{figure*}

We observe cases where the language-memory loop tracks state correctly but the action executor over- or under-shoots by one press. On a Press Button episode with target counts $(\text{left}=5,\text{middle}=2,\text{confirm}=1)$, the memory readout progresses cleanly to \texttt{left 5/5, middle 2/2, confirm, next: done}, yet the physical arm registers three middle presses instead of two (\cref{fig:failure_press_button}). The internal count is right; the task-completion channel fires one step late, so the action expert continues the current subtask for one extra press.

\section{Limitations}
\label{sec:limitations}
\coolname{} exhibits four main failure modes.
\emph{(i) Boundary misfires}: the memory loop tracks state correctly but the task-completion signal $\beta$ fires one step late or early, causing an extra or missed press (\cref{fig:failure_press_button}; detailed case in the Failure Case subsection above).
\emph{(ii) Incorrect subtask prediction}: on near-identical observations with different underlying memory state, the language stream occasionally emits the wrong next subtask.
\emph{(iii) Grasp precision}: complex-geometry objects and cluttered scenes still bound task success.
\emph{(iv) Inference speed}: generating an action chunk of 32 with the language expert at 10 denoising steps runs at 123\,ms (76\,ms without language; \cref{sec:arch}), which may be too slow for high-frequency closed-loop control.

\section{Additional Future Work}
\label{sec:future_work}

Beyond the directions in the main paper, the \coolname{} formulation opens several concrete extensions.

\parahead{Pretrained VLM initialization for the language DiT} Our language DiT is trained from scratch. Initializing it from a pretrained vision-language model would inject broad semantic priors into the language-memory loop without altering the shared-denoising interface.

\parahead{Retrieval-augmented memory with a learned selection criterion} The current textual memory is a single running string. Coupling it with an external episode log and a learned selection policy (choosing which past states or memory snippets to re-inject into the prompt) would extend the loop to long horizons and cross-episode settings.

\parahead{FMC from a replay buffer of the robot's own mistakes} FMC currently perturbs clean actions with Gaussian noise $\epsilon_1$. A stronger variant is to construct the perturbation from the model's actual rollout deviations: store rollout error pairs and train the correction target to recover the nominal path from these empirical errors rather than synthetic noise.

\parahead{Confidence-based flow matching} A learned confidence head $c\in(0,1]$ over the predicted velocity, trained jointly with a log-penalty to prevent trivial collapse, would let the policy down-weight uncertain action predictions and expose an inference-time signal for guarded execution and targeted re-planning.

\parahead{Multi-keyframe visual memory} Our visual-memory stream already supports an arbitrary number of frames (\cref{sec:arch}), but our experiments use a single anchor frame. A natural extension is to maintain a running history of keyframes (either at fixed intervals or selected online by the model) so the policy can attend to a bank of task-relevant past observations rather than a single snapshot at episode start.

\parahead{Balancing multiple conditioning signals} When several conditioning streams are provided (e.g., a multi-keyframe visual-memory history, textual memory, current observation, and proprioception), careful training is required to prevent the policy from shortcutting to whichever stream is most predictive on the training distribution and ignoring the rest: for instance, collapsing onto the most recent keyframe while ignoring earlier ones, or ignoring the textual memory when the current frame alone is sufficient in-distribution. Explicit stream dropout, per-stream masking curricula, and information-theoretic regularizers that encourage the action expert to attend to the situation-relevant signal are promising directions for scaling \coolname{} to richer memory and observation channels.

\parahead{Safety via language traces of intent} Because \coolname{} emits the next subtask and textual memory \emph{before} the corresponding action is executed, the generated string is effectively a human-readable trace of the robot's intention. This exposes a natural handle for safe deployment: downstream guards can veto or re-plan when the declared intention violates a constraint, operators can audit rollouts via the language log, and the trace supplies dense data for intention-conditioned alignment. Making this trace faithful (grounded in the actually-executed action), tamper-resistant, and expressive enough to cover safety-critical constraints is an important direction for future research.




\end{document}